\documentclass[twoside,11pt]{article}

\usepackage{jmlr2e}
\usepackage{amsmath}
\usepackage{graphicx}
\usepackage{natbib}
\RequirePackage{hyperref}
\usepackage{endnotes}
\usepackage{amssymb}
\usepackage{pifont}
\usepackage[english]{babel}
\usepackage{microtype}
\usepackage{subcaption}
\usepackage{changes}
\usepackage{algorithm}
\usepackage{algpseudocode}
\usepackage{todonotes}
\usepackage[margins]{trackchanges}
\usepackage{booktabs}
\addeditor{GT}
\addeditor{TR}

\ShortHeadings{Improved Gibbs Sampling Parameter Estimators for LDA}{Papanikolaou, Foulds, Rubin and Tsoumakas}
\firstpageno{1}

\begin{document}

\title{Dense Distributions from Sparse Samples:\\Improved Gibbs Sampling Parameter Estimators for LDA}

\author{\name Yannis Papanikolaou \email ypapanik@csd.auth.gr \\
       \addr Department of Informatics\\
       Aristotle University of Thessaloniki\\
       Thessaloniki, Greece\\
       \\
       \name James R. Foulds \email jfoulds@ucsd.edu\\
       \addr California Institute for Telecommunications and Information Technology\\ 
       University of California\\
       San Diego, CA, USA\\
       \\
       \name Timothy N. Rubin \email tim.rubin@gmail.com\\
       \addr SurveyMonkey\\
       San Mateo, CA, USA\\
        \\
		\name Grigorios Tsoumakas \email greg@csd.auth.gr\\
       \addr Department of Informatics\\
       Aristotle University of Thessaloniki\\
       Thessaloniki, Greece}

\editor{tba}

\maketitle

\begin{abstract}
We introduce a novel approach for estimating Latent Dirichlet Allocation (LDA) parameters from collapsed Gibbs samples (CGS), by leveraging the full conditional distributions over the latent variable assignments to efficiently average over multiple samples, for little more computational cost than drawing a single additional collapsed Gibbs sample. 
Our approach can be understood as adapting the soft clustering methodology of Collapsed Variational Bayes (CVB0) to CGS parameter estimation, in order to get the best of both techniques.
Our estimators can straightforwardly be applied to the output of any existing implementation of CGS, including modern accelerated variants.
We perform extensive empirical comparisons of our estimators with those of standard collapsed inference algorithms on real-world data for both unsupervised LDA and Prior-LDA,
a supervised variant of LDA for multi-label classification. Our results show a consistent advantage of our approach over traditional CGS under all experimental conditions, and over CVB0 inference in the majority of conditions. More broadly, our results highlight the importance of averaging over multiple samples in LDA parameter estimation, and the use of efficient computational techniques to do so.
\end{abstract}


\section{Introduction}
\label{sec:introduction}

Latent Dirichlet Allocation (LDA) \citep{Blei:2003:LDA:944919.944937} is a probabilistic model which organizes and summarizes large corpora of text documents by automatically discovering the semantic themes, or \emph{topics}, hidden within the data.
Since the model was introduced by \cite{Blei:2003:LDA:944919.944937}, 
LDA and its extensions have been successfully applied to many other data types and application domains, including bioinformatics \citep{zheng2006identifying}, computer vision \citep{4408965}, and social network analysis \citep{4258697}, in addition to text mining and analytics \citep{4781095}.
It has also been the subject of numerous adaptations and improvements, for example to deal with supervised learning tasks \citep{blei2007supervised,Ramage:2009:LLS:1699510.1699543,zhu2009medlda},  time and memory efficiency issues \citep{newman2007distributed,Porteous:2008:FCG:1401890.1401960,yao2009efficient} and large or streaming data settings \citep{Gohr09topicevolution,hoffman2010online,Rubin:2012:STM:2339279.2339301,Foulds:2013:SCV:2487575.2487697}.

When training an LDA model on a collection of documents, two sets of parameters are primarily estimated: the \emph{distributions over word types} $\phi$ for topics, and the \emph{distributions over topics} $\theta$ for documents.
After more than a decade of research on LDA training algorithms, the Collapsed Gibbs Sampler (CGS) \citep{griffiths_steyvers04} remains a popular choice for topic model estimation, and is arguably the de facto industry standard technique for commercial and industrial applications. Its success is due in part to its simplicity, along with the availability of efficient implementations \citep{McCallumMALLET,smola2010architecture} leveraging sparsity \citep{yao2009efficient,li2014reducing} and parallel architectures \citep{newman2007distributed}. It exhibits much faster convergence than the naive uncollapsed Gibbs sampler of \citet{pritchard2000inference}, as well as the partially collapsed Gibbs sampling scheme \citep{asuncion2011distributed}. The CGS algorithm marginalizes out parameters $\phi$ and $\theta$, necessitating a procedure to recover them from samples of the model's latent variables $\mathbf{z}$. \citet{griffiths_steyvers04} proposed such a procedure, which, while very successful, has not been revisited in the last decade to our knowledge. Our goal is to improve upon this procedure.

In this paper, we propose a new method for estimating topic model parameters $\phi$ and $\theta$ from collapsed Gibbs samples. Our approach approximates the posterior mean of the parameters, which leads to increased stability and statistical efficiency relative to the standard estimator. Crucially, our approach leverages the \emph{distributional information given by the form of the Gibbs update equation} to implicitly average over multiple Markov chain Monte Carlo (MCMC) samples, with little more computational cost than that which is required to compute a single sample.  As such, our work is focused on the realistic practical scenario where we can afford a moderate burn-in period, but cannot afford to compute and average over enough post burn-in MCMC samples to accurately estimate the parameters' posterior mean.  In our experiments, this requires several orders of magnitude further sampling iterations beyond burn-in.

Our use of the full conditional distribution of each topic assignment is reminiscent of the CVB0 algorithm for collapsed variational inference  \citep{Asuncion:2009:SIT:1795114.1795118}.  The update equations of CVB0 bear resemblance to those of CGS, except that they involve deterministic updates of dense probability distributions, while CGS draws sparse samples from similar distributions, with corresponding trade-offs in execution time and convergence behavior, as we study in Section \ref{sec:exp2}. 
Our approach can be understood as adapting this dense \emph{soft clustering} methodology of the CVB0 algorithm to CGS parameter estimation, leveraging the corresponding uncertainty information, as in CVB0, but within a Markov chain Monte Carlo framework, in order to draw on the benefits of both techniques.  Our approach does not incur the memory overhead of CVB0, relative to CGS.  The properties of CVB0, CGS, and our proposed estimators are summarized in Table \ref{tab:properties}.
\begin{table}[t]
\centering
\begin{tabular}{ccccccc}
\toprule
        & training &        parameter&               update&          inference&              memory \\
        & phase    &        recovery &               rule&                algorithm&                per token \\
\midrule
CVB0    & dense&               dense&                   deterministic&   variational Bayes&                  $O(K)$ \\
CGS     & sparse&              sparse&                  random&          MCMC&                            $O(1)$ \\
CGS$_p$   & sparse&            dense&                   random&          MCMC&                           $O(1)$ \\
\bottomrule
\end{tabular}
\caption{\label{tab:properties} Properties of CVB0, CGS, and our proposed CGS$_p$ method.}
\end{table}


It is important to note that we do not modify the CGS algorithm, and thus do not affect its runtime or memory requirements.  Rather, we modify the \emph{procedure for recovering parameter estimates from collapsed Gibbs samples}. After running the standard CGS algorithm (or a modern accelerated approximate implementation) for a sufficient number of iterations, we obtain the standard document-topic and topic-word count matrices. At this point, instead of using the standard CGS parameter estimation equations to calculate $\theta$ and $\phi$, we employ our proposed estimators. As a result, our estimators can be plugged in to the output of any CGS-based inference method, including modern variants of the algorithm such as Sparse LDA \citep{yao2009efficient}, Metropolis-Hastings-Walker \citep{li2014reducing},  LightLDA \citep{yuan2015lightlda}, or WarpLDA \citep{chen2016warplda}, allowing for easy and wide adoption of our approach by the research community and in industry applications. 
 Our algorithm is very simple to implement, requiring only slight modifications to existing CGS code.
Popular LDA software packages such as MALLET \citep{McCallumMALLET} and Yahoo\_LDA \citep{smola2010architecture} could straightforwardly be extended to incorporate our methods, leading to improved parameter estimation in numerous practical topic modeling applications. In Section \ref{sec:malletWarpExp}, we provide an application of our estimators to both MALLET's Sparse LDA implementation, and to WarpLDA.  When paired with a sparse CGS implementation, our approach gives the best of two worlds: making use of the \emph{full uncertainty information encoded in the dense Gibbs sampling transition probabilities} in the final parameter recovery step, while still leveraging \emph{sparsity to accelerate the CGS training process}. 

In extensive experiments, we show that our approach leads to improved performance over standard CGS parameter estimators in both unsupervised and supervised LDA models.  Our approach also outperforms the CVB0 algorithm under the majority of experimental conditions. The computational cost of our method is comparable that of a single (dense) CGS iteration, and we show that even when a state-of-the-art sparse implementation of the CGS training algorithm is used, this is a price well worth paying.  Our results further illustrate the benefits of averaging over multiple samples for LDA parameter estimation.
The contributions of our work can be summarized as follows:

\begin{itemize}
\item We propose a novel, theoretically motivated method for improved estimation of the topic-level and document-level parameters of an LDA model 
based on collapsed Gibbs samples.
\item We present an extensive empirical comparison of our proposed estimators against those of standard CGS as well as the CVB0 algorithm \citep{Asuncion:2009:SIT:1795114.1795118} ---a variant of the Collapsed Variational Bayesian inference (CVB) method---in both unsupervised and supervised learning settings, demonstrating the benefits of our approach.
\item We provide an additional experimental comparison of the CGS and CVB0 algorithms regarding their convergence behavior, to further contextualize our empirical results.
\end{itemize}

Our theoretical and experimental results lead to three primary conclusions of interest to topic modeling practitioners:
\begin{itemize}
\item Averaging over multiple CGS samples to construct a point estimate for LDA parameters is beneficial to performance, in both the unsupervised and supervised settings, in cases where identifiability issues can be safely resolved.
\item Using CVB0-style soft clustering to construct these point estimates is both valid and useful in an MCMC setting, and corresponds to implicitly averaging over many samples, thereby improving performance with little extra computational cost.  This ports some of the benefits of CVB0's use of dense uncertainty information to CGS, while retaining the sparsity advantages of that algorithm during training.
\item Increasing the number of topics increases the benefit of our soft clustering/averaging approach over traditional CGS. While CVB0 outperforms CGS with few topics and a moderate number of iterations, CGS-based methods otherwise outperform CVB0, especially when using our improved estimators.
\end{itemize}

As this paper deals with LDA in both the unsupervised and the supervised setting, we will treat the following terms as equivalent when describing the procedure of learning an LDA model from training data: `estimation', `fitting', and `training'. Since we are in a Bayesian setting the word `inference' will refer to the recovery of parameters, which are understood to be random variables, and/or the latent variables, and to the computation of the posterior distributions of parameters and latent variables. 
The term `prediction' will be applied to the case of predicting on test data (i.e. data that was unobserved during training). Additionally, since Prior-LDA is essentially a special case of unsupervised LDA, we will focus our algorithm descriptions on the case of standard unsupervised LDA and specify the exceptions as they apply to Prior-LDA. Furthermore, we note here that we will be using the CGS algorithm described by \citet{griffiths_steyvers04}, which approximates the LDA parameters of interest through iterative sampling in a Markov chain Monte Carlo (MCMC) procedure, unless otherwise specified.

The rest of this paper is organized as follows: Section \ref{sec:background} describes backround information on the LDA model, Bayesian inference, and the CGS and the CVB0 inference methods, as well as the Prior-LDA model, providing an overview of the related literature. 
Sections \ref{sec:cgsp} and \ref{sec:phi_p} introduce our estimators for the relevant LDA parameters.
Sections \ref{sec:expsunsup} and \ref{sec:multi} describe our experiments comparing the estimators of our method to those of CGS and CVB0 in unsupervised and supervised learning settings, respectively. Finally, Section \ref{sec:concl} summarizes the main contributions and conclusions of our work.

\section{Background and Related Work}
\label{sec:background}
In this section we provide the background necessary to describe our methods, and also establish the relevant notation. 
\subsection{Latent Dirichlet Allocation}
\label{sec:lda}
LDA is a hierarchical Bayesian model that describes how a corpus of documents is generated via a set of unobserved {\em topics}. Each topic is parameterized by a discrete probability distribution over word types, which captures some coherent semantic theme within the corpus\footnote{A discrete distribution a multinomial distribution for which a single value is drawn, i.e. where $N=1$.}. Here, we assume that this set of word types is the same for all topics (i.e. we have a single, shared vocabulary). Furthermore, LDA assumes that each document can be described as a discrete distribution over these topics. The process for generating each document then involves first sampling a topic from the document's distribution over topics, and then sampling a word token from the corresponding topic's distribution over word types.

Formally, let us denote as $V$ the size of the vocabulary (the set of unique word types) and as $D_{Train} = |\mathcal{D}_{Train}|$ and $D_{Test}= |\mathcal{D}_{Test}|$ the number of training and testing documents respectively. We will use $K$ to denote the number of topics. Similarly, $v$ denotes a word type, $w_i$ a word token, 
$k$ a topic (or in the case of Prior-LDA, a label), and $z_i$ a topic assignment. The parameter $\phi_{kv},\, k \in\{1...K\},\, v \in\{1...V\} $ represents the probability of word type $v$ under topic (or label) $k$, and $\theta_{dk}$ represents the probability of topic (or label) $k$ for document $d$. Additionally $\alpha_k$ will denote the parameter of the Dirichlet prior on $\theta$ for topic $k$ and $\beta_v$ the parameter of the Dirichlet prior on $\phi$ for word type $v$. Unless otherwise noted we assume for simplicity that the two priors are symmetric, so all the elements of the parameter vector $\boldsymbol{\alpha}$ will have the same value (the same holds for the parameter vector $\boldsymbol{\beta}$). 
The term $N_d$ denotes the number of word tokens in a given document $d$. Lastly, in the multi-label setting, $\mathcal{K}_d$ stands for the set of labels that are associated with $d$. This notation is summarized in Table \ref{tbl:notation}.

 LDA assumes that a given corpus of documents has been generated as follows:
\begin{itemize}
\item For each topic $k \in \{1\ldots K\}$, sample a discrete distribution $\phi_k$ over the word types of the corpus from a Dirichlet$(\boldsymbol{\beta})$.
\item For each document $d$,
	\begin{itemize}
    \item Sample a discrete distribution $\theta_d$ over topics from a Dirichlet($\boldsymbol{\alpha}$).
    \item Sample a document length $N_d$ from a Poisson($\xi$), with $\xi \in \mathbb R_{> 0}$
\item For each word token $w_i$ in document $d$, $i \in \{1...N_d\}$:
		\begin{itemize}
    	\item Sample a topic $z_i=k$ from discrete$(\theta_d)$.
    	\item Sample a word type $w_i=v$ from discrete$(\phi_{z_i})$. 
    	\end{itemize}
	\end{itemize} 
\end{itemize}

\begin{table}[tb]
\centering
\begin{tabular}{cl}
\toprule
\noalign{\smallskip}  
$V$& the size of the vocabulary\\
$K$& the number of topics (or labels)\\
$D_{Train}$& the number of training documents (similarly $D_{Test}$)\\
$\mathcal{D}_{Train}$& the set of training documents (similarly $\mathcal{D}_{Test}$)\\
$d$& a document\\
$v$& a word type\\
$w_i$& a single word token, i.e. an instance of a word type\\
$\mathbf{w}_d$& the vector of word assignments in document $d$\\
$k$& a topic (label)\\
$z_i$ & the topic assignment to a word token $w_i$ \\
$\mathbf{z}_d$ & the vector of topic assignments to all word tokens of a document \\
$z_{djk}$ & binary indicator variable, equals 1 IFF the $j$th word in document $d$ \\
&is assigned to topic $k$ \\
$N_d$& number of word tokens in $d$\\
$\mathcal{K}_d$& set of topics (or labels) in $d$\\
$n_{kv}$& number of times that word type $v$ has been assigned to topic $k$ across\\
&the corpus\\
$n_{dk}$& number of word tokens in $d$ that have been assigned to topic $k$\\
$\phi_k$& word type distribution for topic $k$\\
$\theta_d$& topic distribution for document $d$\\
$\phi_{kv}$ & topic-word type probability \\
$\theta_{dk}$ & document-topic probability\\
$\alpha_k$& Dirichlet prior hyperparameter on $\theta$ for topic $k$ \\ 
$\beta_v$& Dirichlet prior hyperparameter on $\phi$ for word type $v$\\
$\gamma_{dik}$&CVB0 variational probability of word token $w_i$ in document $d$ being\\
&assigned to topic $k$\\

\bottomrule
\end{tabular}
\caption{Notation used throughout the article.}
\label{tbl:notation}
\end{table}

The goal of inference is to estimate the aforementioned $\theta$ and $\phi$ parameters---the discrete distributions for documents over topics, and topics over word types, respectively. In doing so, we learn a lower-dimensional representation of the structure of all documents in the corpus in terms of the topics. In the unsupervised learning context, this lower-dimensional representation of documents is useful for both summarizing documents and for generating predictions about future, unobserved documents. In the supervised, multi-label learning context, extensions of the basic LDA model---such as Prior-LDA ---put topics into one-to-one correspondence with labels, and the model is used for assigning labels to test documents.

\subsection{Bayesian Inference, Prediction, and Parameter Estimation}

LDA models are typically trained using Bayesian  inference techniques. Given the observed training documents $\mathcal{D}_{Train}$, the goal of Bayesian inference is to ``invert'' the assumed generative process of the model, going ``backwards'' from the data to recover the parameters, by inferring the posterior distribution over them. In the case of LDA, the posterior distribution over parameters and latent variables is given by Bayes' rule as 
\begin{align*}
p(\phi, \theta, \mathbf{z}|\mathcal{D}_{Train}, \boldsymbol{\alpha}, \boldsymbol{\beta}) = \frac{p(\mathcal{D}_{Train}|\phi, \theta, \mathbf{z}) p(\phi, \theta, \mathbf{z}| \boldsymbol{\alpha}, \boldsymbol{\beta})}{p(\mathcal{D}_{Train}|\boldsymbol{\alpha}, \boldsymbol{\beta})} \mbox{ .}
\end{align*}
Being able to make use of the uncertainty encoded in the posterior distribution is a key benefit of the Bayesian approach. In the ideal procedure, having computed the posterior it can be used to make predictions on held-out data via the posterior predictive distribution, which computes the probability of new data by averaging over the parameters, weighted by their posterior probabilities. For a test document $\mathbf{w}^{(new)}$, under the LDA model we have
\begin{align*}
p(\mathbf{w}^{(new)}|\mathcal{D}_{Train}, \boldsymbol{\alpha}, \boldsymbol{\beta}) = \int \int p(\mathbf{w}^{(new)}| \phi, \theta^{(new)}) p(\theta^{(new)}|\boldsymbol{\alpha}) d \theta^{(new)}  p(\phi|\mathcal{D}_{Train}, \boldsymbol{\alpha}, \boldsymbol{\beta}) d \phi \mbox{ .}
\end{align*}
Similarly, for supervised variants of LDA (discussed below), the ideal procedure is to average over all possible parameter values when predicting labels of new documents. In practice, however, it is intractable to compute or marginalize over the posterior, necessitating approximate inference techniques such as MCMC methods, which sample from the posterior using an appropriate Markov chain, and hence approximate it with the resulting set of samples. In particular, \citet{nguyen2014sometimes} advocate using multiple samples to approximate the posterior, and averaging over the corresponding approximation to the posterior predictive distribution in order to make predictions for new data.

\begin{algorithm}[h!]
\renewcommand{\algorithmicrequire}{\ \ \ \ \ \textbf{Input:}}
\caption{Pseudocode for our proposed estimators. For simplicity, we consider only the case where a single CGS sample $\mathbf{z}$ is used. \label{alg:methods}
}
\begin{algorithmic}
\Function{$\theta^p$}{}
\Require Burned-in CGS sample $\mathbf{z}$ with associated count matrices, documents $\mathcal{D}$
 \For {$d=1$ to $D$}
\State $\hat{\theta}^p_{d,\mathbf{:}} := \alpha^\intercal$
\For {$j = 1$ to $N_d$}
	\State $v := w_{dj}$ \Comment \tiny Retrieve the word token from the word position j in d.
	\normalsize
    \For {$k=1$ to $K$} 
     	\State  $p_{djk} := \frac{n_{kv\neg{dj}}+\beta_v}{\sum\limits_{v^\prime = 1}^V({n_{kv^\prime\neg{dj}}+\beta_{v^\prime})}} \cdot (n_{dk\neg{dj}}+\alpha_k)$ \Comment{\tiny{During testing, the first term is replaced by $\phi_{kv}$}} 
        \normalsize
     \EndFor 
     \State $p_{dj,\mathbf{:}} := p_{dj,\mathbf{:}} \mbox{ } ./ \mbox{ sum}(p_{dj,\mathbf{:}})$  \Comment{\tiny{Normalize such that $\sum_{k=1}^Kp_{djk} = 1$}, $./$ denotes elementwise division.} \normalsize
\State $\hat{\theta}^p_{d,\mathbf{:}} := \hat{\theta}^p_{d,\mathbf{:}} +  p_{dv,\mathbf{:}}^\intercal$
\EndFor

\State $\hat{\theta}^p_{d,\mathbf{:}} := \hat{\theta}^p_{d,\mathbf{:}} \mbox{ } ./ \mbox{ sum}(\hat{\theta}^p_{d,\mathbf{:}})$ \Comment{\tiny{Normalize $\hat{\theta}^p_{d,\mathbf{:}}$ such that $\sum_{k=1}^K \hat{\theta}^p_{dk}=1$}} \normalsize 
\EndFor
\State \Return $\hat{\theta}^p$
\EndFunction

\State
\State 
\Function{$\phi^p$}{}
\Require Burned-in CGS sample $\mathbf{z}$ with associated count matrices, documents $\mathcal{D}_{Train}$
\For {$k=1$ to $K$}
	\State $\hat{\phi}^p_{k,\mathbf{:}} := \beta^\intercal$
\EndFor
\For {$d=1$ to $D$}
\For {$j = 1$ to $N_d$}
	\State $v := w_{dj}$
	\For {$k=1$ to $K$}
     	\State  $p_{djk} = \frac{n_{kv\neg{dj}}+\beta_v}{\sum\limits_{v^\prime = 1}^V({n_{kv^\prime\neg{dj}}+\beta_{v^\prime})}} \cdot (n_{dk\neg{dj}}+\alpha_k)$ 
        \EndFor 
     \State $p_{dj,\mathbf{:}} := p_{dj,\mathbf{:}} \mbox{ } ./ \mbox{ sum}(p_{dj,\mathbf{:}})$  \Comment{\tiny{Normalize such that $\sum_{k=1}^Kp_{djk} = 1$}.} \normalsize
     
     \State $\hat{\phi}^p_{\mathbf{:},v} := \hat{\phi}^p_{\mathbf{:},v} + p_{dj,\mathbf{:}}$
\EndFor
\EndFor
\For {$k=1$ to $K$}
	\State $\hat{\phi}^p_{k} := \hat{\phi}^p_{k,\mathbf{:}} \mbox{ } ./ \mbox{ sum}(\hat{\phi}^p_{k,\mathbf{:}}) $\Comment{\tiny{Normalize $\hat{\phi}^p_{k,\mathbf{:}}$ such that $\sum_{v=1}^V \hat{\phi}^p_{kv}=1$}} \normalsize
\EndFor
\State \Return $\hat{\phi}^p$
\EndFunction
\end{algorithmic}

\end{algorithm}

However, in the LDA literature it is much more common practice to simply approximate the posterior (and hence posterior predictive) distribution based on a single estimated value of the parameters (a \emph{point estimate}). Although not ideal from a Bayesian perspective, a point estimate $\hat{\phi}$, $\hat{\theta}$ of the parameters is convenient to work with as it is much easier for a human analyst to interpret, and is computationally much cheaper to use at test time when making predictions than using a collection of samples. In this paper, while strongly agreeing with the posterior predictive-based approach of \citet{nguyen2014sometimes} in principle, we take the perspective that a point estimate is in many cases useful and desirable to have.

Following \citet{nguyen2014sometimes}, we do, however, have substantial reservations regarding the ubiquitous standard practice of using \emph{a single MCMC sample} to obtain a point estimate. By interpreting MCMC as a ``stochastic mode-finding algorithm'' we can view this procedure as a poor-man's approximation to the mode of the posterior.  For many models, with a sufficiently large amount of data the posterior will approach a multivariate Gaussian under the Bernstein-von Mises theorem, and eventually become concentrated at a single point under certain general regularity conditions, cf. \citep{kass1990validity, kleijn2012bernstein}. In this regime, a point estimate based on a single sample will in fact suffice. More generally, however, due to posterior uncertainty this procedure is unstable and consequently statistically inefficient: a parameter estimate from a posterior sample has an \emph{asymptotic relative efficiency (ARE) of 2}, meaning, roughly speaking, that the variance of the estimator requires twice as many observations to match the variance of the ideal estimator, in the asymptotic Gaussian regime, under general regularity conditions, cf. \citep{wang2015privacy}.

As a compromise between the expensive full Bayesian posterior estimation procedure and noisy ``stochastic mode-finding'' with one sample, in this paper we propose to use the \emph{posterior mean}, approximated via multiple MCMC samples, as an estimator of choice in many cases. The posterior mean, if it can be precisely computed, is asymptotically efficient (i.e. it has an ARE of 1), under the conditions necessary for the Bernstein-von Mises theorem to hold, cf. \citep{kleijn2012bernstein}. Although we cannot compute the posterior mean exactly, and the conditions necessary for the Bernstein-von Mises theorem have not been rigorously verified for LDA, these results provide a useful intuition for why we should prefer this approach over an estimator based on a single sample. We expect a sample-based approximation to the posterior mean to be much more stable than an estimator based on a single sample, and our empirical results below will show that this is indeed the case in practice. In our proposed algorithm, we will show how to effectively average over multiple samples almost ``for free'', based on samples and their Gibbs sampling transition probabilities.  This is valuable in the very common setting where we are able to afford a moderate number of burn-in MCMC iterations, but we are not able to afford the much larger number of iterations to compute enough samples required to accurately estimate the posterior mean of the parameters.

It should be noted that averaging over MCMC samples for LDA, without care, can be invalidated due to a lack of \emph{identifiability}.  Multiple samples can each potentially encode different permutations of the topics, thereby destroying any correspondences between them, and \citet{griffiths_steyvers04} strongly warn against this.  However, in this work, we will identify certain cases for which it is safe, and indeed highly beneficial, to average over samples in order to construct a point estimate, such as when estimating $\theta$ at test time with the topics held fixed.

\subsection{Collapsed Gibbs Sampling for LDA}
\label{sec:CGS}

The Collapsed Gibbs Sampling (CGS) algorithm for LDA, introduced by \citet{griffiths_steyvers04}, marginalizes out the parameters $\phi$ and $\theta$, and operates only on the latent variable assignments $\mathbf{z}$.  This leads to a simple but effective MCMC algorithm which mixes much more quickly than the naive Gibbs sampling algorithm.  To compute the CGS updates efficiently, the algorithm maintains and makes use of several count matrices during sampling. We will employ the following notation for these count matrices: $n_{kv}$ represents the number of times that word type $v$ is assigned to topic $k$ across the corpus, and $n_{dk}$ represents the number of word tokens in document $d$ that have been assigned to topic $k$. The notation for these count matrices is included in Table \ref{tbl:notation}.

During sampling, CGS updates the hard assignment $z_i$ of a word token $w_i$ to one of the topics $k \in \{1...K\}$. This update is performed sequentially for all word tokens in the corpus, for a fixed number of iterations. The update equation giving the probability of setting $z_{i}$ to topic $k$, conditional on $w_i$, $d$, the hyperparameters $\boldsymbol{\alpha}$ and $\boldsymbol{\beta}$, and the current topic assignments of all other word tokens (represented by $\cdot$) is: 

\begin{equation} 
p(z_i = k|\, w_i = v,\, d,\, \boldsymbol{\alpha},\, \boldsymbol{\beta},\, \cdot) \propto \frac{n_{kv\neg{i}}+\beta_v}{\sum\limits_{v^\prime = 1}^V({n_{kv^\prime\neg{i}}+\beta_{v^\prime})}} \cdot \frac{n_{dk\neg{i}}+\alpha_k}{N_d+\sum\limits_{k^\prime = 1}^K\alpha_{k^\prime}} \mbox{ .}
\label{eq:p}
\end{equation}

\noindent In the above equation, one excludes from all count matrices the current topic assignment of $w_i$, as indicated by the $\neg{i}$ notation. A common practice is to run the Gibbs sampler for a number of iterations before retrieving estimates for the parameters, as the Markov chain is typically in a low probability state initially and so early estimates are expected to be of poor quality. After this \textit{burn-in} period, we can retrieve estimates for both the topic assignments $z_i$ as well as for the $\theta$ and $\phi$ parameters. 
Samples are taken at a regular interval, often called \textit{sampling lag}.\footnote{Averaging over dependent samples does not introduce any bias. Using a sampling lag (a.k.a. ``thinning'') is merely convenient for reducing memory requirements and computation at test time.}  

To compute point estimates for $\theta$ and $\phi$, Rao-Blackwell estimators were employed in \citep{griffiths_steyvers04}. Specifically, a point estimate of the probability of word type $v$ given topic $k$ is computed as:
\begin{equation} 
\hat{\phi}_{kv} = \frac{n_{kv}+\beta_v}{\sum\limits_{v^\prime = 1}^V({n_{kv^\prime }+\beta_{v^\prime})}} \mbox{ .} 
\label{eq:phi}
\end{equation}

\noindent Similarly, a point estimate for the probability of the topic $k$ given document $d$ is given by:
\begin{equation} 
\hat{\theta}_{dk} = \frac{n_{dk}+\alpha_k}{N_d+ \sum\limits_{k^\prime = 1}^K\alpha_{k^\prime}} \mbox{ .}
\label{eq:theta}
\end{equation}

During prediction (i.e. when applying CGS on documents that were unobserved during training), a common practice is to fix the $\phi$ distributions and set them to the ones learned during estimation. The sampling update equation presented in Equation \ref{eq:p} thus becomes:
\begin{equation} 
p(z_i=k|\, w_i = v,\, d,\, \boldsymbol{\alpha},\, \hat{\phi},\, \cdot) \propto \hat{\phi}_{kv} \cdot \frac{n_{dk\neg{i}}+\alpha_k}{N_d+ \sum\limits_{k^\prime = 1}^K\alpha_{k^\prime}} \mbox{ .}
\label{eq:ppred}
\end{equation}

\subsection{Labeled LDA}
\label{sec:llda}
An extension to unsupervised LDA for supervised multi-label document classification was proposed by \citet{Ramage:2009:LLS:1699510.1699543}. Their algorithm, Labeled LDA (LLDA), employs a one-to-one correspondence between topics and labels. During estimation of the model, the possible assignments for a word token to a topic are constrained to the training document's observed labels. Therefore, during training, the sampling update of Equation \ref{eq:p}, 
becomes:
\begin{equation} 
p(z_i = k|\, w_i = v,\, d,\, \boldsymbol{\alpha},\, \boldsymbol{\beta},\, \cdot) \propto 
    \begin{cases}
      \frac{n_{kv\neg{i}}+\beta_v}{\sum\limits_{v^\prime = 1}^V({n_{kv^\prime \neg{i}}+\beta_{v^\prime})}} \cdot \frac{n_{dk\neg{i}}+\alpha_k}{N_d+\sum\limits_{k^\prime = 1}^K\alpha_{k^\prime}},& \text{if}\ k \in \mathcal{K}_d \\
      0,& \text{otherwise.}
    \end{cases}
\label{eq:pllda}
\end{equation}

Inference on test documents is performed similarly to standard LDA; estimates of the label--word types distributions, $\phi$, are learned on the training data and are fixed, and then the test documents' $\theta$ distributions are estimated using Equation \ref{eq:theta}. Unlike in unsupervised LDA, where topics can change from iteration to iteration (the label switching problem), in LLDA topics remain steady (anchored to a label) and therefore it is possible to average point estimates of $\phi$ and $\theta$ over multiple Markov chains, thereby improving performance. 

\cite{Ramage:2011:PLT:2020408.2020481} have introduced PLDA to relax the constraints of LLDA and exploit the unsupervised and supervised forms of LDA simultaneously. Their algorithm attempts to model hidden topics within each label, as well as unlabeled, corpus-wide latent topics.

\cite{Rubin:2012:STM:2339279.2339301} presented two extensions of the LLDA model. The first one, Prior-LDA, takes into account the label frequencies in the corpus via an informative Dirichlet prior over parameter $\theta$ (we describe this process in detail in Section \ref{sec:priorldasetup}). The second extension, Dependency LDA, takes into account label dependencies by following a two stage approach: first, an unsupervised LDA model is trained on the observed labels. The estimated $\theta^\prime$ parameters incorporate information about the label dependencies. Second, the LLDA model is trained as described in the previous paragraph. During prediction, the previously estimated $\theta^\prime$ parameters of the unsupervised LDA model are used to calculate an asymmetrical hyperparameter $\alpha_{dk}$, which is in turn used to compute the $\theta$ parameters of the LLDA model. 

\subsection{Collapsed Variational Bayes with Zero Order Information -- CVB0}
\label{sec:cvb0}

Collapsed Variational Bayesian (CVB) inference \citep{DBLP:conf/nips/TehNW06} is a deterministic inference technique for LDA. The key idea is to improve on Variational Bayes (VB) by marginalizing out $\theta$ and $\phi$ as in a collapsed Gibbs sampler. The key parameters in CVB are the $\gamma_{di}$ variational distributions over the topics, with $\gamma_{dik}$ representing the probability of the assignment of topic $k$ to the word token $w_i$ in document $d$, under the variational distribution. As the exact implementation of CVB is computationally expensive, the authors propose a second-order Taylor expansion as an approximation to compute the parameters of interest, which nevertheless improves over VB inference with respect to prediction performance.

\cite{Asuncion:2009:SIT:1795114.1795118} presented a further approximation for CVB, by using a zero order Taylor expansion approximation.\footnote{The first-order terms are zero, so this is actually equivalent to a first-order Taylor approximation.} The update equation for CVB0 is given by:

\begin{equation}
\label{eq:cvb0}
\gamma_{dik} \propto \frac{n^{\prime}_{kw_i\neg{i}}+\beta_{w_i}}{\sum\limits_{v^\prime = 1}^V(n^{\prime}_{kv^\prime\neg{i}}+\beta_{v^\prime})}(n^{\prime}_{dk\neg{i}}+\alpha_k)
\end{equation}
with $n^{\prime}_{kv} = \sum\limits_{d =1}^D\sum_{j:w_{dj}=v}\gamma_{djk}$, $n^{\prime}_{dk} = \sum_{j=1}^{N_d}\gamma_{djk}$.

Point estimates for $\theta$ and $\phi$ are retrieved from the same equations as in CGS (Equations \ref{eq:phi} and \ref{eq:theta}). Even though the above update equation looks very similar to the one for CGS, the two algorithms have a couple of key differences. First, the counts $n^{\prime}_{kv}$ and $n^\prime_{dk}$ for CVB0 differ from the respective ones for CGS; the former are summations over the variational probabilities while the latter sum over the topic assignments $z_i$ of topics to words. Second, CVB0 differs from CGS in that in every pass, instead of probabilistically sampling a hard topic-assignment for every token based on the sampling distribution in Equation \ref{eq:p}, it keeps (and updates) that probability distribution for  every word token. A consequence of this procedure is that CVB0 is a deterministic algorithm, whereas CGS is a stochastic algorithm \citep{asuncion2010approximate}. The deterministic nature of CVB0 allows it to converge faster than other inference techniques.

On the other hand, the CVB0 algorithm has significantly larger memory requirements as it needs to store the variational distribution $\gamma$ for every token in the corpus.  More recently, a stochastic extension of CVB0, SCVB0, was presented by \citet{Foulds:2013:SCV:2487575.2487697}, inspired by the Stochastic Variational Bayes (SVB) algorithm of \citet{NIPS2010_3902}. Both SCVB0 and SVB focus on efficient and fast inference of the LDA parameters in massive-scale data scenarios. SCVB0 also improves the memory requirements of CVB0.  Since CVB0 maintains full, dense probability distributions for each word token, it is unable to leverage sparsity to accelerate the algorithm, unlike CGS \citep{yao2009efficient, li2014reducing, yuan2015lightlda, chen2016warplda}.  Inspired by CVB0, in our work we aim to leverage the full distributional uncertainty information per word token, in the context of the parameter estimation step of the CGS algorithm, while maintaining the valuable sparsity properties of that algorithm during the training process.

\section{CGS$_p$ for Document-Topic Parameter Estimation}
\label{sec:cgsp}

In this section we present our new estimator for the document-topic ($\theta_d$) 
parameters of LDA that make use of the full distributions of word tokens over topics. For simplicity, we first describe the case of $\theta_d$ estimation during prediction and then extend our theory to $\theta_d$ estimation during training. We present our estimator for the topic-word parameters ($\phi_k$) in the following section.

\subsection{Standard CGS $\theta_d$ Estimator}
The standard estimator for document $d$'s parameters $\theta_d$ from a collapsed Gibbs sample $\mathbf{z}_d$ (Equation \ref{eq:theta}), due to \citet{griffiths_steyvers04}, corresponds to the posterior predictive distribution over topic assignments, i.e. 

\begin{align}
\hat{\theta}_{dk} &\triangleq p(z^{(new)}=k|\mathbf{z}_d, \boldsymbol{\alpha})\\
&=\int p(z^{(new)}=k, \theta_d|\mathbf{z}_d, \boldsymbol{\alpha})\ d \theta_d \\
&=\int p(z^{(new)}=k|\theta_d)  p(\theta_d|\mathbf{z}_d, \boldsymbol{\alpha})\ d \theta_d\\
&=\int \theta_{dk}  p(\theta_d|\mathbf{z}_d, \boldsymbol{\alpha})\ d \theta_d\\
&= \mathbb{E}_{p(\theta_d|\mathbf{z}_d, \boldsymbol{\alpha})}[\theta_{dk}] \label{eqn:gs04PosteriorMean}\\
&= \mathbb{E}_{\mbox{Dirichlet}(\theta_d|\boldsymbol{\alpha} + \mathbf{n}_{d})}[\theta_{dk}] \\
&= \frac{n_{dk} + \alpha_k}{N_d + \sum_{k'=1}^K\alpha_{k'}} \mbox{ ,} \label{eqn:gs04Estimator}
\end{align}
where the last two lines follow from Dirichlet/multinomial conjugacy, and from the mean of a Dirichlet distribution, respectively.  Here, $z^{(new)}$ corresponds to a hypothetical ``new'' word in the document. The posterior predictive distribution can be understood via the urn process interpretation of collapsed Dirichlet-multinomial models.  After marginalizing out the parameters $\theta$ of a multinomial distribution  $\mathbf{x} \sim \mbox{Multinomial}(\theta, N)$ with a Dirichlet prior  $\theta \sim \mbox{Dirichlet}(\boldsymbol{\alpha})$, we arrive at an urn process called the Dirichlet-multinomial distribution, a.k.a. the multivariate Polya urn, cf. \citep{Minka2000estimating}.  The urn process is as follows:
\begin{samepage}
\begin{itemize}
    \item Begin with an empty urn
  	\item For each $k$, $1 \leq k \leq K$
  		\begin{itemize}
  		  \item add $\alpha_k$ balls of color $k$ to the urn
  		\end{itemize}
  	\item For each $i$, $1 \leq i \leq N$
	\begin{itemize}
	  \item Reach into the urn and draw a ball uniformly at random
	  \item Observe its color, $k$.  Count it, i.e. add one to $x_k$
      \item Place the ball back in the urn, along with a \emph{new ball of the same color}.
	\end{itemize}
\end{itemize}
\end{samepage}
We can interpret the posterior predictive distribution above as the probability of the next ball in the urn, i.e. the next word in the document, if we were to add one more word, by reaching into the urn once more.  

\subsection{A Marginalized Estimation Principle}
The topic assignment vector $\mathbf{z}_d$ for test document $d$ is a latent variable which typically has quite substantial posterior uncertainty associated with it, and yet the most common practice is to ignore this uncertainty and construct a point estimate of $\theta_d$ from a single $\mathbf{z}_d$ sample, which can be detrimental to performance \citep{nguyen2014sometimes}. Following \citet{griffiths_steyvers04}, we assume that the predictive probability of a new word's topic assignment $z^{(new)}$ is a principled estimate of $\theta_d$.  Differently to previous work, though, we take the perspective that the latent $\mathbf{z}_d$ is a nuisance variable which should be marginalized out. This leads to a \emph{marginalized} version of Griffiths and Steyvers' estimator,
\begin{equation}
\bar{\theta}_{dk} \triangleq p(z^{(new)}=k|\mathbf{w}_d, \phi, \boldsymbol{\alpha}) = \sum_{\mathbf{z}_d} p(z^{(new)}=k, \mathbf{z}_d|\mathbf{w}_d, \phi, \boldsymbol{\alpha}) \mbox{ .} 
\label{eqn:marginalizedEstimator}
\end{equation}
Due to its treatment of uncertainty in $\mathbf{z}_d$, we advocate for $\bar{\theta}_{d}$ as a gold standard principle for constructing a point estimate of $\theta_d$. We will introduce computationally efficient Monte Carlo algorithms for approximating it below. Note that while previous works have considered averaging strategies which correspond to Monte Carlo approximations of this principle, including \citet{griffiths_steyvers04}, \citet{Asuncion:2009:SIT:1795114.1795118}, \citet{nguyen2014sometimes}, and references therein, discussion on the estimation principle of Equation \ref{eqn:marginalizedEstimator} that these methods approximate appears to be missing in the literature. \citet{griffiths_steyvers04} mention that such averaging strategies are possible but caution against them due to concerns regarding identifiability, however this identifiability issue does not arise at test time with topics $\phi$ held fixed, as in the scenario we consider here. \citet{nguyen2014sometimes} empirically explore averaging procedures for making predictions by approximating the posterior predictive distribution. They do not aim to compute a point estimate from multiple samples, however their \emph{Single Average} prediction strategy can be seen to be equivalent to using a naive Monte Carlo estimate (given in Equation \ref{eqn:naiveMonteCarloMarginalizedEstimator}) of Equation \ref{eqn:marginalizedEstimator} in the context of predicting held-out words.

\subsubsection{Interpretations}
We can interpret the estimator from Equation \ref{eqn:marginalizedEstimator} in several other ways.
\paragraph{Expected Griffiths and Steyvers estimator:}
The estimator $\bar{\theta}_{d}$ can readily be seen to be the posterior mean of the Griffiths and Steyvers estimator $\hat{\theta}_{d}$:
\begin{align}
\bar{\theta}_{dk} = p(z^{(new)}=k|\mathbf{w}_d, \phi, \boldsymbol{\alpha}) &= \sum_{\mathbf{z}_d} p(z^{(new)}=k, \mathbf{z}_d|\mathbf{w}_d, \phi, \boldsymbol{\alpha})\\
&= \sum_{\mathbf{z}_d} p(z^{(new)}=k|\mathbf{z}_d, \boldsymbol{\alpha}) p(\mathbf{z}_d|\mathbf{w}_d, \phi, \boldsymbol{\alpha}) \label{eqn:expectedGS}\\
&= E_{p(\mathbf{z}_d|\mathbf{w}_d, \phi, \boldsymbol{\alpha})} [p(z^{(new)}=k|\mathbf{z}_d, \boldsymbol{\alpha})]\\
&= E_{p(\mathbf{z}_d|\mathbf{w}_d, \phi, \boldsymbol{\alpha})}[\hat{\theta}_{dk}] \mbox{ .}
\end{align}

\paragraph{Posterior mean of $\theta_d$:}
Griffiths and Steyvers' estimator can be viewed as the posterior mean of $\theta_d$, given $\mathbf{z}_d$ (Equation \ref{eqn:gs04PosteriorMean}).  Along these lines, we can interpret $\bar{\theta}_{d}$ as the marginal posterior mean:
\begin{align}
\bar{\theta}_{dk} = p(z^{(new)}=k|\mathbf{w}_d, \phi, \boldsymbol{\alpha}) &= E_{p(\mathbf{z}_d|\mathbf{w}_d, \phi, \boldsymbol{\alpha})}[\hat{\theta}_{dk}] \\
&= E_{p(\mathbf{z}_d|\mathbf{w}_d, \phi, \boldsymbol{\alpha})} \Big [\mathbb{E}_{p(\theta_d|\mathbf{z}_d, \boldsymbol{\alpha})}[\theta_{dk}] \Big ]\\
&= E_{p(\theta_d|\mathbf{w}_d, \phi, \boldsymbol{\alpha})} [\theta_{dk}] \mbox{ , }
\end{align}
where the last line follows by the law of total expectation.

\paragraph{Urn model interpretation:}
To interpret the estimator from an urn model perspective, we can plug in Equation \ref{eqn:gs04Estimator}: 
\begin{align}
\bar{\theta}_{dk} = p(z^{(new)}=k|\mathbf{w}_d, \phi, \boldsymbol{\alpha}) 
&= \sum_{\mathbf{z}_d} p(z^{(new)}=k|\mathbf{z}_d, \boldsymbol{\alpha}) p(\mathbf{z}_d|\mathbf{w}_d, \phi, \boldsymbol{\alpha}) \label{eqn:mixtureOfUrnProcesses}\\
&= \sum_{\mathbf{z}_d} \frac{n_{dk}+ \alpha_k}{N_d + \sum_{k'=1}^K\alpha_{k'}} p(\mathbf{z}_d|\mathbf{w}_d, \phi, \boldsymbol{\alpha}) \mbox{ . }
\end{align}
  We can view this as a mixture of urn models.  It can be simulated by picking an urn with colored  balls in it corresponding to the count vector $\mathbf{n}_{d} + \boldsymbol{\alpha}$, with probability according to the posterior $p(\mathbf{z}_{d}|\mathbf{w}_d, \phi, \boldsymbol{\alpha})$, then selecting a colored ball, corresponding to a topic assignment, from this urn uniformly at random.  Alternatively, we can model this equation with a single urn, with (fractional) colored balls placed in it according to the count vector,
\begin{align}
\bar{\theta}_{dk} \propto \sum_{\mathbf{z}_d} (n_{dk} + \alpha_k)p(\mathbf{z}_d|\mathbf{w}_d, \phi, \boldsymbol{\alpha}) &= \sum_{\mathbf{z}_d} p(\mathbf{z}_d|\mathbf{w}_d, \phi, \boldsymbol{\alpha})n_{dk} +  \alpha_k \mbox{ , } \label{eqn:oneUrnInterpretation}
\end{align}
and with the next ball (i.e. topic) $z^{(new)}$ selected according to the total number of balls of each color (i.e. topic) in the urn.

\subsubsection{Monte Carlo algorithms for computing the estimator}
Due to the intractable sum over $\mathbf{z}_d$, we cannot compute $\bar{\theta}_{d}$ exactly.  A straightforward Monte Carlo estimate of $\bar{\theta}_{d}$ is given by
\begin{align}
\bar{\theta}_{dk} = E_{p(\mathbf{z}_d|\mathbf{w}_d, \phi, \boldsymbol{\alpha})} [p(z^{(new)}=k|\mathbf{z}_d, \boldsymbol{\alpha})] &= E_{p(\mathbf{z}_{d}|\mathbf{w}_d, \phi, \boldsymbol{\alpha})} \Big [\frac{n_{dk} + \alpha_k}{N_d + \sum_{k'=1}^K\alpha_{k'} }\Big ] \label{eqn:expectedCounts}\\
&\approx \frac{1}{S} \sum_{i=1}^S \frac{n^{(i)}_{dk} + \alpha_k}{N_d + \sum_{k'=1}^K\alpha_{k'}} \mbox{ , } \label{eqn:naiveMonteCarloMarginalizedEstimator}
\end{align}
where each of the count variables $\mathbf{n}^{(i)}_d$ corresponds to a sample $\mathbf{z}_d^{(i)} \sim p(\mathbf{z}_d|\mathbf{w}_d, \phi, \boldsymbol{\alpha})$.  Algorithmically, this corresponds to drawing multiple MCMC samples of $\mathbf{z}_d$ and averaging the Griffiths and Steyvers estimator $\hat{\theta}_{d}$ over them.  Alternatively, by linearity of expectation we can shift the expectation inwards to rewrite the above as
\begin{align}
E_{p(\mathbf{z}_{d}|\mathbf{w}_d, \phi, \boldsymbol{\alpha})} \Big [\frac{n_{dk} + \alpha_k }{N_d + \sum_{k'=1}^K\alpha_{k'} }\Big ] &= E_{p(\mathbf{z}_{d}|\mathbf{w}_d, \phi, \boldsymbol{\alpha})} \Big [\frac{\sum_{j=1}^{N_d} z_{djk} + \alpha_k }{N_d + \sum_{k'=1}^K\alpha_{k'} }\Big ]\\
&= \frac{\sum_{j=1}^{N_d} E_{p(\mathbf{z}_{d}|\mathbf{w}_d, \phi, \boldsymbol{\alpha})}[z_{djk}] + \alpha_k}{N_d + \sum_{k'=1}^K\alpha_{k'} } \mbox{ .} \label{eqn:expectationInwardsTheta}
\end{align}
This leads to another Monte Carlo estimate,
\begin{align}
\bar{\theta}_{dk} \approx \frac{\sum_{j=1}^{N_d} \frac{1}{S}\sum_{i=1}^Sz^{(i)}_{djk} + \alpha_k }{N_d + \sum_{k'=1}^K\alpha_{k'}} \mbox{ .} \label{eqn:MonteCarloPerWord}
\end{align}

In this case we would once again draw multiple MCMC samples of $\mathbf{z}_d$, but average over the samples to compute the proportion of times that each word is assigned to each topic, applying Griffiths and Steyvers' estimator to the resulting expected total counts.  It can straightforwardly be shown that this estimator is mathematically equivalent to the estimator of Equation \ref{eqn:naiveMonteCarloMarginalizedEstimator}, so this algorithm will perform identically to that method.  We mention this formulation for expository purposes, as it will be used as a launching point for deriving our proposed CGS$_p$ algorithm.

\subsection{CGS$_p$ for $\theta_d$}
\label{sec:motivatingCGS_p}
We aim to design a statistically and computationally efficient Monte Carlo algorithm which improves on the above algorithms.
To do this, we first consider a hypothetical Monte Carlo algorithm which is more expensive but easily understood, using Equation \ref{eqn:MonteCarloPerWord} as a starting point.  We then propose our algorithm, CGS$_p$, as a more computationally efficient version of this hypothetical method.
Let $\mathbf{z}_d^{(i)} \sim p(\mathbf{z}_d|\mathbf{w}_d, \phi, \boldsymbol{\alpha})$,  $i \in \{1,\ldots,S \}$ be $S$ samples from the posterior for the topic assignments, and let $\mathbf{z}_d^{(i,l)\rightarrow j}$, $l \in \{1,\ldots, L\}$ be $L$ Gibbs samples starting from $\mathbf{z}_d^{(i)}$, where a Gibbs update to the $j$th topic assignment is performed.  The Bayesian posterior distribution is a stationary distribution of the Gibbs sampler, so these Gibbs updated samples are still samples from the posterior, i.e. $\mathbf{z}_d^{(i,l)\rightarrow j} \sim p(\mathbf{z}_d|\mathbf{w}_d, \phi, \boldsymbol{\alpha})$.  We can therefore estimate posterior expectations of any quantity of interest using these samples instead of the $\mathbf{z}_d^{(i)}$'s, and in particular:
\begin{align}
\bar{\theta}_{dk} \approx \frac{\sum_{j=1}^{N_d} \frac{1}{SL}\sum_{i=1}^S\sum_{l=1}^Lz^{(i,l)\rightarrow j}_{djk} + \alpha_k }{N_d + \sum_{k'=1}^K\alpha_{k'}} \mbox{ .} \label{eqn:MonteCarloPerWord_OneStep}
\end{align}
This method increases the sample size, and also the effective sample size, relative to the method of Equation \ref{eqn:MonteCarloPerWord}, but also greatly increases the computational effort to obtain the Monte Carlo estimator, making it relatively impractical.  Fortunately, a variant of this procedure exists which is both computationally faster and has lower variance.  Let $T(\mathbf{z}_{d}^{(i)\rightarrow j}|\mathbf{z}_d^{(i)}, \phi)$ be the transition operator for a Gibbs update on the $j$th topic assignment.  Holding the other assignments $\mathbf{z}_{d\neg j}^{(i)}$ fixed, we have the partially collapsed Gibbs update
\begin{equation}
T(\mathbf{z}_{djk}^{(i)\rightarrow j} = 1|\mathbf{z}_d^{(i)}, \phi) \propto \phi_{k w_{d,j}} \frac{n^{(i)}_{dk\neg j} + \alpha_k }{N_{d\neg j} + \sum_{k'=1}^K\alpha_{k'}} \mbox{ .} 
\end{equation}
By the strong law of large numbers,
\begin{align}
\frac{1}{L}\sum_{l=1}^L z^{(i,l)\rightarrow j}_{djk} \rightarrow E_{T(\mathbf{z}_d^{(i)\rightarrow j}|\mathbf{z}_d^{(i)}, \phi)}[z_{djk}^{(i)\rightarrow j}] = T(z_{djk}^{(i)\rightarrow j}|\mathbf{z}_d^{(i)}, \phi)\mbox{ as } L \rightarrow \infty \mbox{  with probability 1.}
\end{align}
So for the cost of a single Gibbs sweep through the document, we can obtain an effective number of inner loop samples $L = \infty$ by plugging in the Gibbs sampling probabilities as the expected topic counts in Equation \ref{eqn:MonteCarloPerWord_OneStep}.  This leads to our proposed estimator for $\theta_{dk}$: 
\begin{align}
 \hat{\theta}^{p}_{dk} \triangleq & \frac{\sum_{j=1}^{N_d} \frac{1}{S}\sum_{i=1}^S T(\mathbf{z}_{djk}^{(i)\rightarrow j}|\mathbf{z}_d^{(i)}, \phi) + \alpha_k}{ N_d + \sum_{k'=1}^K\alpha_{k'}} \mbox{ .}
\label{eq:thetap}
\end{align}
This method is equivalent to averaging over an infinite number of Gibbs samples adjacent to our initial samples $i = 1, \ldots, S$ in order to estimate each word's topic assignment probabilities, and thereby estimate the expected topic counts for the document, which correspond to our estimate of $\theta_d$. 
By the Rao-Blackwell theorem, it has lower variance than the method in Equation \ref{eqn:MonteCarloPerWord_OneStep} with finite $L$.  We refer to this Monte Carlo procedure as CGS$_p$, for collapsed Gibbs sampling with probabilities maintained.

The above argument makes it clear that $\hat{\theta}^{p}_{d}$ is an unbiased Monte Carlo estimator of the posterior predictive distribution, but for added rigor we can also show this directly:

\begin{align}
 \hat{\theta}^{p}_{dk} &\propto \sum_{j=1}^{N_d} \frac{1}{S}\sum_{i=1}^S T(z_{djk}^{(i)\rightarrow j}|\mathbf{z}_d^{(i)}, \phi) + \alpha_k \\
 &= \sum_{j=1}^{N_d} \frac{1}{S}\sum_{i=1}^S E_{T(\mathbf{z}_d^{(i)\rightarrow j}|\mathbf{z}_d^{(i)}, \phi)}[z_{djk}^{(i)\rightarrow j}] + \alpha_k \\
 &\approx \sum_{j=1}^{N_d} E_{p(\mathbf{z}_d|\mathbf{w}_d, \phi, \boldsymbol{\alpha})} \Big [E_{T(\mathbf{z}_d^{\rightarrow j}|\mathbf{z}_d, \phi)}[z_{djk}^{\rightarrow j}] \Big ] + \alpha_k \\
 &= \sum_{j=1}^{N_d} E_{p(\mathbf{z}_d^{\rightarrow j }|\mathbf{w}_d, \phi, \boldsymbol{\alpha})}[ z_{djk}^{\rightarrow j}] + \alpha_k \tag*{\mbox{ (by the law of total expectation)}}\\
 &= \sum_{j=1}^{N_d} E_{p(\mathbf{z}_d|\mathbf{w}_d, \phi, \boldsymbol{\alpha})}[z_{djk}] + \alpha_k \tag*{ \mbox{  (by stationarity of the Gibbs sampler)}} \\
 &= E_{p(\mathbf{z}_d|\mathbf{w}_d, \phi, \boldsymbol{\alpha})}[\sum_{j=1}^{N_d} z_{djk}] + \alpha_k\\
 &= E_{p(\mathbf{z}_d|\mathbf{w}_d, \phi, \boldsymbol{\alpha})}[n_{dk} + \alpha_k]\\
 &\propto E_{p(\mathbf{z}_d|\mathbf{w}_d, \phi, \boldsymbol{\alpha})}[\frac{n_{dk} + \alpha_k}{N_d + \sum_{k' = 1}^K\alpha_{k'}}]\\
 &= p(z^{(new)}=k|\mathbf{w}_d, \phi, \boldsymbol{\alpha}) \mbox{ .} \tag*{\mbox{ (by Equation \ref{eqn:expectedCounts})}}
\end{align}
We advocate the use of this procedure even when we can only afford a single burned-in sample, i.e. $S=1$, as a cheap approximation to $p(z^{(new)}=k|\mathbf{w}_d, \phi, \boldsymbol{\alpha})$ which can be used as a plug-in replacement to \citet{griffiths_steyvers04}'s single sample estimator $p(z^{(new)}=k|\mathbf{z}_d, \boldsymbol{\alpha})$.  Indeed, the increased stability of our estimator, due to implicit averaging, is expected to be especially valuable when $S=1$.  We emphasize three primary observations arising from the above analysis.
\begin{enumerate}
\renewcommand*\labelenumi{(\theenumi)}
\item We have shown that it is valid and potentially beneficial to use ``soft'' probabilistic counts instead of hard assignment counts when computing parameter estimates from collapsed Gibbs samples.  This ports the soft clustering methodology of CVB0 to the CGS setting at estimation time. While our approach computes soft counts via Gibbs sampling transition probabilities and CVB0 uses variational distributions, the corresponding equations to compute them are very similar, cf. Equations \ref{eq:p}, \ref{eq:ppred}, and \ref{eq:cvb0}.
\item Averaging these (hard or soft) count matrices over multiple samples can be performed to improve the stability of the estimators, as long as identifiability is resolved, e.g. at test time with the topics held fixed.
\item Averaging over either hard or soft count matrices will converge to the same solution in the long run: the marginalized version of the Griffiths and Steyvers estimator, given by Equation \ref{eqn:marginalizedEstimator}.  However, we expect the use of soft counts to be much more efficient in the number of samples $S$, as it implicitly averages over larger numbers of samples.
\end{enumerate}

This last point is crucial in many common practical applications of topic modeling where we have a limited computational budget per document at prediction time, and can afford only a modest burn-in period, and few samples after burn-in.  This situation frequently arises when evaluating topic models, for which a novel approach is typically compared to multiple baseline methods on a test set containing up to tens of thousands of documents, with the entire experiment being repeated across multiple corpora and/or training folds.  The computational challenge is greatly exacerbated when evaluating topic model training algorithms, for which the above must be repeated at many stages of the training process \citep{foulds2014Annealing}.  For instance, \citet{Asuncion:2009:SIT:1795114.1795118} state that ``\emph{in our experiments we don't perform averaging over samples
for CGS \ldots [in part] for computational reasons}.''  Computational efficiency at test time is also essential when using topic modeling for multi-label document classification \citep{Ramage:2009:LLS:1699510.1699543,Ramage:2011:PLT:2020408.2020481,Rubin:2012:STM:2339279.2339301}, for which a deployed system may need to perform inference on a never-ending stream of incoming test documents.

Pseudocode illustrating the proposed method 
is provided in 
Algorithm \ref{alg:methods}
.  It should be noted than any existing CGS implementation can very straightforwardly be modified to apply our proposed technique, including MALLET \citep{McCallumMALLET}, Yahoo\_LDA \citep{smola2010architecture}, and WarpLDA \citep{chen2016warplda}.  Our method is compatible with modern sparse and parallel implementations of CGS, as the final estimation procedure does not need to alter the training process.  This can lead to a ``best of both worlds'' situation, where sparsity is leveraged during the expensive CGS inference process, but the full dense distributions are used during the relatively inexpensive parameter recovery step, which in many topic modeling applications is performed only once.   In this setting, the improved performance is typically worth the computational overhead (which is similar to that of a single dense CGS update), as we demonstrate in Section \ref{sec:overhead}.  A practical implementation of our estimators can further leverage the fact that the count matrices are not modified during the procedure, by computing the sampling distribution only once for each distinct word type in each document, and by performing the computations in parallel. In the following we will sometimes abuse notation and denote the procedure of using $\hat{\theta}^{p}_d$ to estimate the $\theta_d$'s as $\theta^p$ (and similarly for the procedure to estimate $\phi$, denoted $\phi^p$, introduced below).

\subsubsection{CGS$_p$ for estimating $\theta_d$ for training documents}
During the training phase, we can apply the same procedure to estimate $\theta_d$ for training documents based on a given collapsed Gibbs sample $\mathbf{z}^{(i)}$:
\begin{align}
 \hat{\theta}^{p}_{dk} = & \frac{\sum_{j=1}^{N_d} T(z_{djk}^{(i)\rightarrow j}|\mathbf{z}^{(i)}, \phi) + \alpha_k}{ N_d + \sum_{k'=1}^K\alpha_{k'}} \mbox{ .}
\label{eq:thetapEst}
\end{align}
The collapsed Gibbs sampler does not explicitly maintain an estimate of $\phi$ to plug into the above. However, the Griffiths and Steyvers estimator for $\phi$ (Equation \ref{eq:phi}) can be obtained from the same count matrices that CGS computes, as it corresponds almost exactly to the first term on the right hand side of Equation \ref{eq:p}. Alternatively, below we propose a novel estimator for $\phi$ which can also be used here with a little bit more computational effort.

Similarly to the Griffiths and Steyvers estimator, it should be noted that averaging this estimator over multiple samples from different training iterations raises issues with the identifiability of the topics, which can potentially be problematic \citep{griffiths_steyvers04}. 
We do not consider this procedure further here.  It is however safe to apply Equation \ref{eq:thetap} to multiple MCMC samples for $\mathbf{z}_d$ that are generated while holding $\phi$ fixed, i.e. treating document $d$ as a test document for the purposes of estimating $\theta_d$ given the $\hat{\phi}$ from $\mathbf{z}^{(i)}$.

\section{CGS$_p$ for Topic - Word Types Parameter Estimation}
\label{sec:phi_p}
By applying the same approach as for $\theta^p$, plugging in Gibbs transition probabilities instead of indicator variables into Equation \ref{eq:phi}, we arrive at an analogous technique for estimating $\phi$:
\begin{align}
\hat{\phi}^{p}_{kv} = \frac{\sum_{d=1}^{D_{Train}}\sum_{j:w_{d,j} = v} T(z_{djk}^{(i)\rightarrow j}|\mathbf{z}^{(i)}) + \beta_v}{ \sum_{d=1}^{D_{Train}}\sum_{j=1}^{N_d} T(z_{djk}^{(i)\rightarrow j}|\mathbf{z}^{(i)})+  \sum_{v^\prime = 1}^V \beta_{v^\prime}} \mbox{ ,}
\label{eq:phip}
\end{align}
where $T(z_{d,j}^{(i)\rightarrow j}|\mathbf{z}^{(i)})$ corresponds to the collapsed Gibbs update in Equation \ref{eq:p}.  While the intuition of the above method is clear, there are several technical complications to an analogous derivation for it. Nevertheless, we show that the same reasoning essentially holds for $\phi$, with  minor approximations.  In the following, we outline a justification for the $\phi^p$ estimator, with emphasis on the places where the argument from $\theta^p$ does not apply, along with well-principled approximation steps to resolve these differences.

First, consider the estimation principle underlying the technique.  For topics $\phi$, the standard Rao-Blackwellized estimator of \citet{griffiths_steyvers04}, given by Equation \ref{eq:phi}, once again corresponds to the posterior predictive distribution,
\begin{align}
\hat{\phi}_{kv} &\triangleq p(w^{(new)} = v| \mathbf{z},  \mathbf{w}, \boldsymbol{\beta}, z^{(new)} = k)\nonumber \\
&= \int p(w^{(new)} = v, \phi_k| \mathbf{z}, \mathbf{w}, \boldsymbol{\beta}, z^{(new)} = k) d\phi_k \nonumber\\
&= \int p(w^{(new)} = v| \phi_k, z^{(new)} = k) p(\phi_k|\mathbf{w}_{z=k}, \boldsymbol{\beta}) d\phi_k \nonumber\\
&= E_{Dirichlet(\mathbf{n}_k + \boldsymbol{\beta})}[\phi_{kv}] \nonumber\\
&= \frac{n_{kv}+\beta_v}{\sum\limits_{v^\prime = 1}^V({n_{v^\prime k}+\beta_{v^\prime})}} \mbox{ ,}
\end{align}
where $\mathbf{w}_{z=k}$ is the collection of word tokens assigned to topic $k$.
This estimator again conditions on the latent variable assignments $\mathbf{z}$, which are uncertain quantities.  We desire a marginalized estimator, summing out the latent variables $\mathbf{z}$.  However, without conditioning on $\mathbf{z}$, the topic index $k$ loses its meaning because the topics are not identifiable.  The naive marginalized estimator
\begin{equation}
\label{eqn:phiMarginalized}
\bar{\phi}_{kv} \triangleq p(w^{(new)} = v| \mathbf{w}, \boldsymbol{\beta}, z^{(new)} = k) = \sum_{\mathbf{z}}p(w^{(new)} = v, \mathbf{z}|  \mathbf{w}, \boldsymbol{\beta}, z^{(new)} = k)
\end{equation}
sums over all permutations of the (collapsed representation of the) topics, and so conditioning on $z^{(new)} = k$ has no effect. The estimator in Equation \ref{eqn:phiMarginalized} is therefore unfortunately not well defined. On the other hand, a sum over $\mathbf{z}$'s \emph{is} meaningful as long as \emph{the topics are aligned for all values of $\mathbf{z}$ considered}.  Suppose, by some oracle, we know that $\mathbf{z}$ belongs to some \emph{identified subset} of assignments $\dot{Z}$, for which topic indices are aligned between all elements $\mathbf{z}' \in \dot{Z}$. We are then able to properly define our idealized estimator with respect to $\dot{Z}$,
\begin{equation}
\label{eqn:phiMarginalizedIdentified}
\bar{\phi}_{kv}^{(\dot{Z})} \triangleq p(w^{(new)} = v| \mathbf{w}, \boldsymbol{\beta}, z^{(new)} = k, \mathbf{z} \in \dot{Z})  = \sum_{\mathbf{z} \in \dot{Z}}p(w^{(new)} = v, \mathbf{z}|  \mathbf{w}, \boldsymbol{\beta}, z^{(new)} = k, \mathbf{z} \in \dot{Z}) \mbox{ .}
\end{equation}
Regarding our proposed technique, the Monte Carlo estimator in Equation \ref{eq:phip} implicitly averages over the set of possible Gibbs samples $Z_{Gibbs}(\mathbf{z}^{(i)})$ that are adjacent to sample $\mathbf{z}^{(i)}$, i.e. the candidate $\mathbf{z}$ assignments that differ from $\mathbf{z}^{(i)}$ in a single entry.  A difference of a single word is extremely unlikely to cause label switching for a realistic corpus, so the topics will almost certainly be aligned for each of the adjacent Gibbs samples that the estimator implicitly averages over. The empirical results of \citet{nguyen2014sometimes} also support this, as they found that averaging over multiple samples from the same MCMC chain was beneficial to performance, which suggests that identifiability was not a problem even for MCMC samples that are many iterations apart.  Hence, we take $\bar{\phi}_{kv}^{(Z_{Gibbs}(\mathbf{z}^{(i)}))}$ as our idealized estimator, under the supposition that $Z_{Gibbs}(\mathbf{z}^{(i)})$ is identified, and we aim to derive Equation \ref{eq:phip} as a Monte Carlo algorithm that efficiently approximates it.

To accomplish this, following the derivation for $\hat{\theta}^p$, we would like to show that 
\begin{enumerate}
\renewcommand*\labelenumi{(\theenumi)}
\item we can use a Monte Carlo estimator in which the $z_{d,j}$'s can be sampled individually, as in Equation \ref{eqn:MonteCarloPerWord}, and \label{item:individualZs}
\item the transition probabilities in Equation \ref{eq:phip} can be used to implement this estimator with an effectively infinite number of adjacent Gibbs samples, as in Section \ref{sec:motivatingCGS_p}. \label{item:infiniteGibbs}
\end{enumerate}
While the previous stationarity argument for (\ref{item:infiniteGibbs}) still applies, there is a further complication for (\ref{item:individualZs}).
For $\hat{\theta}^{p}$, linearity of expectation was sufficient to show in Equation \ref{eqn:expectationInwardsTheta} that
\begin{align*}
 E_{p(\mathbf{z}_{d}|\mathbf{w}_d, \phi, \boldsymbol{\alpha})} \Big [\frac{\sum_{j=1}^{N_d} z_{djk} + \alpha_k }{N_d + \sum_{k'=1}^K\alpha_{k'} }\Big ] &= \frac{\sum_{j=1}^{N_d} E_{p(\mathbf{z}_{d}|\mathbf{w}_d, \phi, \boldsymbol{\alpha})}[z_{djk}] + \alpha_k}{N_d + \sum_{k'=1}^K\alpha_{k'} } \mbox{ .} 
\end{align*}
We cannot use this argument directly for $\hat{\phi}^p$ because in this case the denominator of the Griffiths and Stevyers estimator, which is averaged over, also includes $\mathbf{z}$ terms which are involved in the expectation:
\begin{align}
\label{eq:phiExpectation}
\bar{\phi}_{kv}^{(Z_{Gibbs}(\mathbf{z}^{(i)}))} &= E_{p(\mathbf{z}| \mathbf{w}, \boldsymbol{\beta}, \mathbf{z} \in Z_{Gibbs}(\mathbf{z}^{(i)}))}\Big [\frac{\sum_{d=1}^{D_{Train}}\sum_{j:w_{d,j} = v} z_{djk} + \beta_v}{\sum_{d=1}^{D_{Train}}\sum_{j=1}^{N_d} z_{djk} +  \sum_{v^\prime = 1}^V\beta_{v^\prime}} \Big ] \mbox{ .}
\end{align}
The issue here is that the adjacent Gibbs samples modify the overall counts per topic $n_{k} \triangleq \sum_{d=1}^{D_{Train}}\sum_{j=1}^{N_d} z_{djk}$.  However, between adjacent Gibbs samples $\mathbf{z}$ and $\mathbf{z}^{(i)}$, the corresponding topic counts $n_k$ are within $\pm1$ of $n^{(i)}_{k}$ and for a typical corpus we have that $n_k >> 1$, so the impact of a single word is negigible, i.e. 
\begin{align}
\nonumber
& \sum_{d=1}^{D_{Train}}\sum_{j=1}^{N_d}z_{djk} \in \Big \{\sum_{d=1}^{D_{Train}}\sum_{j=1}^{N_d}z_{djk}^{(i)}, \sum_{d=1}^{D_{Train}}\sum_{j=1}^{N_d}z_{djk}^{(i)} \pm1 \Big \} \approx \sum_{d=1}^{D_{Train}}\sum_{j=1}^{N_d}z_{djk}^{(i)} \mbox{ ,}\\
& \mbox{ for } \mathbf{z} \in Z_{Gibbs}(\mathbf{z}^{(i)}) \mbox{ , } \sum_{d=1}^{D_{Train}}\sum_{j=1}^{N_d}z_{djk} >> 1 \mbox{ .} \label{eq:approximationphip1}
\end{align}
In this case, we see from Equation \ref{eq:approximationphip1} that for a corpus in which $n_k >> 1$, for all practical purposes the denominator in Equation \ref{eq:phiExpectation} is essentially constant over all adjacent Gibbs samples, and so (\ref{item:individualZs}) still holds approximately, and we have
\begin{align}
\label{eq:approximationphip2}
\bar{\phi}_{kv}^{(Z_{Gibbs}(\mathbf{z}^{(i)}))} &= E_{p(\mathbf{z}| \mathbf{w}, \boldsymbol{\beta}, \mathbf{z} \in Z_{Gibbs}(\mathbf{z}^{(i)}))}\Big [\frac{\sum_{d=1}^{D_{Train}}\sum_{j:w_{d,j} = v} z_{djk} + \beta_v}{ \sum_{d=1}^{D_{Train}}\sum_{j=1}^{N_d} z_{djk} + \sum_{v^\prime = 1}^V\beta_{v^\prime}} \Big ] \nonumber \\
&\approx E_{p(\mathbf{z}| \mathbf{w}, \boldsymbol{\beta}, \mathbf{z} \in Z_{Gibbs}(\mathbf{z}^{(i)}))}\Big [\frac{\sum_{d=1}^{D_{Train}}\sum_{j:w_{d,j} = v} z_{djk} + \beta_v}{ \sum_{d=1}^{D_{Train}}\sum_{j=1}^{N_d} z_{djk}^{(i)} + \sum_{v^\prime = 1}^V\beta_{v^\prime}} \Big ] \nonumber \\
&= \frac{\sum_{d=1}^{D_{Train}}\sum_{j:w_{d,j} = v} E_{p(\mathbf{z}| \mathbf{w}, \boldsymbol{\beta}, \mathbf{z} \in Z_{Gibbs}(\mathbf{z}^{(i)})) }[z_{djk}] + \beta_v}{ \sum_{d=1}^{D_{Train}}\sum_{j=1}^{N_d} z_{djk}^{(i)} +  \sum_{v^\prime = 1}^V\beta_{v^\prime}} \mbox{ .}
\end{align}
A related argument was also made by \citet{Asuncion:2009:SIT:1795114.1795118}, who note that several topic modeling inference algorithms differ only by offsets of 1 or 0.5 to the counts, and state that ``\emph{since $n_k$ is usually large, we do not expect [a small offset] to play a large role in learning}.'' To formalize this intuition, as we show in Appendix \ref{sec:boundPhiPApprox}, the approximation can be bounded from below and from above as a function of $n_k^{(i)}$ as:
\begin{align}
\frac{n_k^{(i)} + \sum_{v^\prime = 1}^V\beta_{v^\prime}}{n_k^{(i)} + \sum_{v^\prime = 1}^V\beta_{v^\prime} + 1} & E_{p(\mathbf{z}| \mathbf{w}, \boldsymbol{\beta}, \mathbf{z} \in Z_{Gibbs}(\mathbf{z}^{(i)}))}\Big [\frac{\sum_{d=1}^{D_{Train}}\sum_{j:w_{d,j} = v} z_{djk} + \beta_v}{ \sum_{d=1}^{D_{Train}}\sum_{j=1}^{N_d} z_{djk}^{(i)} + \sum_{v^\prime = 1}^V\beta_{v^\prime}} \Big ] \nonumber \\
\leq \ & E_{p(\mathbf{z}| \mathbf{w}, \boldsymbol{\beta}, \mathbf{z} \in Z_{Gibbs}(\mathbf{z}^{(i)}))}\Big [\frac{\sum_{d=1}^{D_{Train}}\sum_{j:w_{d,j} = v} z_{djk} + \beta_v}{ \sum_{d=1}^{D_{Train}}\sum_{j=1}^{N_d} z_{djk} + \sum_{v^\prime = 1}^V\beta_{v^\prime}} \Big ] \nonumber \\
\leq \frac{n_k^{(i)} + \sum_{v^\prime = 1}^V\beta_{v^\prime}}{n_k^{(i)} + \sum_{v^\prime = 1}^V\beta_{v^\prime} - 1} & E_{p(\mathbf{z}| \mathbf{w}, \boldsymbol{\beta}, \mathbf{z} \in Z_{Gibbs}(\mathbf{z}^{(i)}))}\Big [\frac{\sum_{d=1}^{D_{Train}}\sum_{j:w_{d,j} = v} z_{djk} + \beta_v}{ \sum_{d=1}^{D_{Train}}\sum_{j=1}^{N_d} z_{djk}^{(i)} + \sum_{v^\prime = 1}^V\beta_{v^\prime}} \Big ] \mbox{ ,}
\end{align}
 for $n_k^{(i)} > 0$. Since $\lim_{n_k^{(i)} \rightarrow \infty} \frac{n_k^{(i)} + \sum_{v^\prime = 1}^V\beta_{v^\prime}}{n_k^{(i)} + \sum_{v^\prime = 1}^V\beta_{v^\prime} + 1} = 1$ and $\lim_{n_k^{(i)} \rightarrow \infty} \frac{n_k^{(i)} + \sum_{v^\prime = 1}^V\beta_{v^\prime}}{n_k^{(i)} + \sum_{v^\prime = 1}^V\beta_{v^\prime} - 1} = 1$, the left and right hand sides of Equation \ref{eq:approximationphip2} will converge in the limit as the $n_k^{(i)}$'s go to infinity, e.g. when increasing the counts via a stream of incoming documents.  In Appendix \ref{sec:validateEq47}, we provide a further empirical illustration of this approximation, by taking subsets of the documents in the corpus to vary $n_k$, and observing that as $n_k$ increases the difference between the left and right sides of Equation \ref{eq:approximationphip2} approaches zero.

Finally, using the same stationarity and law of large numbers arguments as before, we obtain the $\hat{\phi}^{p}$ estimator by plugging in the transition operator, and then renormalizing,
\begin{align}
\hat{\phi}^{p}_{kv}&\propto \sum_{d=1}^{D_{Train}}\sum_{j:w_{d,j} = v} E_{p(\mathbf{z}| \mathbf{w}, \boldsymbol{\beta}, \mathbf{z} \in Z_{Gibbs}(\mathbf{z}^{(i)})) }[z_{djk}] + \beta_v = \sum_{d=1}^{D_{Train}}\sum_{j:w_{d,j} = v} T(z_{djk}^{(i)\rightarrow j}|\mathbf{z}^{(i)}) + \beta_v \mbox{ .}
\end{align}
More formally, the full argument is given mathematically as
\begin{align}
 \hat{\phi}^{p}_{kv} &\propto \sum_{d=1}^{D_{Train}}\sum_{j:w_{d,j} = v} T(z_{djk}^{(i)\rightarrow j}|\mathbf{z}^{(i)}) + \beta_v \\
 &= \sum_{d=1}^{D_{Train}}\sum_{j:w_{dj} = v} E_{T(\mathbf{z}_d^{(i)\rightarrow j}|\mathbf{z}_d^{(i)})}[z_{djk}^{(i)\rightarrow j}] + \beta_v \\
 &= \sum_{d=1}^{D_{Train}}\sum_{j:w_{dj} = v} E_{p(\mathbf{z}| \mathbf{w}, \boldsymbol{\beta}, \mathbf{z} \in Z_{Gibbs}(\mathbf{z}^{(i)}))}[z_{djk}] + \beta_v \tag*{ \mbox{  (by stationarity of the Gibbs sampler)}} \\
 &\propto \frac{\sum_{d=1}^{D_{Train}}\sum_{j:w_{dj} = v} E_{p(\mathbf{z}| \mathbf{w}, \boldsymbol{\beta}, \mathbf{z} \in Z_{Gibbs}(\mathbf{z}^{(i)})) }[z_{djk}] + \beta_v}{ \sum_{d=1}^{D_{Train}}\sum_{j=1}^{N_d} z_{djk}^{(i)} +  \sum_{v^\prime = 1}^V\beta_{v^\prime}}\\
 &=   E_{p(\mathbf{z}| \mathbf{w}, \boldsymbol{\beta}, \mathbf{z} \in Z_{Gibbs}(\mathbf{z}^{(i)})) } \Big [\frac{\sum_{d=1}^{D_{Train}}\sum_{j:w_{dj} = v}z_{djk} + \beta_v}{ \sum_{d=1}^{D_{Train}}\sum_{j=1}^{N_d} z_{djk}^{(i)} +   \sum_{v^\prime = 1}^V\beta_{v^\prime}} \Big ]\\
 &\approx E_{p(\mathbf{z}| \mathbf{w}, \beta, \mathbf{z} \in Z_{Gibbs}(\mathbf{z}^{(i)}))}\Big [\frac{\sum_{d=1}^{D_{Train}}\sum_{j:w_{dj} = v} z_{djk} + \beta_v}{ \sum_{d=1}^{D_{Train}}\sum_{j=1}^{N_d} z_{djk} +   \sum_{v^\prime = 1}^V\beta_{v^\prime}} \Big ] \tag*{ (if $n_k >> 1$)}\\
 &= p(w^{(new)} = v| \mathbf{w}, \boldsymbol{\beta}, z^{(new)} = k, \mathbf{z} \in Z_{Gibbs}(\mathbf{z}^{(i)})) \mbox{ .}
\end{align}


In Algorithm \ref{alg:methods}, along with $\theta^p$'s procedure, 
we also present the pseudocode for the $\phi^p$ estimator.



\section{Unsupervised Learning Experiments --- LDA}
\label{sec:expsunsup}
This section describes the experiments that we performed to study the behavior of our proposed estimators in the context of unsupervised LDA models. We describe the data sets, the evaluation procedure and the experiments performed in the unsupervised LDA setting.
Apart from the experiments reported here, we have performed two additional ones reported in the Appendix, (a) comparing the document-topic counts of the Collapsed Gibbs Sampling (CGS) algorithm against the \emph{soft} document-topic counts, in order to study their differences as the algorithm converges and (b) providing an empirical validation of the approximation that we use in Equation \ref{eq:approximationphip2} regarding our $\phi^p$ estimator's derivation. The relevant code of the following experiments (along with the code of the multi-label experiments of Section \ref{sec:multi}) can be found in \url{https://github.com/ypapanik/cgs_p}.

\subsection{Data Sets}
\label{sec:datasets1}
Four data sets were used in the unsupervised setting: a) BioASQ, b) New York Times, c) Reuters-21578 and d) TASA. The number of documents in the training and test sets, the average length of the training documents and the size of the vocabulary of word types found in the training set are presented in Table \ref{tbl:unsupdatasets}.

The first data set originates from the \textit{BioASQ} challenge \citep{BalikasPNKP14} that deals with large-scale online multi-label classification of biomedical journal articles. This learning task is particularly challenging as the taxonomy of labels includes around $27,000$ terms, with highly imbalanced frequencies. For the unsupervised experiments of this section, we used the $30,000$ last documents of the BioASQ corpus of the year 2014, using the first $20,000$ as training documents and the rest as test documents. To construct the data set we concatenated the abstract and title of each article and removed common stopwords. The remaining unigrams were used as word types. 

The second data set contains articles published by the {\em New York Times}, manually annotated via their indexing service. We used the same data set as used by \citet{Rubin:2012:STM:2339279.2339301}, with the same training set (14,668 documents) and keeping the first 7,000 documents for testing (out of the 15,989 of the original paper). 

The third data set\footnote{\url{http://disi.unitn.it/moschitti/corpora/Reuters21578-Apte-115Cat.tar.gz}} contains documents from the {\em Reuters} news-wire and has been widely used among researchers for almost two decades. The split that we used has 11,475 documents for training and 4,021 documents for testing. We preprocessed the corpus by removing common stopwords.

The \textit{TASA} data set contains $37,650$ documents of diverse educational materials (e.g., health, sciences, etc) collected by Touchstone Applied Science Associates \citep{citeulike:2243850}. We used the first 30,121 documents as a training set and the remaining as a test set. The corpus already had stopwords and infrequent words removed, so we did not perform any further preprocessing.

\begin{table}[tb]

\centering
\begin{tabular}{ccccc}
\toprule
\noalign{\smallskip}
Data Set & $D_{Train}$ & $D_{Test}$ & $\overline{N_d}$ & $V$\\
\midrule
\noalign{\smallskip}
BioASQ &20,000&10,000&113.55&135,186\\
New York Times &14,668&7,000&581.13&19,259\\
Reuters-21578 &11,475&4,021&49.68&41,014\\
TASA & 30,121 &7,529 &107.62 &21,970\\
\bottomrule
\end{tabular}
\caption{Statistics for the data sets used in the unsupervised learning experiments. Average document length and number of word types refer to the training sets.}
\label{tbl:unsupdatasets}
\end{table}

\subsection{Evaluation}

Evaluation of LDA models typically focuses on the probability of a set of held-out documents given an already trained model \citep{Wallach:2009:EMT:1553374.1553515}. In this context, one must compute the model's posterior predictive likelihood of all words in the test set, given estimates of the topic parameters $\phi$ and the document-level mixture parameters $\theta$.

The log likelihood of a set of test documents $D_{Test}$, given an already estimated model $M$, is given by \citet{Heinrich04parameterestimation} as:

\begin{equation}
\textrm{Log Likelihood} = \sum\limits_{d=1}^{D_{Test}} \textrm{log} p(\mathbf{w}_d|M) =\sum\limits_{d=1}^{D_{Test}} \sum\limits_{i=1}^{N_d} \textrm{log}\sum\limits_{k=1}^{K}(\phi_{kv} \cdot \theta_{dk}) 
\label{eq:likelihood}
\end{equation}
with $w_i=v$. The perplexity will be:
\begin{equation}
 \textrm{Perplexity} = \textrm{exp}({-\frac{ \textrm{Log Likelihood}}{\sum\limits_{d=1}^{D_{Test}} N_d}})
\end{equation}

\noindent where lower values of perplexity signify a better model.

Since $\theta_d$ is unknown for test documents, and is intractable to marginalize over, a common practice in the literature in order to compute the above likelihood is to run the CGS algorithm for a few iterations on the first half of each document, and then to compute the perplexity of the held-out data (the second half of each document), based on the trained model's posterior predictive distribution over words \citep{Asuncion:2009:SIT:1795114.1795118}. This is the approach we follow.

\subsection{Comparison between CGS and CGS$_p$}
\label{sec:exp1}
In this first experiment, the motivation is to validate the previously presented theory behind the $\phi^p$ and $\theta^p$ methods. Specifically, we are interested in verifying the following hypothesis: given a single burned-in sample, $\hat{\phi^p}$ or $\hat{\theta^p}$ will more effectively estimate the respective LDA parameters compared to the standard estimators, since the CGS$_p$ estimators make use of the full dense probabilities, through the infinite l-steps of Eq. \ref{eq:thetap}. This advantage would correspond to lower perplexity values  for a single burned-in sample for the CGS$_p$ estimators compared to the standard CGS estimator. Furthermore, when averaging over multiple samples to estimate the $\theta$ parameters during prediction, we expect that the $\theta^p$ and standard $\theta$ estimators will eventually converge to the same solution given enough samples or Markov chains, since $\theta^p$ aims to compute the posterior mean.  However, we expect that $\theta^p$ will converge to the posterior mean more rapidly in the number of samples that are averaged over than the standard estimator $\theta$, with correspondingly faster improvement in perplexity.

For this experiment we used all four data sets and considered four different topic number configurations (20, 50, 100, 500). For brevity, we report only the results for $K=100$ and include the rest of the plots in Appendix \ref{app1}. During training, we ran one Markov chain for 200 iterations and took a single sample to calculate $\phi$ and $\phi^p$ from the same chain. During prediction, since $\phi$ is fixed and topics are not exchanged through the Gibbs sampler's iterations, we took multiple samples from multiple Markov chains, and averaged over these samples using the $\theta$ (standard CGS) and $\theta^p$ estimators. The burn-in period for the Gibbs sampler was set to 50 iterations and a sampling lag of 5 iterations was used. The $\boldsymbol{\alpha}$ and $\boldsymbol{\beta}$ hyperparameters were symmetrical across all topics and words respectively and set such that $\alpha_k = 0.1$ and $\beta_v = 0.01$. Finally, to ensure fairness between the $\theta$ and $\theta^p$ estimators, we used the same samples from the same chains to compute the respective estimates. 

\begin{figure}[tb]
\begin{minipage}{\textwidth}
\begin{minipage}{0.49\textwidth}
  \includegraphics[width=\linewidth]{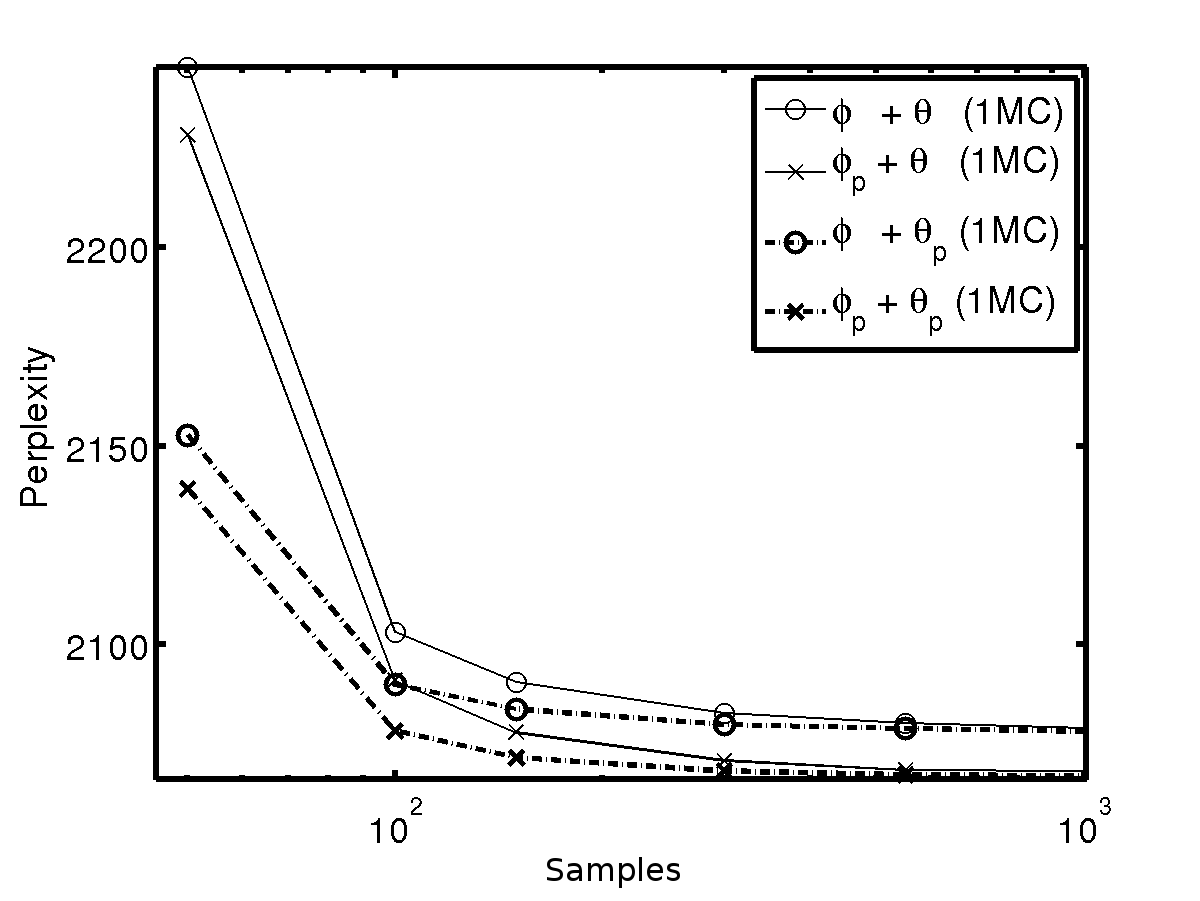}
\end{minipage}  
\hspace*{\fill}
\begin{minipage}{0.49\textwidth}
  \includegraphics[width=\linewidth]{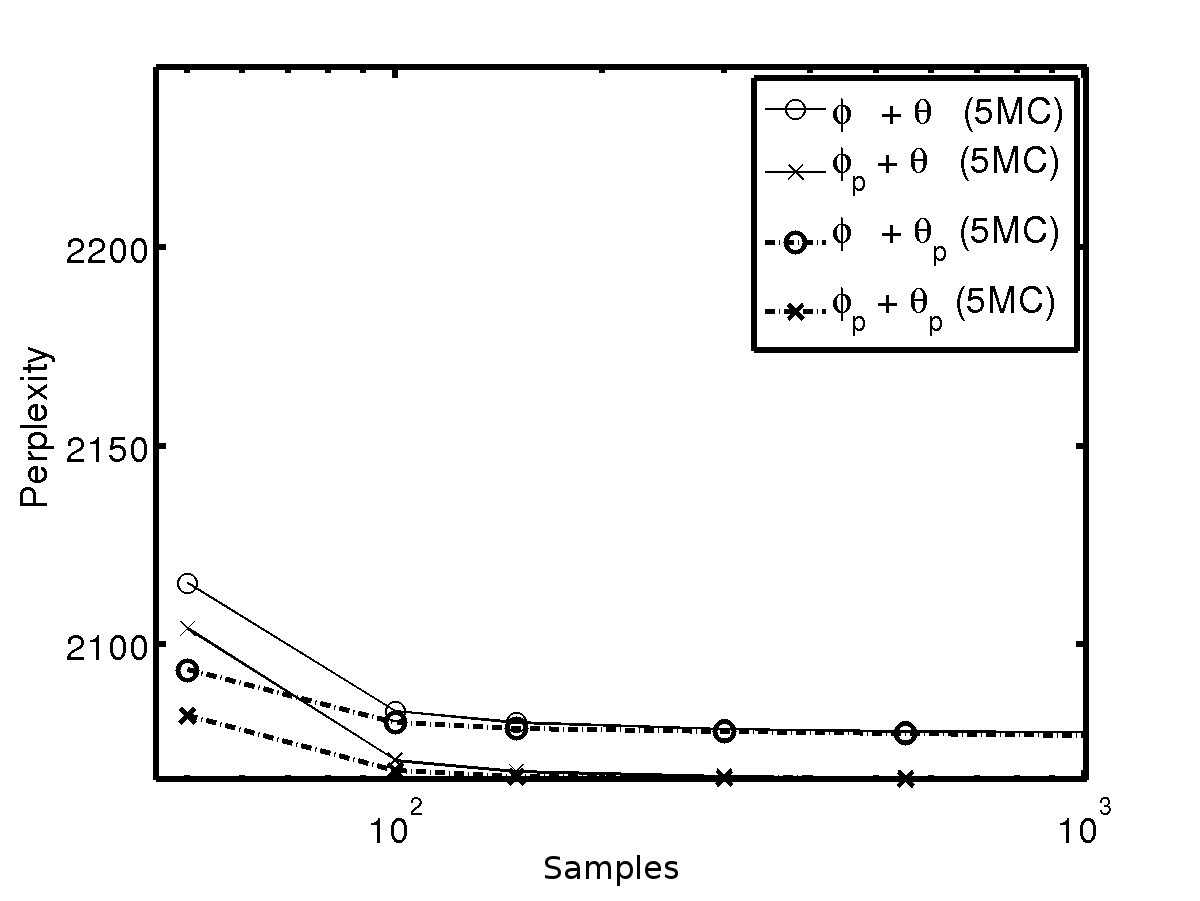}
\end{minipage}

\subcaption{BioASQ}
\end{minipage}
\begin{minipage}{\textwidth}
\begin{minipage}{0.49\textwidth}
  \includegraphics[width=\linewidth]{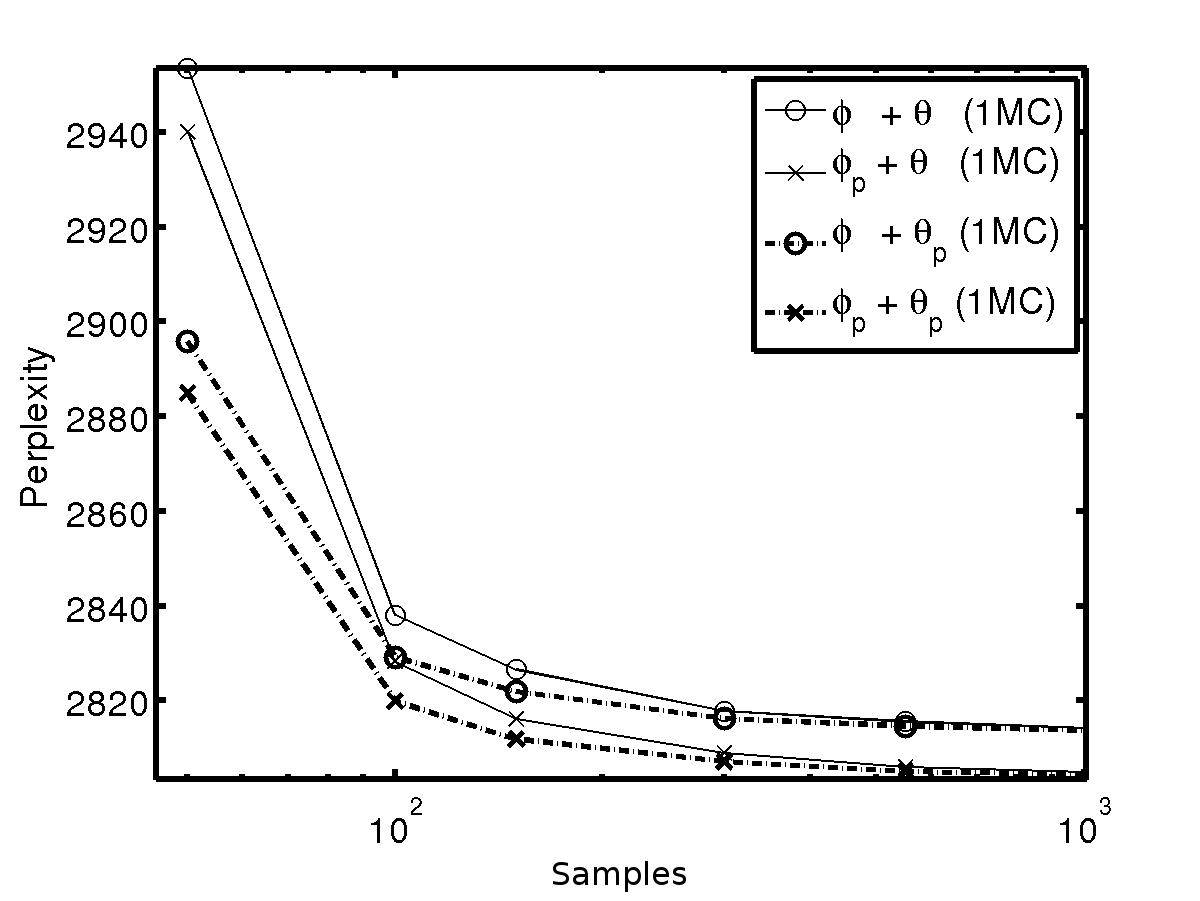}
\end{minipage}  
\hspace*{\fill}
\begin{minipage}{0.49\textwidth}
  \includegraphics[width=\linewidth]{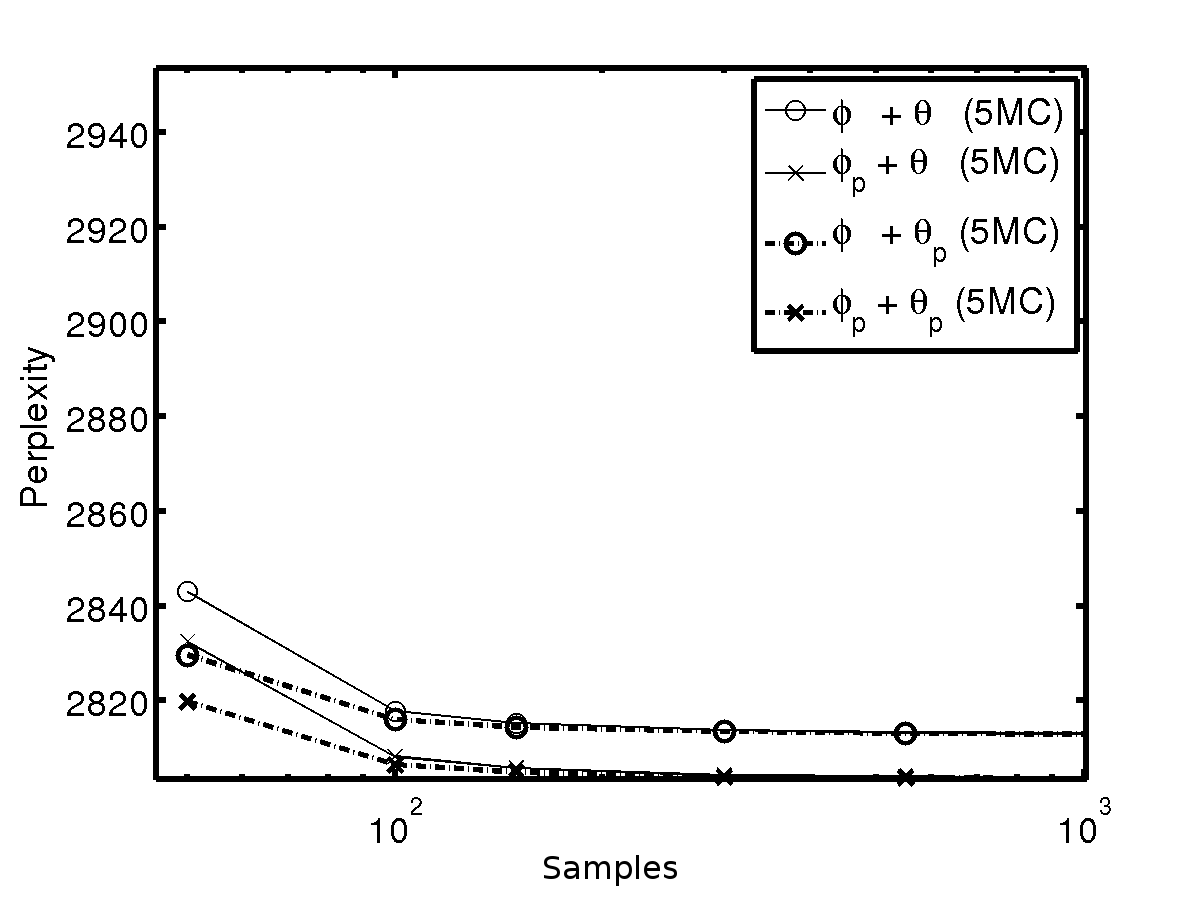}
\end{minipage}
  \subcaption{New York Times}
\end{minipage}
\caption{Perplexity against the number of samples averaged over, for the CGS$_p$ estimators and standard CGS. Results are taken by averaging over 5 different runs. Samples are taken after a burn-in period of 50 iterations.}
\label{fig:exp1a}
\end{figure}

\begin{figure}[tb]
\begin{minipage}{\textwidth}
 \begin{minipage}{0.49\textwidth}
  \includegraphics[width=\linewidth]{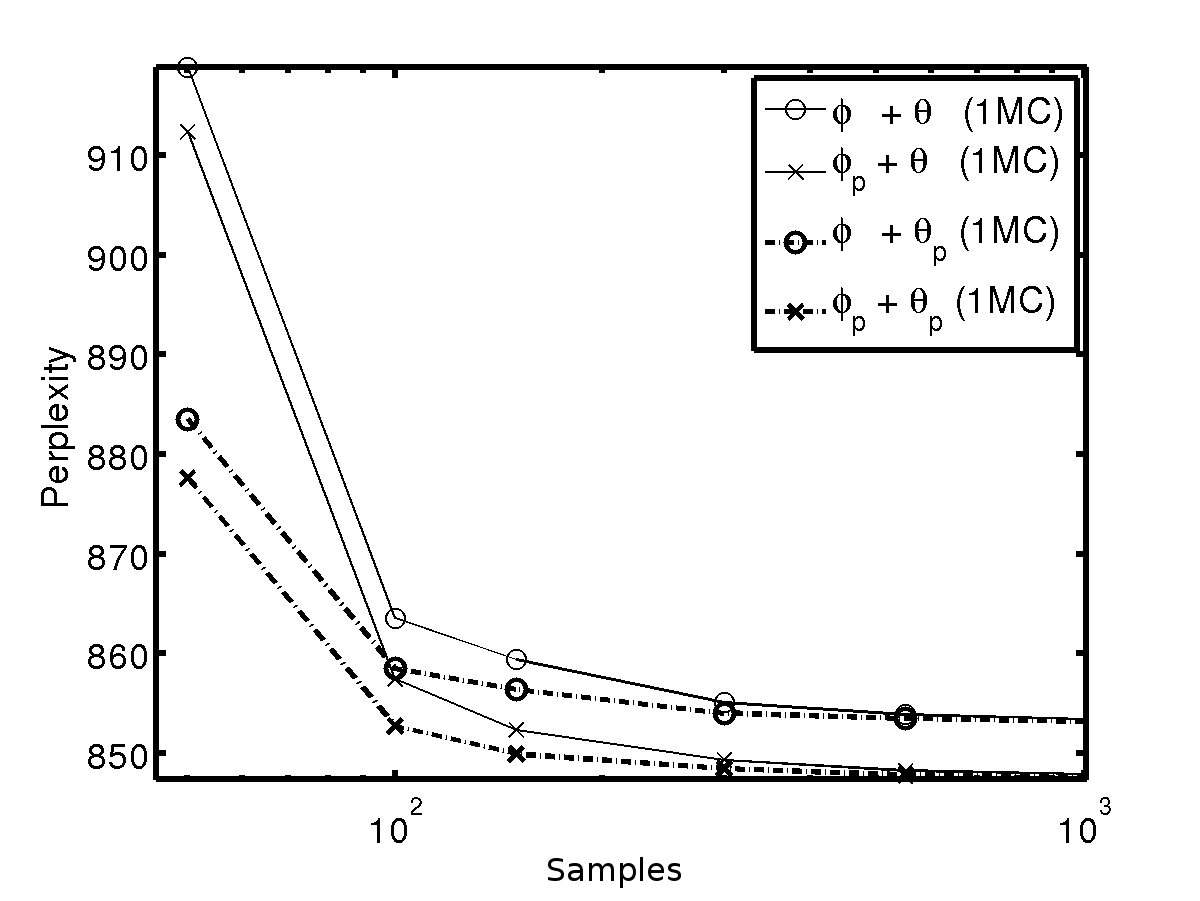}
\end{minipage}  
\hspace*{\fill}
\begin{minipage}{0.49\textwidth}
  \includegraphics[width=\linewidth]{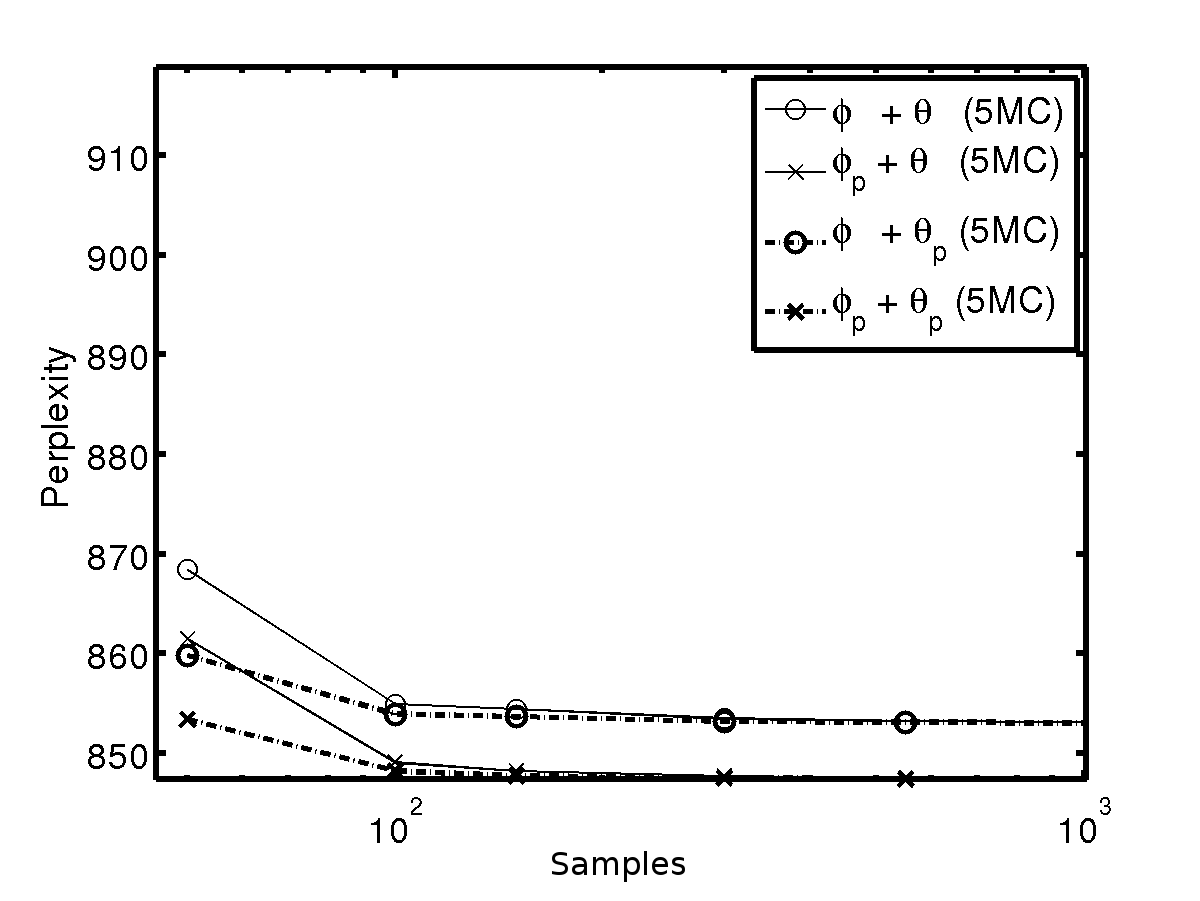}
\end{minipage}
\subcaption{Reuters-21578}
\end{minipage}
\begin{minipage}{\textwidth}
\begin{minipage}{0.49\textwidth}
  \includegraphics[width=\linewidth]{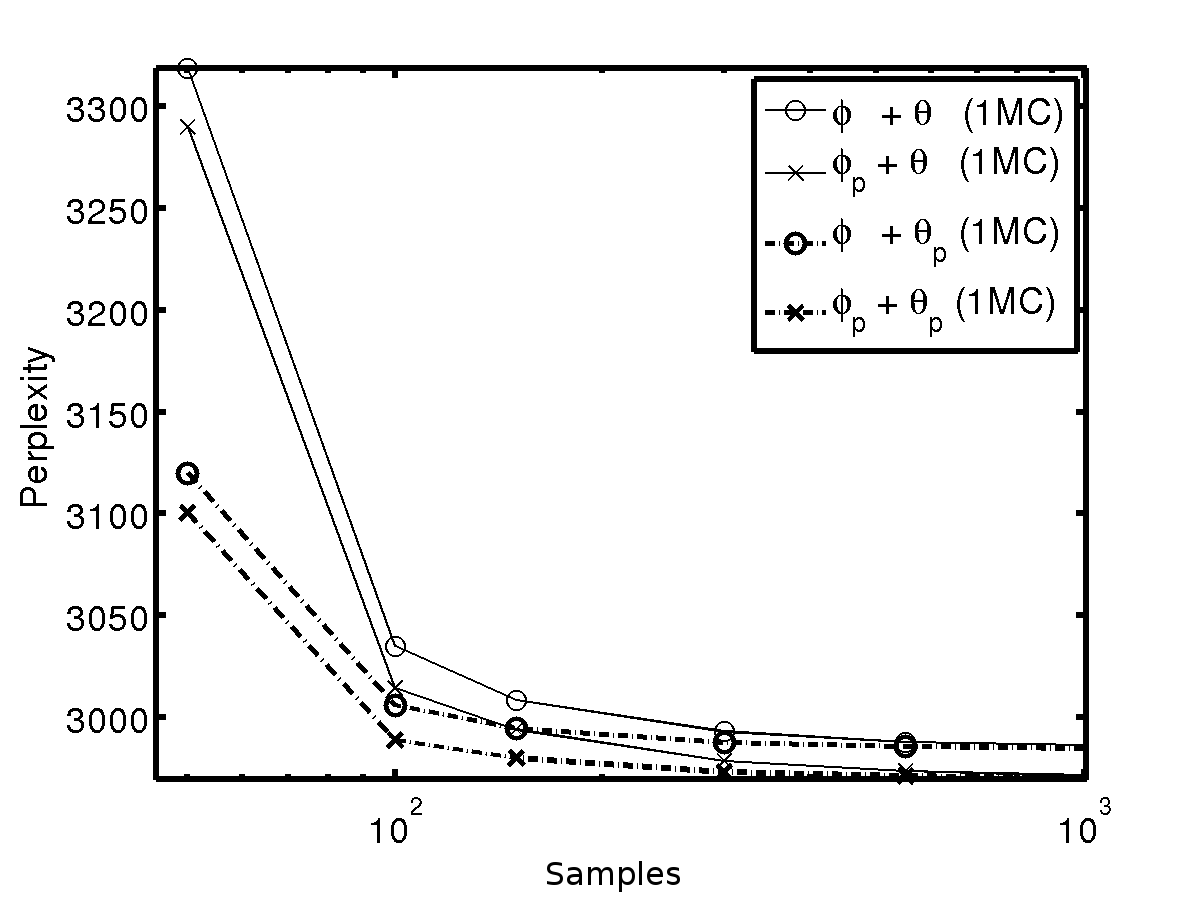}
\end{minipage}  
\hspace*{\fill}
\begin{minipage}{0.49\textwidth}
  \includegraphics[width=\linewidth]{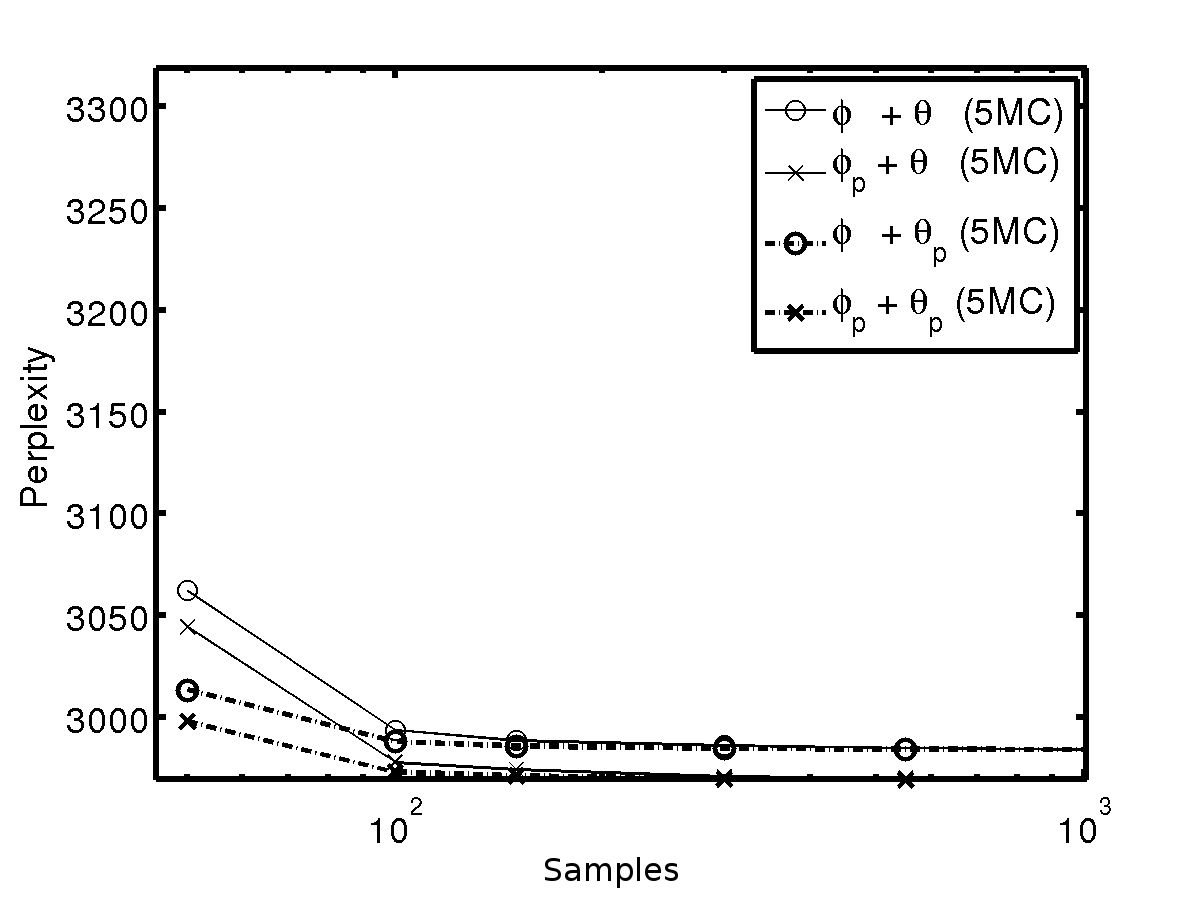}
\end{minipage}
  \subcaption{TASA}
\end{minipage}
\caption{Perplexity against the number of samples averaged over, for the CGS$_p$ methods and standard CGS. Results are taken by averaging over 5 different runs. Samples are taken after a burn-in period of 50 iterations.}
\label{fig:exp1b}
\end{figure}

In Figures \ref{fig:exp1a} and \ref{fig:exp1b} we show the perplexity scores against the number of samples that are averaged over, after burn-in, for one and five Markov chains respectively. All four combinations of the estimators are depicted. By observing the results we notice some key points across all plots. First, if we consider the case where only one burned-in sample is used, corresponding to the left-most points in the plots, we can notice that both $\theta^p$ methods ($\phi + \theta^p$ and $\phi^p + \theta^p$) decisively outperform the other two methods. This observation is particularly important for scenarios where we can't afford to average over many samples due to computational or time limitations, as frequently occurs during the comparative evaluation of topic modeling methods, and in multi-label document classification.

We also see that the $\phi^p$ methods ($\phi^p + \theta$ and $\phi^p + \theta^p$) have a short but steady advantage over the respective $\phi$ estimators. This advantage is not diminished for more samples or Markov chains, suggesting that $\phi^p$ is actually a more accurate estimator than $\phi$.  We also remind here that in unsupervised scenarios we cannot average over samples to compute a $\phi$ estimate during training, since topics may be interchanged between iterations. A third remark, also aligned with our theoretical results, is that $\theta$ and $\theta^p$ converge to the same solution after a sufficient number of samples are averaged over, although $\theta^p$ converges much more rapidly. Overall, these findings confirm that indeed both $\theta^p$ and $\phi^p$ constitute improved estimators compared to the respective standard ones: $\theta^p$ provides more rapid convergence than using the standard estimator $\theta$ when averaging over samples for prediction on test documents, and $\phi^p$ improves perplexity due to implicit averaging. The rest of the plots in Appendix \ref{app1} are in accordance with the above observations.

\subsection{Word Association: $\phi^p$ vs $\phi$}
\label{sec:associations}

\begin{figure}[h]
\begin{minipage}{\textwidth}
 \begin{minipage}{0.49\textwidth}
  \includegraphics[width=\linewidth]{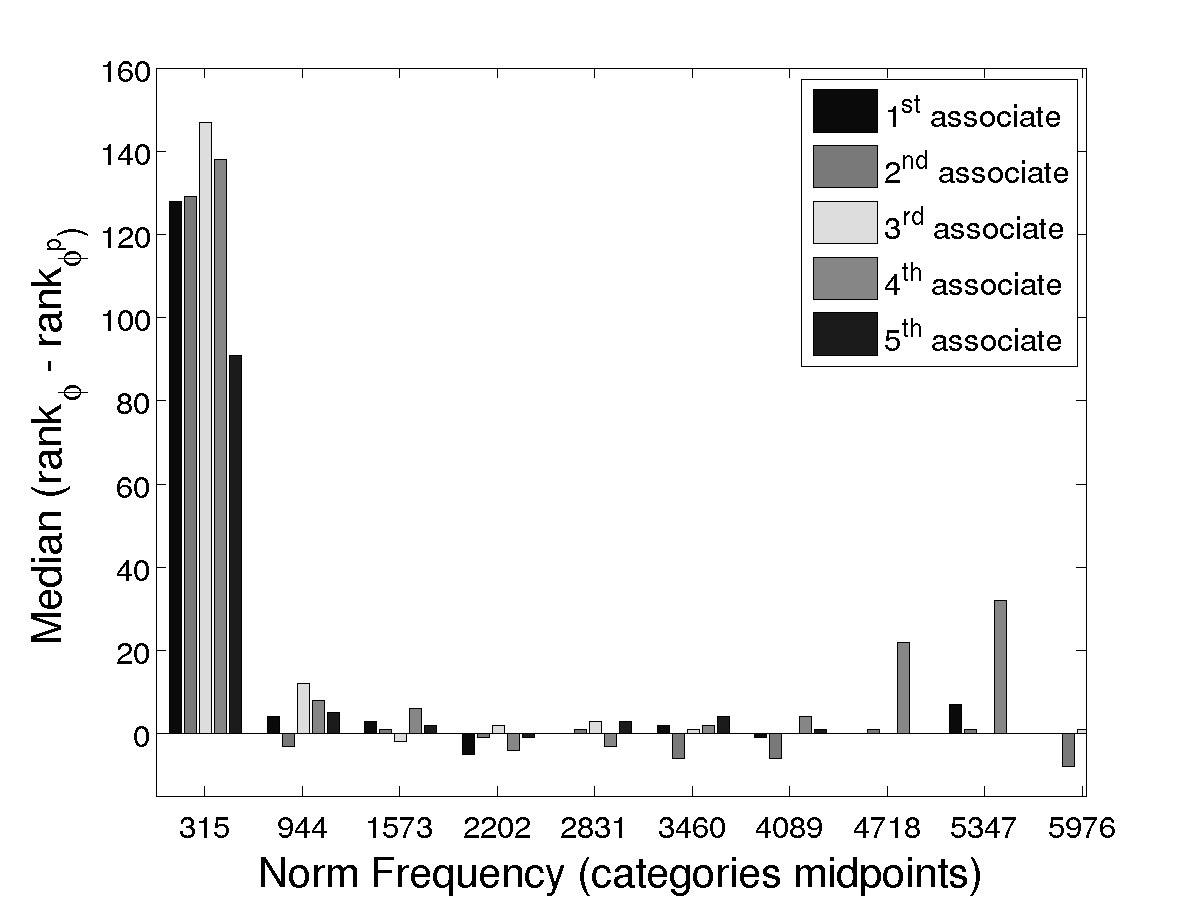}
  \subcaption{K = 50}
\end{minipage}  
\hspace*{\fill}
\begin{minipage}{0.49\textwidth}
  \includegraphics[width=\linewidth]{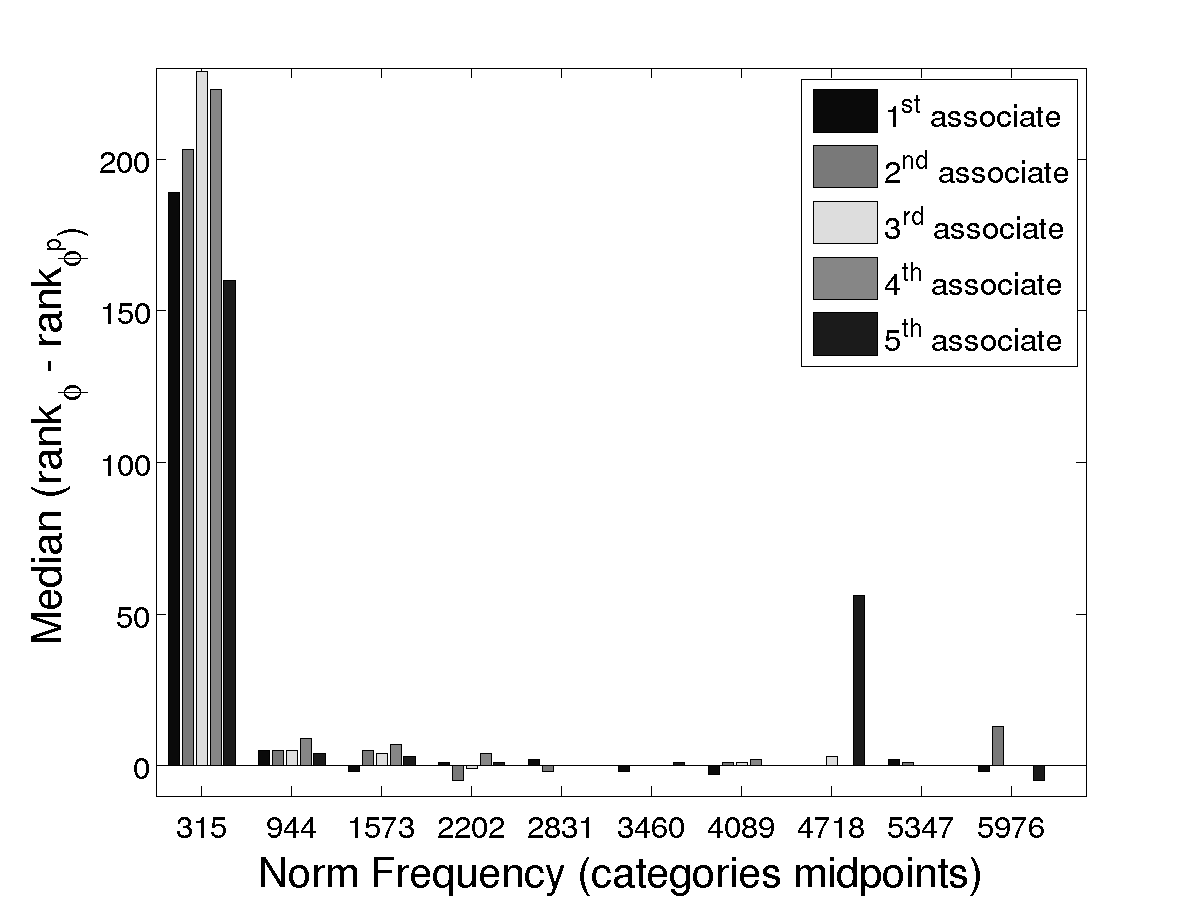}
  \subcaption{K = 100}
\end{minipage}

\end{minipage}
\begin{minipage}{\textwidth}
\begin{minipage}{0.49\textwidth}
  \includegraphics[width=\linewidth]{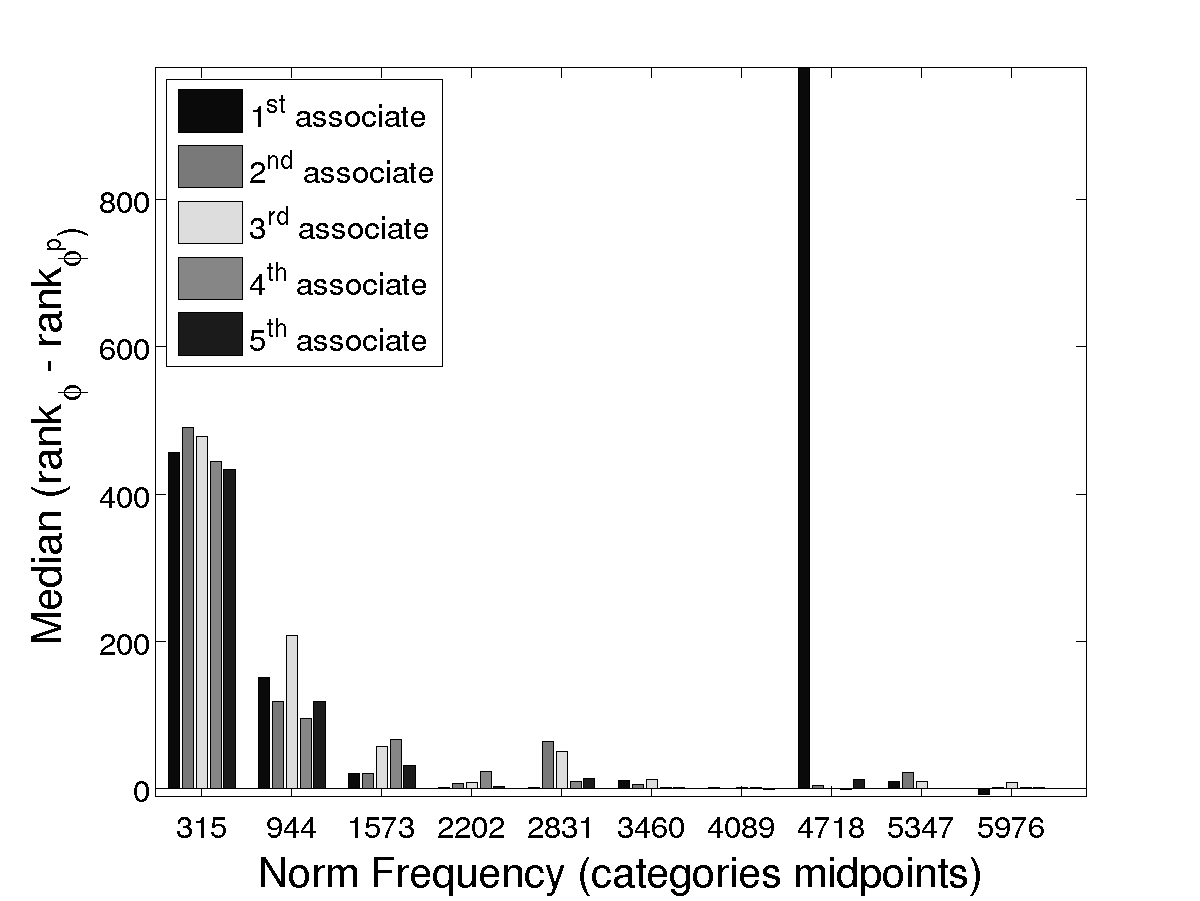}
  \subcaption{K = 500}
\end{minipage}  
\hspace*{\fill}
\begin{minipage}{0.49\textwidth}
  \includegraphics[width=\linewidth]{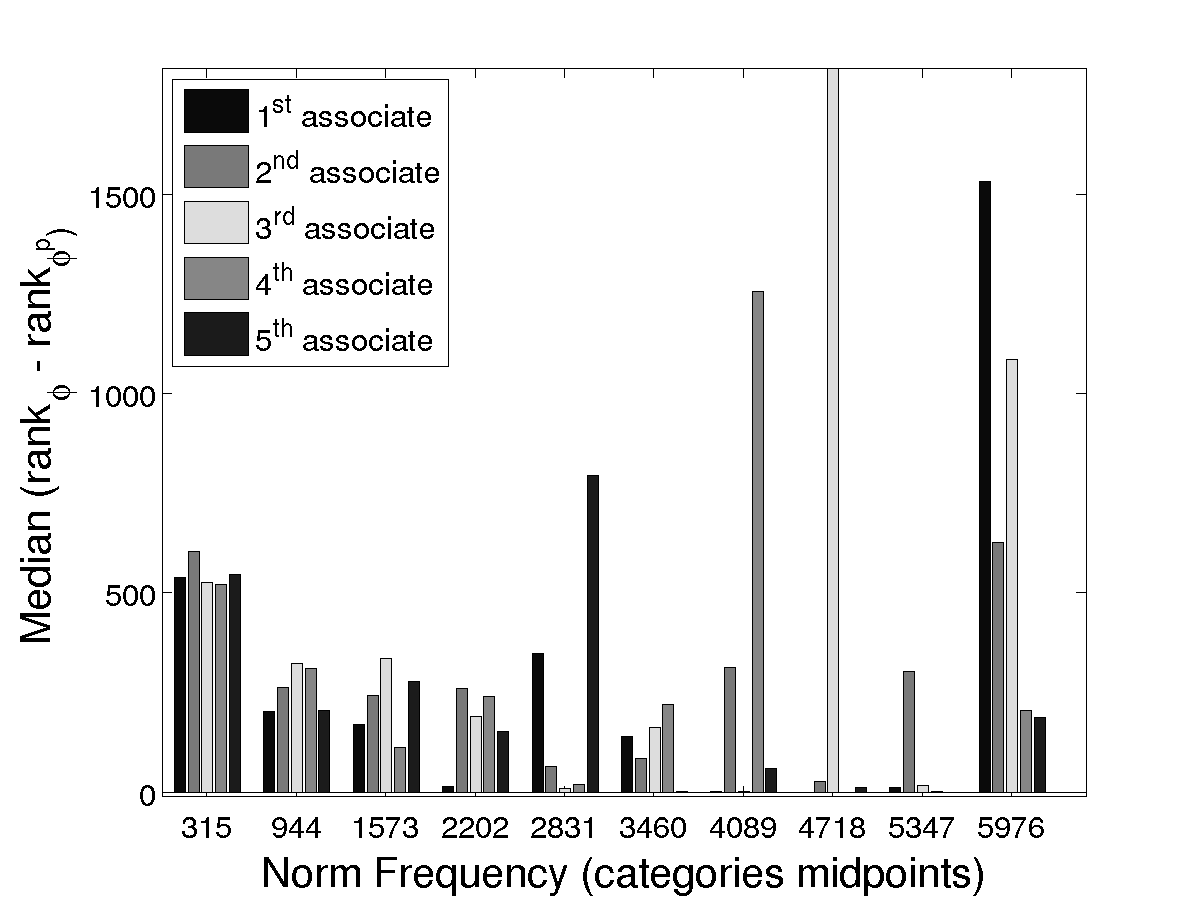}
  \subcaption{K = 1000}
\end{minipage}

\begin{center}
\begin{minipage}{0.4\textwidth}
  \includegraphics[width=\linewidth]{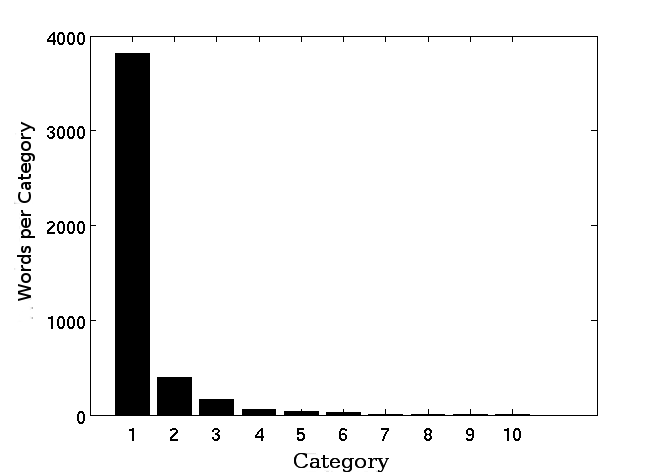}
  \subcaption{Norms per category.}
  \end{minipage}
\end{center}

\end{minipage}
\caption{Median difference in ranks produced by $\phi^p$ and $\phi$ estimators for the first five associate words, for a set of 4,506 norms \citep{nelson2004university}. The estimators were computed after training LDA on the TASA corpus. A positive value indicates an advantage for $\phi^p$.
}
\label{fig:expAssociations}
\end{figure}

In this experiment, we compare $\phi^p$ with $\phi$ on a word association task. In word association, a given word, called the \emph{norm} or \emph{cue} word, is associated with a number of semantically related words, called \emph{targets} or \emph{associates}. We consider the data set provided by \citet{nelson2004university}, which contains a set of 5,019 norm words and for each of them a set of associate words provided by human annotators.  We aim to see how, given a specific cue word, the two LDA estimators rank the corresponding associates, in order to assess which of the two performs better at predicting the targets, in terms of the median difference in ranks.  The word association task provides a useful benchmark for evaluating the extent to which the topic representations are a good model of human semantic representation \citep{griffiths2007topics}.

We largely follow the same setup as that described by \citet{griffiths2007topics}, pages 220-224. Specifically, we use the TASA data set, removing only stopwords and word types with fewer than five and more than ten thousand occurrences in the corpus, resulting in a vocabulary of 26,186 word types. There are 4,506 norms that belong to the obtained vocabulary. Subsequently, we train an LDA  model on the corpus for $K = 50, 100, 500, 1000$ and for 250 iterations, obtaining a single point estimate for $\phi^p$ and $\phi$ for each value of $K$ at the end of the procedure. Our goal is to see how each of the two estimators rank the five most commonly associated responses to each of the norm words, as given by human subjects. Similarly to \cite{griffiths2007topics} (refer to p.220 and Appendix B, p.244), in order to assess semantic association we use the conditional probability $p(w_2|w_1)$ where $w_1$ stands for a given cue word and $w_2$ denotes each of the rest of the vocabulary word types. Supposing that $w_1$ and $w_2$ belong to the same topic $z$, and assuming a uniform prior on $z$, we have\footnote{A uniform prior $p(z)$ follows if $z$ is understood to be the first word in a document under the collapsed LDA urn model, with symmetric hyperparameters $\alpha$.  With asymmetric hyperparameters, $p(z = k) \propto \alpha_k$.}

\begin{align}
    p(w_2|w_1) = \sum_z p(w_2|z) p(z|w_1) &= \sum\limits_z{p(w_2|z)\frac{p(w_1|z)p(z)}{\sum\limits_{z'}{p(w_1|z')p(z')}}} \nonumber\\
    &= \sum\limits_z{p(w_2|z)\frac{p(w_1|z)}{\sum\limits_{z'}{p(w_1|z')}}} \mbox{ .}
\end{align}

Next, we consider the first five associate words of the norm $w_1$ and obtain the rank for each of them according to the probabilities $p(w_2|w_1)$, for each of $\phi^p$ and $\phi$. In Figure \ref{fig:expAssociations}, we report the median difference in ranks, $rank_{\phi} - rank_{\phi^p}$ against word frequency, a positive value indicating an advantage for $\phi^p$. To enhance readability, we have grouped the norms according to their frequencies within the corpus, into 10 categories.

We observe two steady trends. First, $\phi^p$ clearly outperforms $\phi$ for rare norms (the first category), which is also the most populous (3,814 norms). For more frequent norms, $\phi^p$ and $\phi$ behave almost on par. This trend reveals the distinctive benefit of our proposed estimator compared to its counterpart, especially if we take into account that word type frequencies typically follow a power law like distribution in document collections.  A second trend, consistent with our other experiments, relates to the number of topics $K$: as $K$ increases and the complexity of the LDA model correspondingly increases, the advantage of $\phi^p$ becomes wider compared to $\phi$.  We similarly observe that $\phi^p$ provides a benefit for increasingly frequent words as $K$ increases.

These results are consistent with our analysis in Section \ref{sec:phi_p} which showed that $\phi^p$ performs implicit averaging in order to improve the stability of the topics.  The nuanced uncertainty information, and corresponding stability, provided by $\phi^p$ is most important for less frequent words, and for larger $K$, for which the posterior uncertainty in the (smaller) count values  $n_{kv}$ per (word,topic) pair is most consequential. For larger (word, topic) counts $n_{kv}$, the expected counts estimated by $\phi^p$ (Equation \ref{eq:approximationphip2}) approach the observed counts computed by $\phi$ due to the law of large numbers.  Consequently, we typically observe little difference in the top-words lists generated by $\phi^p$ and $\phi$, which are often used to assess the interpretability of the topics.  Instead, the results of this experiment suggest that $\phi^p$ conveys the most benefit for tasks that depend on the word/topic probabilities for the less frequent to moderately frequent words, as in word association, and any other task which requires semantic representations of the words.

\subsection{Comparison between CGS, CGS$_p$ and CVB0 }
\label{sec:exp2}

In this experiment we compared the CGS, CGS$_p$ and CVB0 algorithms in terms of perplexity, across different numbers of topics. The motivation here was to examine how the CGS$_p$ method would perform compared to the rest of the algorithms, in a variety of configurations and data sets.

\begin{figure}[tb]
\begin{minipage}{0.49\textwidth}
  \includegraphics[width=\linewidth]{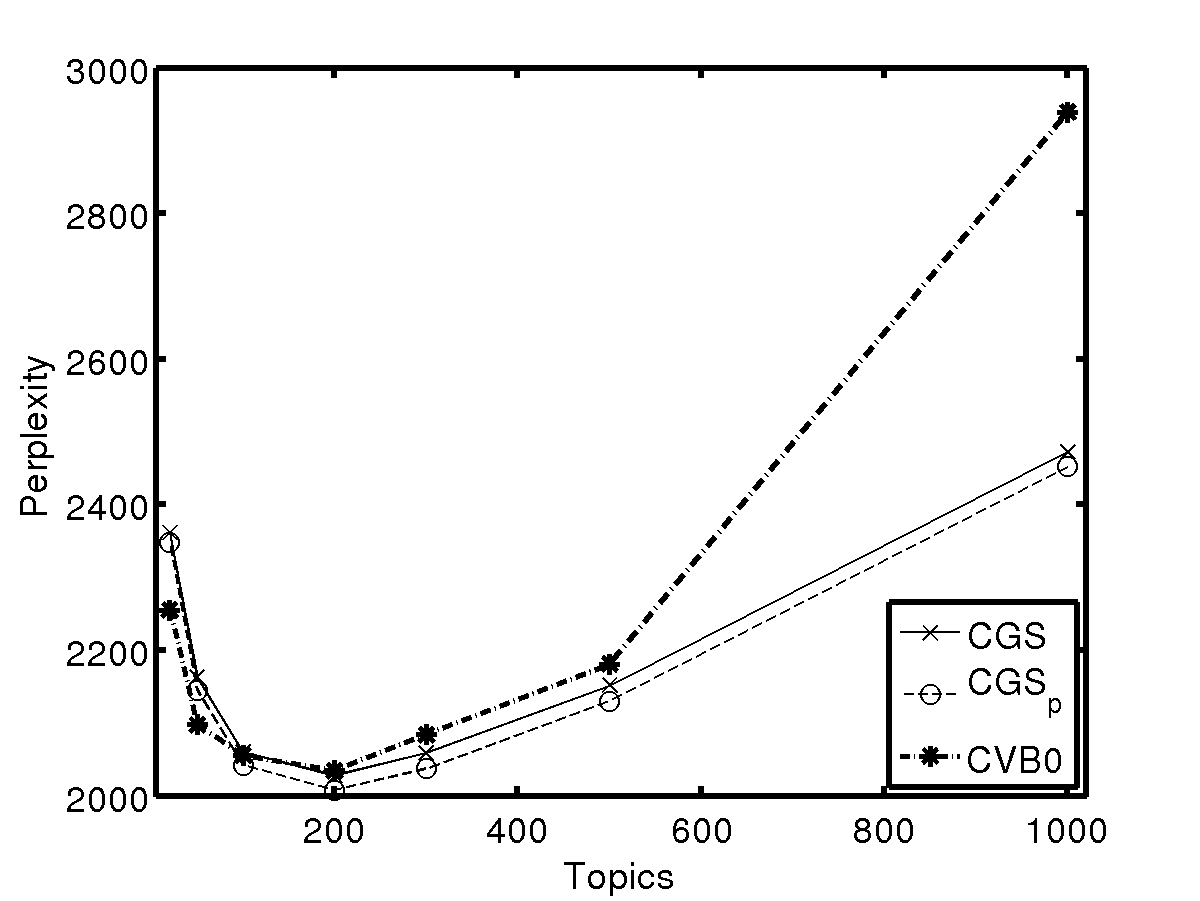}
  \subcaption{BioASQ}
\end{minipage}  
\hspace*{\fill}
\begin{minipage}{0.49\textwidth}
  \includegraphics[width=\linewidth]{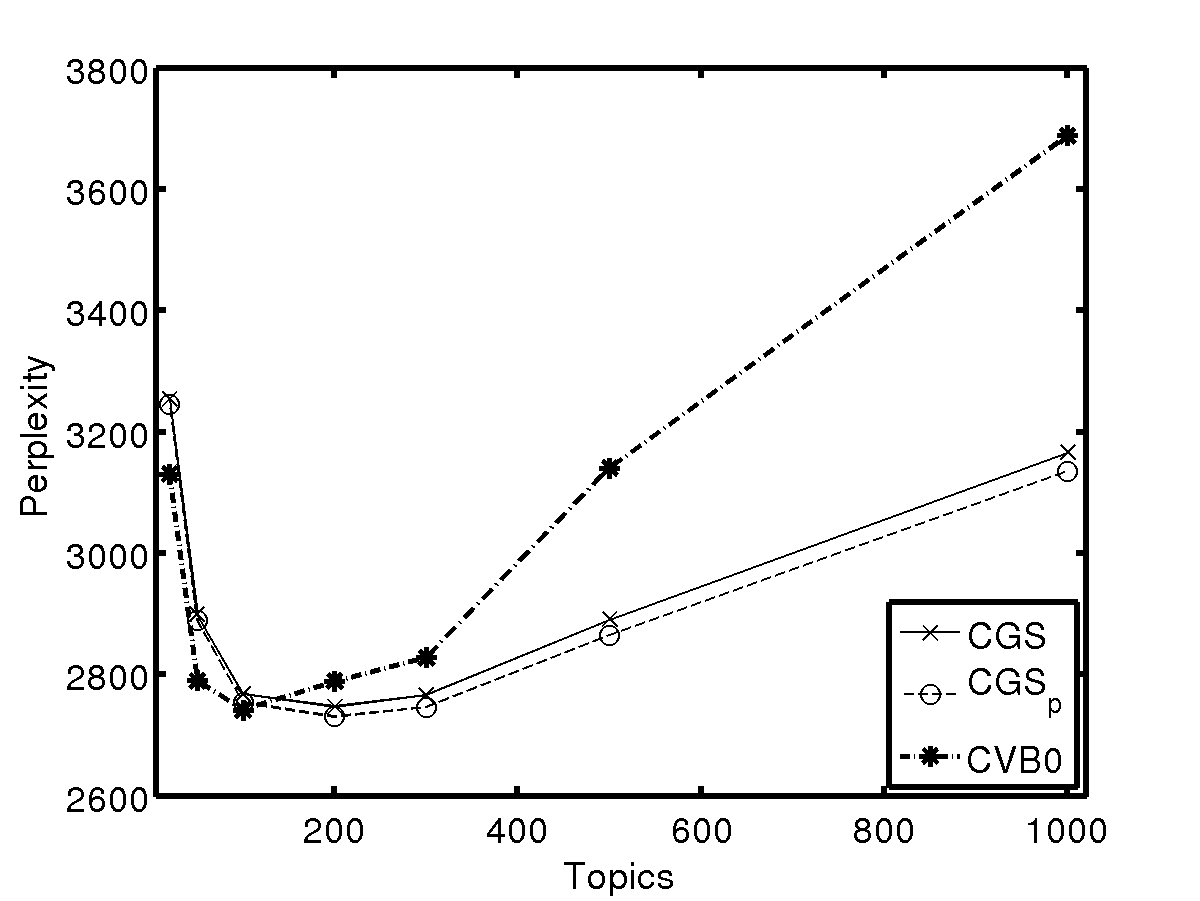}
  \subcaption{New York Times}
\end{minipage}
\begin{minipage}{0.49\textwidth}
  \includegraphics[width=\linewidth]{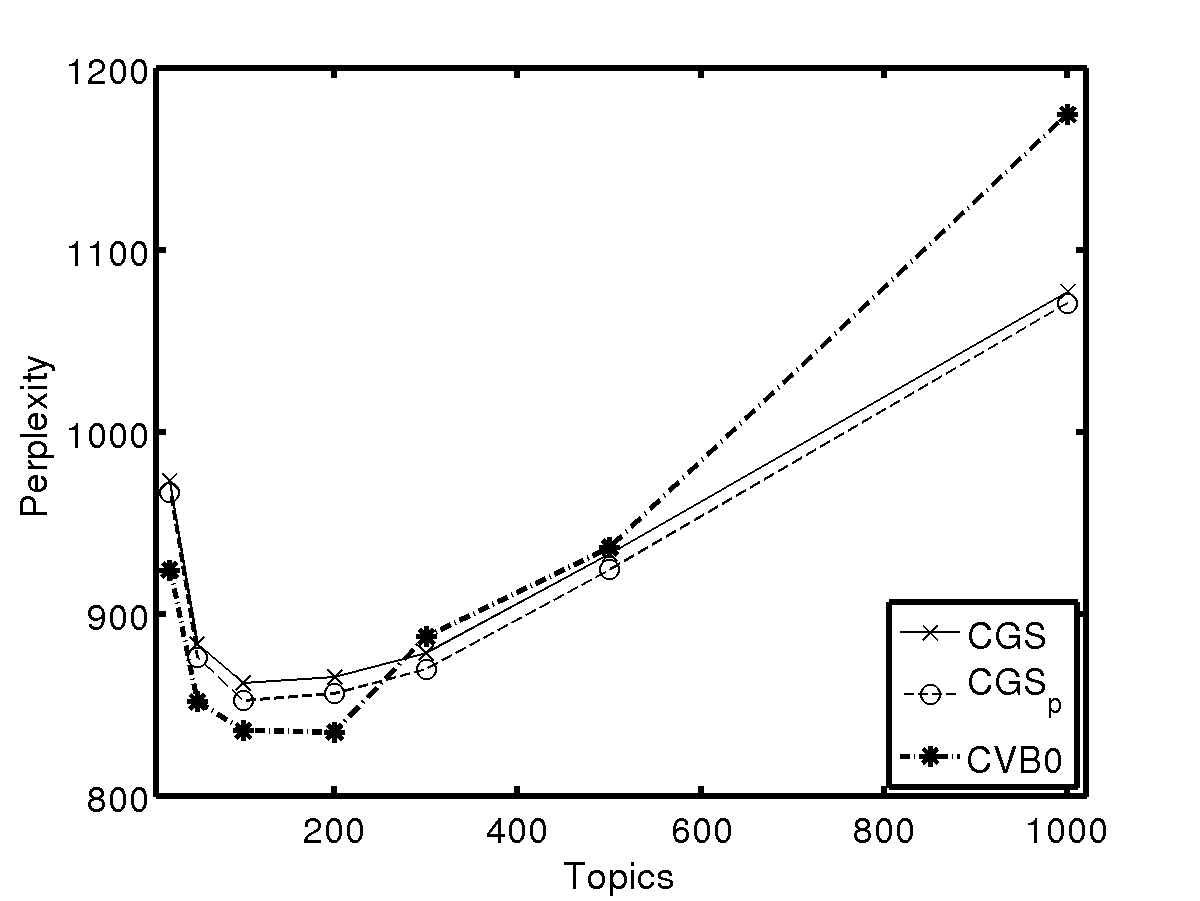}
  \subcaption{Reuters-21578}
\end{minipage}  
\hspace*{\fill}
\begin{minipage}{0.49\textwidth}
  \includegraphics[width=\linewidth]{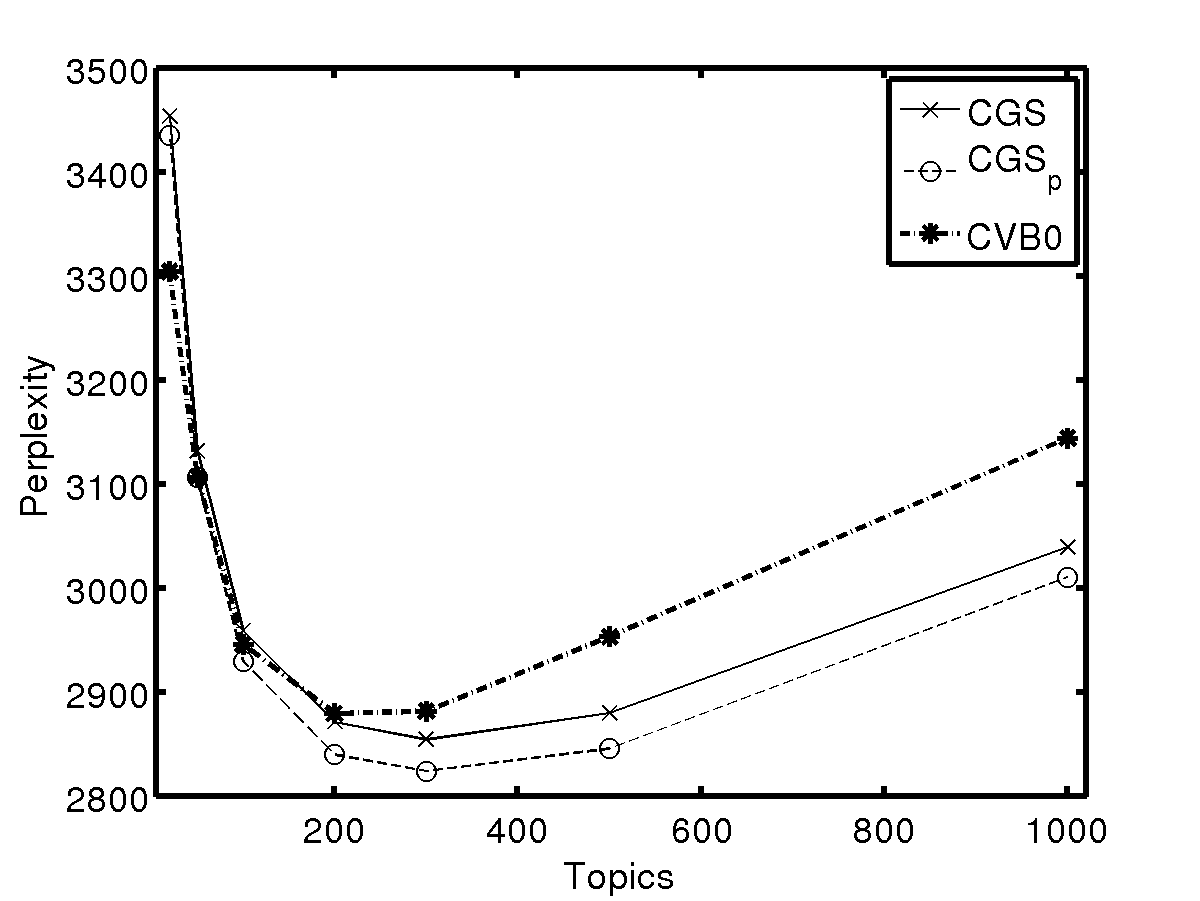}
  \subcaption{TASA}
\end{minipage}
\caption{Perplexity against number of topics for the CGS$_p$ method, standard CGS and CVB0. Results are taken by averaging over 5 different runs.}
\label{fig:ppx}
\end{figure}

In Appendix \ref{app:perplexityVB} we report also the results of this experiment, comparing the three aforementioned algorithms with Variational Bayes (VB). Since the results for VB were steadily worse, we excluded them from Figure \ref{fig:ppx}, to allow for easier comparison among CVB0, CGS and CGS$_p$.  
\citet{Asuncion:2009:SIT:1795114.1795118} note that the disadvantage of VB versus the other algorithms can potentially be mitigated via hyperparameter learning.

For this experiment we followed a similar approach to the one described by   \citet{Asuncion:2009:SIT:1795114.1795118}. During training we ran each chain for 200 iterations to obtain a single point estimate of the $\phi$ and $\phi^p$ distributions. During prediction, using the previously estimated $\phi$, we ran 200 iterations from one chain for the first half of each document to obtain an estimate of $\theta$. We followed the same approach to compute $\theta^p$, using $\phi^p$ in that case.\footnote{When referring to CGS$_p$, we mean that the $\phi^p +\theta^p$ estimators were used.} We then computed perplexity on the second half of each document. The $\alpha_k$ and $\beta_v$ hyperparameters were fixed across all data sets to uniform values, $0.1$ and $0.01$ respectively. CVB0 was run with the same parameterization. 

Figure \ref{fig:ppx} shows the perplexity results for all data sets across different settings (20, 50, 100, 200, 300, 500 and 1000 topics) for all methods. Each topics configuration was run for five times, obtaining the average perplexity value. There is a strong interaction between the perplexity scores, the data sets, and the number of topics, probably due to the diverse statistics of the data sets, such as the average document length and the average number of features (i.e. word types) per document (see Table \ref{tbl:unsupdatasets} in Section \ref{sec:datasets1}), that characterize the data sets. All algorithms achieve their lowest perplexity values at around 200 topics.

Despite the peculiarities of individual data sets, we can identify a broad general trend in these results. CVB0 outperforms (marginally in three out of four cases) the other methods in all data sets for lower topic number values: for the BioASQ and TASA data sets this happens up to 50 topics, for the New New York Times data set up to 100 topics, while for Reuters-21578 up to 200 topics. As the number of topics increases though, this behavior is reversed and CGS and CGS$_p$ have the upper hand after the aforementioned numbers of topics.

A possible explanation for this observation is related to the deterministic nature of CVB0; compared to its stochastic counterpart CGS, CVB0 is more prone to getting stuck in local maxima. As the number of topics increases, we expect the hypothesis space to grow bigger, making it more difficult for CVB0 to find a global optimum. CGS on the other hand, can exploit its stochastic nature to escape local maxima and converge to a better global representation of the data. Therefore, CVB0 seems to be better suited for configurations with a small number of topics (in which case the fact that CVB0 converges a lot faster than CGS as shown by \citet{Asuncion:2009:SIT:1795114.1795118} is an additional advantage), while CGS fits better in the opposite case. Similar to the previous section, when comparing CGS and CGS$_{p}$ we can observe a slight but steady advantage of our method over the standard estimator, across all experimental settings.

\subsection{Convergence During Estimation for CGS, CGS$_p$ and CVB0}
\label{sec:cgscvb0conv}

The perplexity results of the previous section motivated us to further investigate the convergence behavior of CGS, CGS$_p$ and CVB0 algorithms. In particular, we wanted to know the conditions under which CGS (and our modification of it) outperformed CVB0 and vice versa. Furthermore, we wanted to see how these differences evolved as the two algorithms converged upon their respective solutions.

\begin{figure}[bt]
\includegraphics[width=\linewidth]{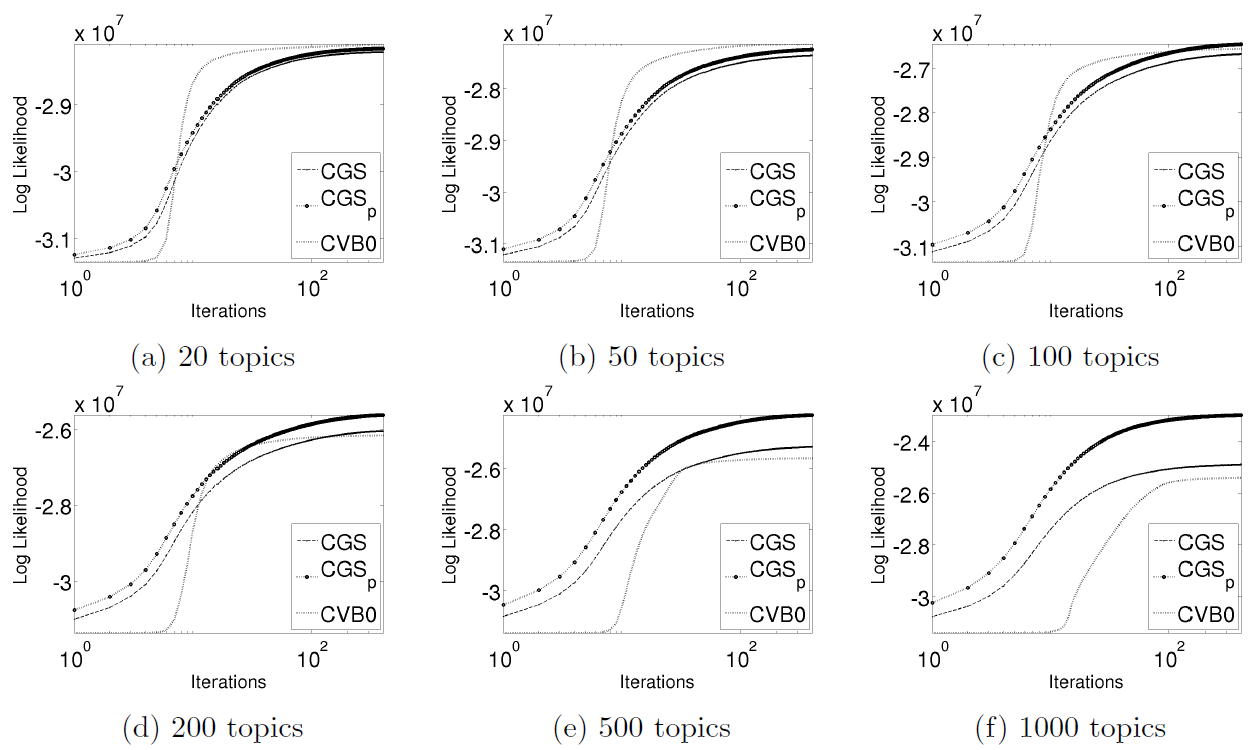}

\caption{CGS, CGS$_p$ and CVB0 convergence in terms of log likelihood during 400 training iterations in the TASA subset for different numbers of topics. $\alpha_k$ and $\beta_v$ were fixed to $0.1$ and $0.01$ respectively.}
\label{fig:ll}
\end{figure}

Figure \ref{fig:ll} presents a comparison of CVB0, CGS and CGS$_p$ (during the estimation phase) on the TASA data set in terms of their log likelihoods over iterations, under different numbers of topics. To calculate the log likelihood for CGS$_p$ we used the $\phi^p$ and $\theta^p$ estimators. For each algorithm and each topics configuration, we performed five runs and obtained the mean log-likelihood values. 

The results seem to validate our previous observations with CVB0 converging faster and to a higher log likelihood than CGS and CGS$_p$ when $|K|\leq 100$. As the number of topics grows and the number of training iterations increases, though, the performance of CGS and CGS$_p$ catches up, and after 100 topics, they outperform CVB0. This is consistent with our hypothesis that CVB0 performance is worse than that of CGS and CGS$_p$ as the hypothesis space grows, due to it being a deterministic algorithm and getting stuck in local maxima.
These results seem to be consistent with the observations made by \citet[Section~4]{DBLP:conf/nips/TehNW06} and could eventually emerge from either (or both) of the following factors:

\begin{itemize}
\item Variational inference methods (VB, CVB and CVB0) approximate the true posterior with a more tractable variational distribution \citep{Blei:2003:LDA:944919.944937,DBLP:conf/nips/TehNW06,Asuncion:2009:SIT:1795114.1795118}. Furthermore, in the case of CVB0, as illustrated by \citep[Section~4.4.1]{foulds2014latent}, the algorithm is a result of three consecutive approximations of the initial problem. On the other hand, CGS performs exact collapsed Gibbs updates and will eventually sample from the true posterior.
\item CVB0 seems to be unable to avoid local maxima due to its deterministic nature, especially as the hypothesis space grows bigger, while CGS takes advantage of its stochastic nature in order to avoid them.
\end{itemize}
 A secondary observation is that CGS$_p$ increases its advantage over CGS as the number of topics grows. A final noteworthy observation from Figure \ref{fig:ll} is that CVB0 exhibits very slow convergence behavior in the first 5 -- 10 iterations, before reaching a regime of rapid convergence.  While this phenomenon was not immediately visible in the results reported by \citet{Asuncion:2009:SIT:1795114.1795118} (Figure 2), this apparent discrepancy is readily explained: in those results a linear scale was used on the X-axis, with data points plotted only every 20 iterations.
\begin{table}[t]

\centering
\begin{tabular}{ccccc}
\toprule
\noalign{\smallskip}  
Method &\multicolumn{4}{c}{Data sets} \\
\midrule
\noalign{\smallskip}  
 & BioASQ & New York Times & Reuters-21578 & TASA\\
\midrule
&\multicolumn{4}{c}{MALLET} \\
\noalign{\smallskip}  
$\phi$ + $\theta$ &-1.921&-6.770&-0.590&-2.572\\
$\phi^p$ + $\theta$ &-1.912&-6.763&-0.585&-2.567\\
$\phi$ + $\theta^p$ &-1.911&-6.762&-0.584&-2.562\\
$\phi^p$ + $\theta^p$ &-1.903&-6.751&-0.580&-2.560\\
&\multicolumn{4}{c}{WarpLDA} \\
\noalign{\smallskip}  
$\phi$ + $\theta$ &-3.22&-7.092&-1.014&-2.730\\
$\phi^p$ + $\theta$ &-3.207&-7.081&-0.997&-2.724\\
$\phi$ + $\theta^p$ &-3.208&-7.079&-0.992&-2.718\\
$\phi^p$ + $\theta^p$ &-3.203&-7.074&-0.990&-2.716\\
\bottomrule
\end{tabular}
\caption{CGS$_p$ vs standard CGS estimator on MALLET and WarpLDA in terms of the Log Likelihood. All values are $\times 10^7$ and represent mean values across five runs.}
\label{tbl:malletWarp}
\end{table}
This slow convergence behavior is in alignment with the well-known convergence speed issues for the closely related EM algorithm, which can also be understood as a variational method. For EM, \citet{salakhutdinov2003optimization} showed that convergence is very slow when the missing data distributions are highly uncertain. This is likely to be the case in the initial stages of CVB0 and CGS, where the count histograms are relatively flat.  We hypothesize that the hard assignments of CGS allow the algorithm to escape the slow-converging high-entropy regime more rapidly than CVB0, leading to better convergence behavior in the early iterations, before the determinism of CVB0 once again gives it the upper hand (at least, with a small number of topics).

\subsection{CGS$_p$ with MALLET and WarpLDA}
\label{sec:malletWarpExp}
In order to exhibit how straightforward it is to plug our methods into already existing CGS LDA implementations and, most importantly, in modern CGS variants, we have employed CGS$_p$ on top of the MALLET software package \citep{mccallum2002mallet} and the WarpLDA \citep{chen2016warplda} implementation. In all cases, we have used $K=100$, $\alpha = 0.1$, $\beta=0.01$. After running the respective implementations on the four datasets for 200 iterations, we have obtained a single sample for the z-assignments. Subsequently, and after calculating the $n_{kw}$ and $n_{dk}$ counters, we have computed the standard estimator and our proposed estimators. Table \ref{tbl:malletWarp} depicts the results in terms of the training data log-likelihood, showing a steady advantage of our methods compared to the standard estimator.

\begin{figure}[tb]
\begin{minipage}{\textwidth}
\begin{minipage}{0.49\textwidth}
  \includegraphics[width=\linewidth]{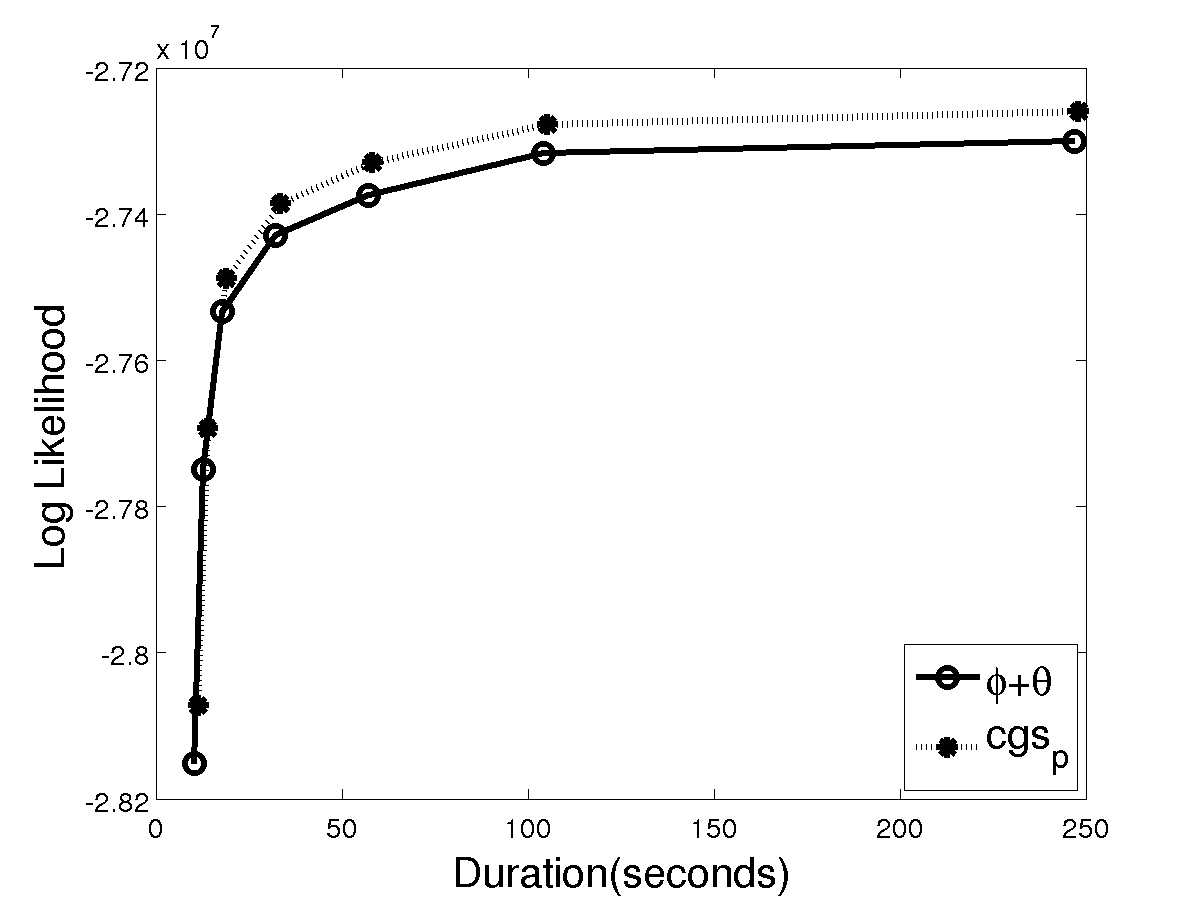}
  \subcaption{$K = 50$}
\end{minipage}  
\hspace*{\fill}
\begin{minipage}{0.49\textwidth}
  \includegraphics[width=\linewidth]{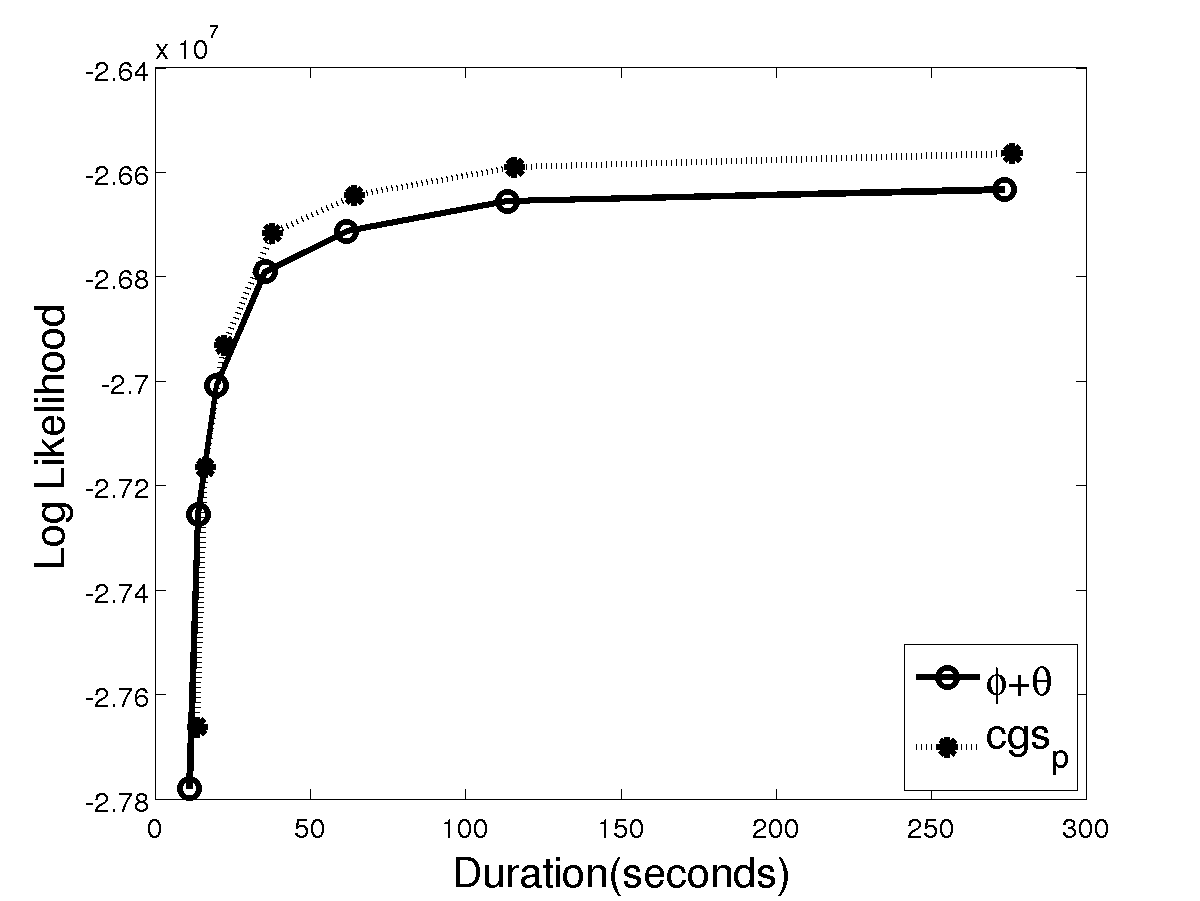}
  \subcaption{$K = 100$}
\end{minipage}
\end{minipage}
\begin{minipage}{\textwidth}
\begin{minipage}{0.49\textwidth}
  \includegraphics[width=\linewidth]{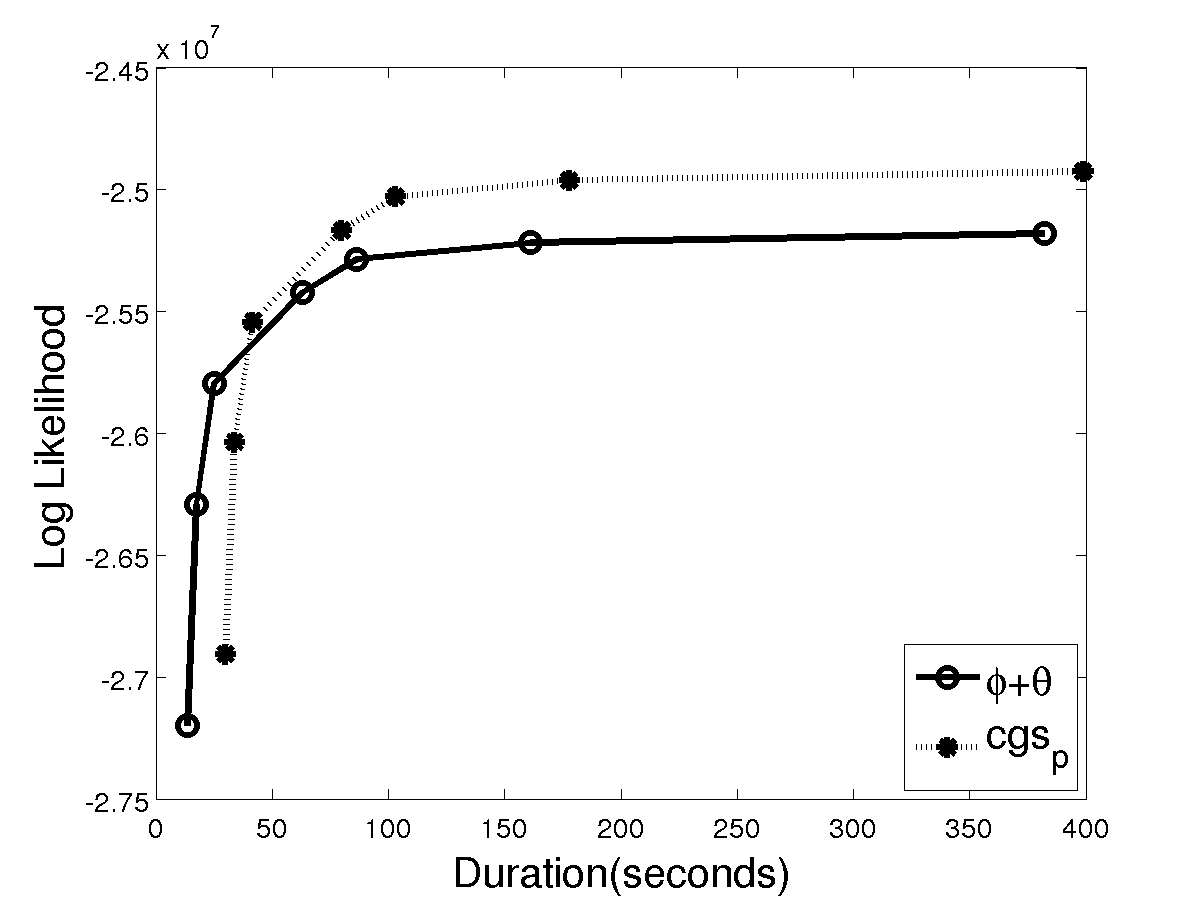}
  \subcaption{$K = 500$}
\end{minipage}  
\hspace*{\fill}
\begin{minipage}{0.49\textwidth}
  \includegraphics[width=\linewidth]{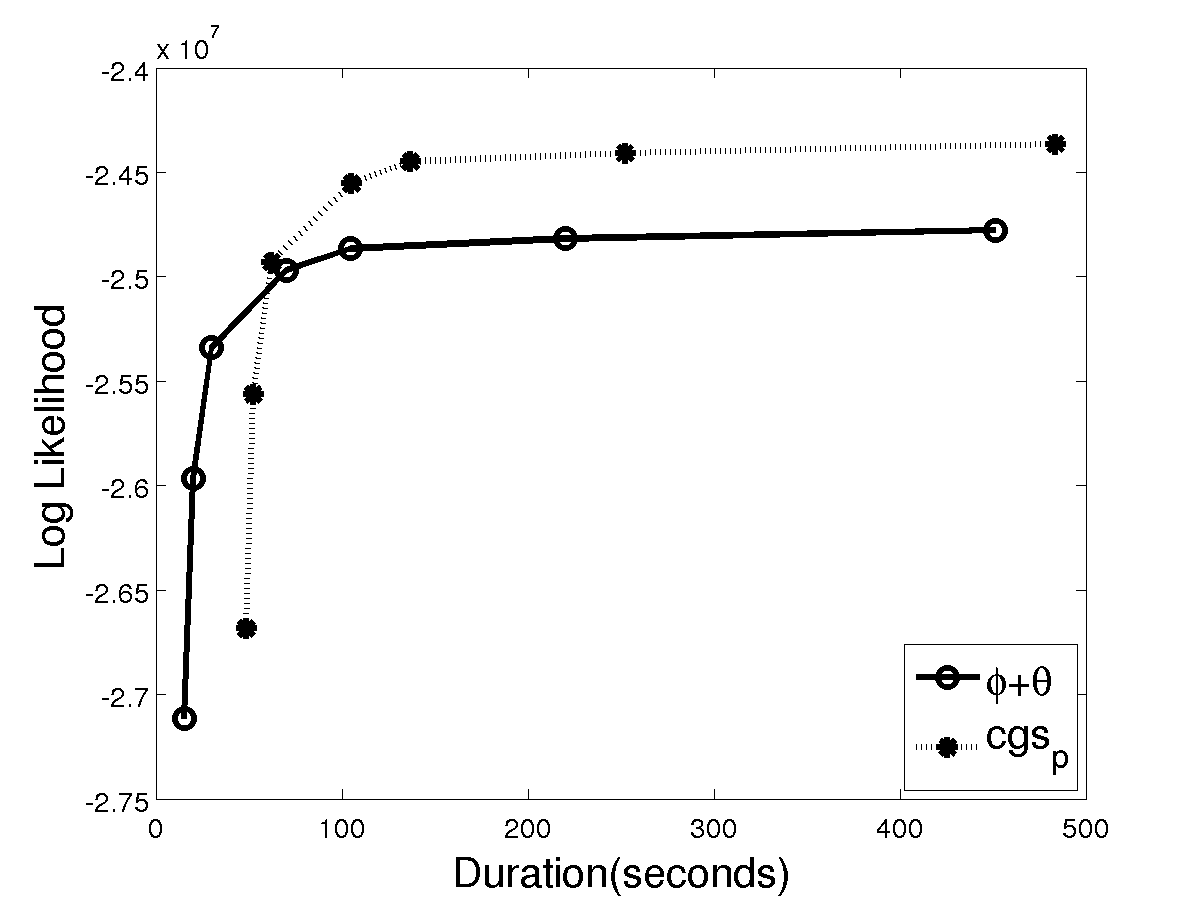}
  \subcaption{$K = 1000$}
\end{minipage}
\end{minipage}
\caption{Log Likelihood vs duration for CGS$_p$ against the standard CGS estimators on TASA, with WarpLDA. The time duration for each point in the plot corresponds to the running time of WarpLDA for different numbers of iterations (50, 100, 200, 500, 1000, 2000, 5000), plus the overhead needed to compute each of the estimators at that timestep.}
\label{fig:timeoverheadExp}
\end{figure}

\subsection{Run-time Overhead: CGS$_p$ vs CGS}
\label{sec:overhead}
In the previous experiments, we have observed that our proposed estimators steadily outperformed the standard CGS estimators. Nevertheless, our methods entail a computational time overhead since they require computing the full CGS probability distributions, which is roughly equivalent to a single dense standard CGS update.  This overhead is most noticeable when modern sparse CGS implementations are used with a large number of topics $K$, since these methods can be executed in time sublinear in $K$, unlike the dense CGS update. In this section, we study the trade-off between the improved performance and the extra cost of our methods, as compared with the standard estimators. In particular, we are interested in examining the performance of our methods when paired with the fastest known CGS implementation, WarpLDA.

In most applications of topic models, especially when dealing with large-scale data, LDA inference is run on a given corpus for a number of iterations and a single point estimate of the $\phi$ and $\theta$ parameters is calculated at the end of the procedure, so we focus on that scenario for our experiment.
%
When considering performance versus running time, based on our previous experiments we expect CGS$_p$ to attain higher log likelihood values, but with the extra overhead of the estimators calculation, while the standard CGS estimators will need more iterations to converge to those same log likelihood values, without the overhead. Therefore, we are interested to see if, and at which point during the process of training an LDA model, employing the CGS$_p$ estimators will be more beneficial time-wise than using the standard estimators.

In this experiment, reported in Figure \ref{fig:timeoverheadExp}, we run WarpLDA on the TASA data set for $K = 50, 100, 500, 1000$ and plot the training data log likelihood of CGS and CGS$_p$ against time duration. In Appendix \ref{app:timeoverhead} we report the relevant results for the rest of the data sets. The time duration for each point in the plot corresponds to the running time of WarpLDA for different numbers of iterations (50, 100, 200, 500, 1000, 2000, 5000), plus the overhead needed to compute each of the estimators at that timestep. We observe that in the early iterations, while LDA is still converging, the standard estimators attain a higher log likelihood value faster than the CGS$_p$ ones for $K=500, 1000$ (the two estimators are almost identical during burn-in for $K=50, 100$). This tendency is reversed after convergence of the procedure, with CGS$_p$ attaining a higher log likelihood value faster than CGS for all topic configurations. In most cases we are interested in obtaining an estimate of the LDA parameters well after convergence, so we consider that the CGS$_p$ estimators are to be preferred over the standard CGS ones, even under limited time resources.  We also see that convergence takes longer with more topics, and the time before CGS$_p$ overtakes CGS is correspondingly longer, but the improvement in log-likelihood for CGS$_p$ over CGS becomes greater.  

\section{Multi-Label Learning Experiments --- Prior-LDA}
\label{sec:multi}

Another important context for comparing different approaches to LDA prediction is in a multi-label, supervised setting. Here, we considered a multi-label learning extension of LDA---Prior-LDA---and used it as a basis for comparisons of model predictions. Following the same organization as the previous section, we present the data sets, the evaluation measures, the experimental setup, and lastly the results with a discussion of their implications.

\subsection{Data Sets}
\label{sec:datasets2}

In this series of experiments we kept the BioASQ data set and used three additional labeled ones: Delicious, Bibtex and Bookmarks. Table \ref{tbl:datasets} presents their main statistics. These data sets were chosen as representative of the diversity of real-world data, where there are often many labels, and sometimes not many word tokens per training instance.

\begin{table}[tb]

\centering
\begin{tabular}{cccp{1.1cm}cccc}
\toprule
\noalign{\smallskip}  
& \multicolumn{3}{c}{Documents} &\multicolumn{3}{c}{Labels} & \\
\midrule
\noalign{\smallskip}  
Data set & $D_{Train}$ & $D_{Test}$ & $\overline{N_d}$ & $K$ & $\overline{|\mathcal{K}_d|}$ & Avg. Freq. & $V$\\
\midrule
\noalign{\smallskip}  
Delicious &12,910&3,181&16.72&983&19.06&250.34&500\\
BioASQ &20,000&10,000&113.55&1,032&9.05&174.97&135,186\\
Bibtex &4,880&2,515&54.06&159&2.38&73.05&1,806\\
Bookmarks &70,285&17,570&124.26&208&2.03&682.90&2,136\\
\bottomrule
\end{tabular}
\caption{Statistics for the data sets used in the supervised learning experiments. Column ``Avg. Freq.'' refers to the average label frequency. All figures concerning labels and word types are given for the respective training sets.}
\label{tbl:datasets}
\end{table}

Delicious, Bibtex and Bookmarks\footnote{\url{http://mulan.sourceforge.net/datasets-mlc.html}}, are three widely used multi-label data sets \citep{katakis08,tsoumakas+etal:2008}. We did not perform any further preprocessing of these data sets. A notable aspect of Delicious is that it contains very few word tokens per instance both in the training and test sets, making accurate classification difficult. Concerning the BioASQ data set, we used the same corpus as for the unsupervised learning experiments and the same preprocessing procedure. We did not use the entire labelset of the original data set, filtering out labels appearing in less than 40 instances.

\subsection{Evaluation Metrics}
\label{eval}
We considered three widely-used performance measures: the micro-averaged F1 measure (Micro-F for brevity) and the macro-averaged F1 measure (Macro-F for brevity) and the example-based F1 measure \citep{tsoumakas+etal:2010b}. These measures are a weighted function of precision and recall, and emphasize the need for a model to perform well in terms of both of these underlying measures. The Macro-F score is the average of the F1-scores that are achieved across all labels, while the example-based F1 score is the respective average of the F1 score across all test documents. The Micro-F score is the average F1 score weighted by each label's frequency. Macro-F tends to emphasize performance on infrequent labels, while Micro-F tends to emphasize performance on frequent labels.

\subsection{Setup of Prior-LDA (CGS and CVB0) }
\label{sec:priorldasetup}
The Prior-LDA model takes into account the relative label frequencies in the corpus. Test documents are biased to assign words to more frequent labels by using a non-symmetric vector $\alpha_k$ on the $\theta$ distributions. 
During training we kept the $\alpha_k$ parameter symmetrical and set it to:

\begin{equation}
\alpha_k = \frac{50}{K}
\end{equation}

\noindent while during prediction we incorporated the label frequencies by setting it to:

\begin{equation}
\alpha_k = 50 \cdot \frac{f_k}{\sum{f_k}}+\frac{30}{K}
\end{equation} 

\noindent with $f_k$ standing for the frequency of label $k$. The $\beta_v$ parameter was set to $0.1$ across all data sets.
For all data sets, we calculated performance under two different configurations: 
\begin{itemize}
\item First, we wanted to emulate a scenario where only one sample can be afforded both for training and testing. In this case, during training we ran only one Markov chain and took a single estimate from the same chain for both $\phi$ and $\phi^p$. Similarly, during testing, we ran one chain with the estimated $\phi$ and one with the estimated $\phi^p$ to obtain a single estimate for $\theta$ and $\theta^p$ from each chain. In this manner, we obtained the predictions for all combinations of $\phi$ and $\theta$ estimators.

\item In the second configuration, we used 5 Markov chains and 30 samples from each chain, both during training and testing, following the same approach as above to obtain predictions from all four combinations of the $\phi$ and $\theta$ estimators. In this case, the motivation was to examine performance with full convergence of the relevant estimators.
\end{itemize}

In both training and prediction, we used a burn-in period of 50 iterations and a lag of 5 iterations between each sample. All samples from all chains were averaged to obtain the respective parameter estimates for each method.

For CVB0 we used the same setup as the one described above. However, as CVB0 is a deterministic algorithm that does not benefit greatly from averaging samples, in the second configuration each chain was initialized with a different random order of the documents and a different random initialization of the $\gamma$ parameters.

Finally, in order to obtain a hard assignment of labels to documents from the $\theta$ distributions, we used the Meta-Labeler approach \citep{tang+etal:2009}. In particular, we used a linear regression model to predict the number of labels per instance. To train this model, the same feature space as for the LLDA models was employed and the LibLinear package \citep{Fan:2008:LLL:1390681.1442794} was used for the implementation.

\begin{table}[t]
\centering
\begin{small}
\begin{tabular}{lccccc}
\toprule
\noalign{\smallskip}  
Method&Delicious&BioASQ&Bibtex &Bookmarks &Avg Rank\\
\midrule
\noalign{\smallskip}  
\multicolumn{6}{c}{1 MC $\times$ 1 sample}\\
\midrule
\noalign{\smallskip}  
CGS&0.2126$\pm$0.0034$\triangle$&0.3075$\pm$0.0041$\triangle$&0.2388$\pm$0.0024$\triangle$&0.1546$\pm$0.0042$\triangle$&4.5\\
$\phi+\theta^p$&0.2472$\pm$0.0038$\triangle$&0.3616$\pm$0.0037$\triangle$&\textbf{0.2985}$\pm$\textbf{0.0029}&0.1696$\pm$0.0036$\triangle$&2.5\\
$\phi^p+\theta$&0.2127$\pm$0.0027$\triangle$&0.3081$\pm$0.0038$\triangle$&0.2337$\pm$0.0026$\triangle$&0.1542$\pm$0.0046$\triangle$&4.5\\
CGS$_{p}$ &0.2531$\pm$0.0032$\triangle$&0.3623$\pm$0.0041$\triangle$&0.2900$\pm$0.0021&0.1712$\pm$0.0027$\triangle$&2.25\\
CVB0&\textbf{0.2761}$\pm$\textbf{0.0015}& \textbf{0.4122}$\pm$\textbf{0.0032}&0.2907$\pm$0.0014&\textbf{0.1810}$\pm$\textbf{0.0032}&1.25\\
\midrule
\noalign{\smallskip}  
\multicolumn{6}{c}{5 MC $\times$ 30 samples}\\
\midrule
\noalign{\smallskip}  
CGS&0.2699$\pm$0.0017$\triangle$&0.4509$\pm$0.0021$\triangle$&0.2873$\pm$0.0011$\triangle$&0.1819$\pm$0.0011$\triangle$&4.25\\
$\phi+\theta^p$&0.3184$\pm$0.0015&0.4628$\pm$0.0019&\textbf{0.3704}$\pm$\textbf{0.0013}&0.2035$\pm$0.0018&1.75\\
$\phi^p+\theta$&0.2699$\pm$0.0019$\triangle$&0.4509$\pm$0.0024$\triangle$&0.2842$\pm$0.0015$\triangle$&0.1820$\pm$0.0023$\triangle$&3.75\\
CGS$_{p}$&\textbf{0.3189}$\pm$\textbf{0.0012}&\textbf{0.4633}$\pm$\textbf{0.0009}&0.36277$\pm$0.0005&\textbf{0.2039}$\pm$\textbf{0.0024}&1.25\\
CVB0&0.2773$\pm$0.0008$\triangle$&0.4125$\pm$0.0006$\triangle$&0.2910$\pm$0.0002$\triangle$&0.1811$\pm$0.0004$\triangle$&4\\
\bottomrule
\end{tabular}
\end{small}
\caption{Results for the Micro-F1 measure comparing the Prior-LDA models. A $\triangle$ symbol represents statistically significant difference compared to the best performing model, at a level of $0.05$.}
\label{tbl:microF}
\end{table}

\subsection{Results and Discussion}

Tables \ref{tbl:microF} -- \ref{tbl:dmacroF} show the Micro-F, the Macro-F and the example-based F1 results respectively for all algorithms on the four data sets. We additionally show the average rank of each model, in terms of how it performs among the models on average across the four data sets. All results represent averages over five runs of the respective Gibbs sampler (or CVB0 algorithm). Also, to test statistical significance of the differences, we performed a one sample z-test for proportions at a significance level of $0.05$.   

First, let us consider the results for when only a single sample was used. A first remark is that CGS$_p$ and $\phi + \theta^p$ are significantly better than CGS and $\phi^p + \theta$ over all data sets and evaluation measures, confirming $\theta^p$'s superiority over $\theta$. Secondly, CVB0 has a clear advantage over the rest of the methods, in the majority of the data sets and evaluation measures. Only CGS$_p$ and $\phi + \theta^p$ manage to outperform it in three out of twelve cases (four data sets and three evaluation measures). 

At first glance, these results suggest that, when only one sample from one Markov chain can be afforded, CVB0 should be perhaps considered first, followed by CGS$_p$ and $\phi + \theta^p$ as alternatives. In order to more carefully evaluate the two algorithms for scenarios where computational time is a critical parameter, in Appendix \ref{app2} we performed a short investigation in order to study if and when CGS$_p$ performance surpasses that of CVB0. Specifically, we plotted performance (Micro-F, Macro-F, example-based F1) against the number of samples taken, for all four data sets. The results show that CGS$_p$ manages to outperform CVB0 in seven out of twelve cases, by taking only two samples from one Markov chain, while, after taking ten samples from one chain, CGS$_p$ outperforms CVB0 in eleven out of twelve cases. Considering that there exist a number of approaches for CGS that exploit sparsity to speed up LDA estimation and prediction \citep{Porteous:2008:FCG:1401890.1401960,yao2009efficient,li2014reducing}, while this is not the case for CVB0, we could claim that in many cases CGS$_p$ is equally fit or even superior to CVB0, in terms of total computational time.\footnote{Note that the stochastic variant SCVB0 \citep{Foulds:2013:SCV:2487575.2487697} is much more scalable than CGS in the amount of data, but less scalable in the number of topics due to dense updates.}

\begin{table}[t]
\centering
\begin{small}
\begin{tabular}{lccccc}
\toprule
\noalign{\smallskip}  
Method&Delicious&BioASQ&Bibtex&Bookmarks&Avg\\
&&&&&Rank\\
\midrule
\noalign{\smallskip}  
\multicolumn{6}{c}{1 MC $\times$ 1 sample}\\
\midrule
\noalign{\smallskip}  
CGS&0.0337$\pm$0.0012)$\triangle$&0.1767$\pm$0.0039$\triangle$&0.1582$\pm$0.0033$\triangle$&0.0928$\pm$0.0017$\triangle$&4.5\\
$\phi+\theta^p$&\textbf{0.0597}$\pm$\textbf{0.0020}&0.2160$\pm$0.0031$\triangle$&0.2019$\pm$0.0030$\triangle$&0.1009$\pm$0.0023$\triangle$&2.25\\
$\phi^p+\theta$&0.0332$\pm$0.0015$\triangle$&0.1769$\pm$0.0040$\triangle$&0.1548$\pm$0.0031$\triangle$&0.0930$\pm$0.0019$\triangle$&4.5\\
CGS$_{p}$&0.0585$\pm$0.0018&0.2168$\pm$0.0036$\triangle$&0.1946$\pm$0.0036$\triangle$&0.1015$\pm$0.0016$\triangle$&2.25\\
CVB0&0.0466$\pm$0.0010$\triangle$&\textbf{0.3039}$\pm$0.0014&\textbf{0.2232}$\pm$\textbf{0.0021}&\textbf{0.1169}$\pm$\textbf{0.0008}&1.5\\
\midrule
\noalign{\smallskip}  
\multicolumn{6}{c}{5 MC $\times$ 30 samples}\\
\midrule
\noalign{\smallskip}  
CGS&0.0373$\pm$0.0006$\triangle$&0.3011$\pm$0.0014$\triangle$&0.2144$\pm$0.0013$\triangle$&0.1167$\pm$0.0009$\triangle$&4.5\\
$\phi+\theta^p$&\textbf{0.0874}$\pm$\textbf{0.0004}&0.3138$\pm$0.0016&\textbf{0.2693}$\pm$\textbf{0.0011}&0.1294$\pm$0.0008&1.5\\
$\phi^p+\theta$&0.0362$\pm$0.0011$\triangle$&0.3016$\pm$0.0015$\triangle$&0.2086$\pm$0.0017$\triangle$&0.1168$\pm$0.0013$\triangle$&4.5\\
CGS$_{p}$&0.0855$\pm$0.0012&\textbf{0.3146}$\pm$\textbf{0.0016}&0.2579$\pm$0.0021&\textbf{0.1299}$\pm$\textbf{0.0011}&1.5\\
CVB0&0.0481$\pm$0.0003$\triangle$&0.3075$\pm$0.0005&0.2237$\pm$0.0008$\triangle$&0.1173$\pm$ 0.0006$\triangle$&3\\
\bottomrule
\end{tabular}
\end{small}
\caption{Results for the Macro-F measure. A $\triangle$ symbol represents statistically significant difference compared to the best performing model, at a level of $0.05$.}
\label{tbl:macroF}
\end{table}

Let us now focus on the scenario where we have averaged over multiple samples and multiple Markov chains to calculate predictions. In this case, model performance shifts in favor of the CGS-based predictions. Whereas the CGS algorithm benefits from averaging samples---due to the fact that each sample is ideally a draw from the posterior distribution---CVB0 is deterministic, and achieves very small improvements (if any), presumably since it tends to converge to the same specific maximum. When five chains are used, CGS$_p$ and $\phi + \theta^p$ are significantly better than CVB0 in all data sets and for all evaluation measures. CGS and $\phi^p + \theta$ on the other hand, seem to achieve equivalent performance to CVB0 overall. 

Among the CGS-based methods, we observe that, even if $\theta$ and $\theta^p$ should theoretically converge to the same solution (at least in an unsupervised learning context) even after a total of 150 samples (30 samples from five chains) their difference is significantly large for all evaluation measures and all data sets (the BioASQ data set represents partly an exception with smaller differences between the four methods).
Another important note concerns the comparison between models predicting with $\phi^p$ and those predicting with $\phi$. Results are mixed in this case, with models predicting with $\phi^p$ having a slight, but not statistically significant advantage.

\begin{table}[t]
\centering
\begin{small}
\begin{tabular}{lccccc}
\toprule
\noalign{\smallskip}  
Method & Delicious & BioASQ& Bibtex & Bookmarks &Avg Rank\\
\midrule
\noalign{\smallskip}  
\multicolumn{6}{c}{1 MC $\times$ 1 sample}\\
\midrule
\noalign{\smallskip}  
CGS&0.2047$\pm$0.0019$\triangle$&0.2930$\pm$0.0037$\triangle$&0.2546$\pm$0.0031$\triangle$&0.2007$\pm$0.0031$\triangle$&4.25\\
$\phi+\theta^p$&0.2389$\pm$0.0023$\triangle$&0.3400$\pm$0.0042&\textbf{0.3127}$\pm$\textbf{0.0036}&0.2152$\pm$0.0032$\triangle$&2.5\\
$\phi^p+\theta$&0.2035$\pm$0.0014$\triangle$&0.2942$\pm$0.0034$\triangle$&0.2472$\pm$0.0028$\triangle$&0.1999$\pm$0.0029$\triangle$&4.75\\
CGS$_{p}$&0.2444$\pm$0.0019$\triangle$&0.3402$\pm$0.0037&0.3048$\pm$0.0033&0.2166$\pm$0.0031$\triangle$&2\\
CVB0&\textbf{0.2672}$\pm$\textbf{0.0015}&\textbf{0.3504}$\pm$\textbf{0.0018}&0.2995$\pm$0.0017$\triangle$&\textbf{0.2225}$\pm$\textbf{0.0021}&1.5\\
\midrule
\noalign{\smallskip}  
\multicolumn{6}{c}{5 MC $\times$ 30 samples}\\
\midrule
\noalign{\smallskip}  
CGS&0.2625$\pm$0.0007$\triangle$&0.4180$\pm$0.0029&0.2958$\pm$0.0023$\triangle$&0.2226$\pm$0.0005$\triangle$&4.25\\
$\phi+\theta^p$&\textbf{0.3118}$\pm$\textbf{0.0011}&0.4269$\pm$0.0020&\textbf{0.3807}$\pm$\textbf{0.0011}&0.2424$\pm$0.0016&1.5\\
$\phi^p+\theta$&0.2610$\pm$0.0012$\triangle$&0.4181$\pm$0.0023&0.2922$\pm$0.0016$\triangle$&0.2229$\pm$0.0007$\triangle$&4\\
CGS$_{p}$&0.3086$\pm$0.0007&\textbf{0.4276}$\pm$\textbf{0.0025}&0.3733$\pm$0.0021&\textbf{0.2431}$\pm$\textbf{0.0009}&1.5\\
CVB0&0.2690$\pm$0.0004$\triangle$&0.3568$\pm$0.0021$\triangle$&0.2997$\pm$0.0005$\triangle$&0.2227$\pm$0.0003$\triangle$&3.75\\
\bottomrule
\end{tabular}
\end{small}
\caption{Results for the example-based F-measure. A $\triangle$ symbol represents statistically significant difference compared to the best performing model, at a level of $0.05$.}
\label{tbl:dmacroF}
\end{table}

Overall, the supervised experiments show a consistent advantage of the CGS$_p$ and $\phi+\theta^p$ estimators over the respective standard CGS estimator. Also, CGS$_p$ and $\phi+\theta^p$ are competitive with CVB0 in scenarios where only one sample can be afforded. Finally, when averaging over many samples and Markov chains, CGS$_p$ and $\phi+\theta^p$ distinctly outperform CVB0.

\section{Conclusions}
\label{sec:concl}

In this work we proposed a novel method for accurately estimating the document-wise and topic-wise LDA parameters from collapsed Gibbs samples by implicitly averaging over many samples, leading to improved estimation with little extra computational cost.
Our algorithm can be interpreted as using soft clustering information, as in the CVB0 collapsed variational inference algorithm, but in a Markov chain Monte Carlo setting.
This allows us to use all of the uncertainty information encoded in the dense Gibbs sampling transition probabilities to recover the parameters from a CGS sample, while still leveraging the sparsity inherent in the CGS algorithm during the training process.
We extensively investigated the performance of the proposed estimators for both unsupervised and supervised topic models, and with comparison to CVB0.
Our results demonstrate that our $\theta^p$ and $\phi^p$ estimators are more effective than the standard estimator in terms of perplexity, and that  $\theta^p$ also improves multi-label classification performance for the supervised Prior-LDA model.

Our experiments showed an advantage of our CGS-based estimators over CVB0 in the majority of cases, and also revealed the consequences of the deterministic nature of that algorithm. In the unsupervised scenario, CVB0 has a clear advantage for LDA models with few topics, while this phenomenon is reversed as the posterior landscape becomes more complicated with a greater number of topics. In the multi-label learning case, CVB0 has the upper hand when only one sample can be afforded, while it is decisively outperformed by CGS$_p$ when averaging over multiple samples.  
The success of our methods illustrates the value of averaging over multiple MCMC samples for LDA, even when only a point estimate is required. In future work, we anticipate that our ideas can be adapted to other models with collapsed representations, including latent variable blockmodels for social networks and other network data \citep{kemp2006learning}, and  hidden Markov models \citep{goldwater2007fully}.

\acks{The authors would like to thank the BioASQ team for kindly providing the BioASQ corpus that was used throughout the experiments. Furthermore, we would like to thank the anonymous reviewers of the previous versions of this paper for their elaborate and constructive comments that greatly helped in improving and extending this work. Yannis Papanikolaou and Grigorios Tsoumakas have been partially funded by Atypon Systems Inc., Santa Clara, California, USA.}

\appendix

\begin{figure}[tb]
\begin{minipage}{\textwidth}
	\begin{minipage}{0.48\textwidth}
   		\includegraphics[width=\linewidth]{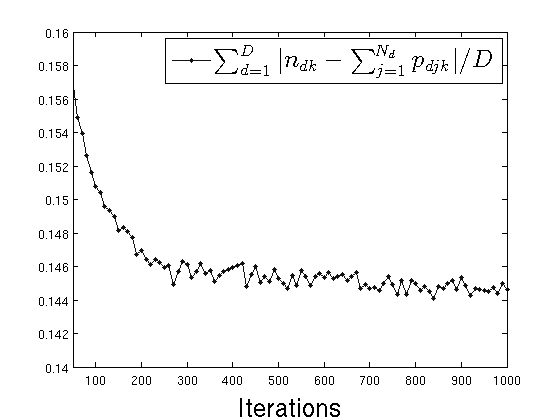}
   		\subcaption{TASA}
   \end{minipage} 
	\begin{minipage}{0.48\textwidth}
   		\includegraphics[width=\linewidth]{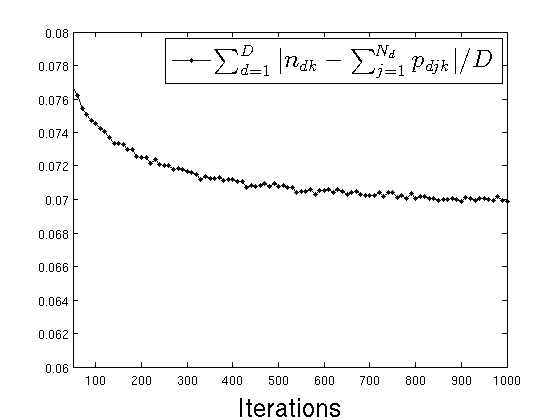}
   		\subcaption{New York Times}
 	\end{minipage}  
\end{minipage}

\begin{minipage}{\textwidth}
	\begin{minipage}{0.48\textwidth}
   		\includegraphics[width=\linewidth]{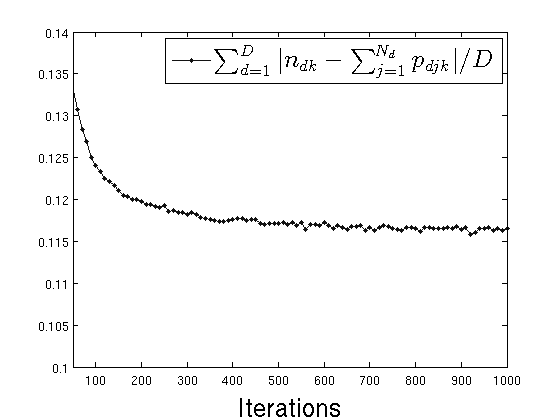}
   		\subcaption{BioASQ}
   \end{minipage} 
	\begin{minipage}{0.48\textwidth}
   		\includegraphics[width=\linewidth]{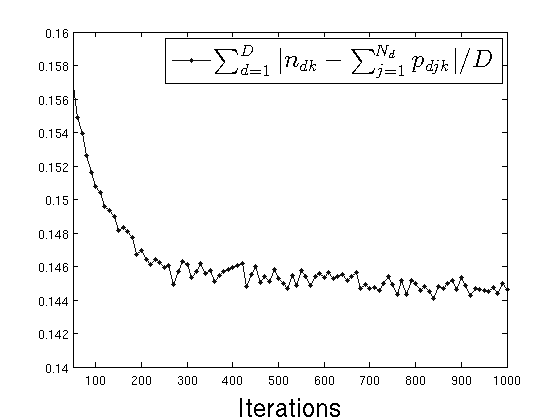}
   		\subcaption{Reuters-21578}
 	\end{minipage}  
\end{minipage}

\caption{Hard and soft document-topic counts absolute difference against the number of iterations. The counts are normalized to 1 before taking their difference, hence the difference taking values between 0 and 1.}
\label{fig:hardvssoftcounts}
\end{figure}

\section{Hard Counts vs Soft Counts}
Algorithmically (refer to Algorithm \ref{alg:methods}), the difference between the $\theta$ and $\theta^p$ estimators lies in the difference between the hard counts $n_{dk}$ and the soft counts given by $\sum_{j=1}^{N_d} p(d, j, k)$, p being the sampling update equation of CGS given in equation \ref{eq:p}. In this experiment, we examine how similar are the two counts. For this reason, we have run CGS on the four data sets setting $K=100$, $\alpha = 0.1$, $\beta=0.01$ for 50 iterations and calculated the sum of the absolute differences between the two counts, or, in other words:
\begin{align}
\frac{\sum_{d=1}^D \sum_{k=1}^K |n_{dk} - \sum_{j=1}^{N_d} p(d, j, k)|}{D}
\end{align}

\begin{figure}[tb]
 \begin{minipage}{0.48\textwidth}
   \includegraphics[width=\linewidth]{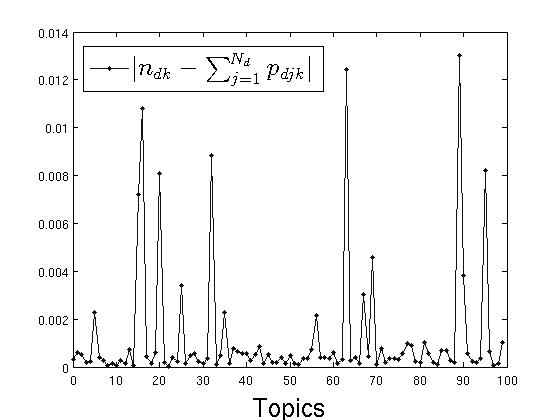}
   \subcaption{$iteration$ = 50}
 \end{minipage}  
 \hspace*{\fill}
 \begin{minipage}{0.48\textwidth}
   \includegraphics[width=\linewidth]{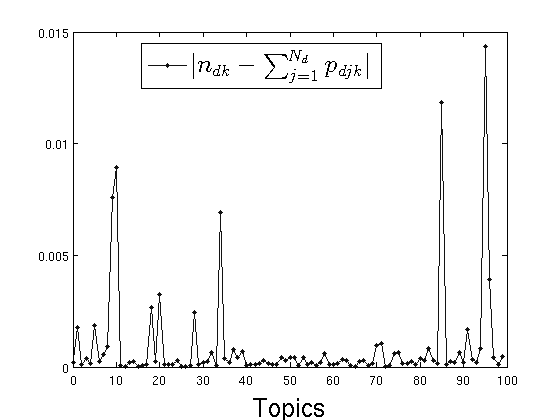}
   \subcaption{$iteration$ = 100}
 \end{minipage}
  \hspace*{\fill}
 \begin{minipage}{0.48\textwidth}
   \includegraphics[width=\linewidth]{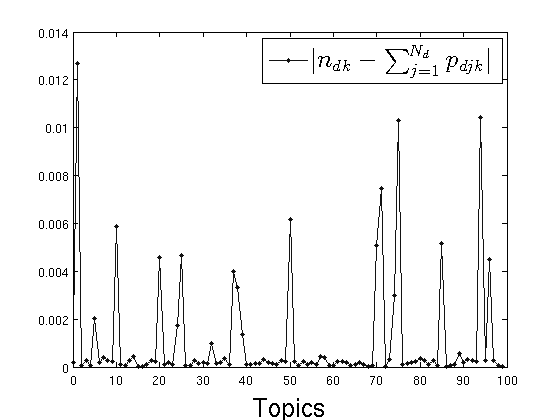}
   \subcaption{$iteration$ = 500}
 \end{minipage}
\hspace*{\fill} 
\begin{minipage}{0.48\textwidth}
   \includegraphics[width=\linewidth]{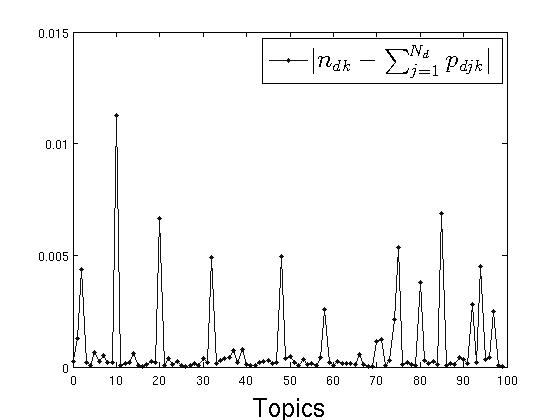}
   \subcaption{$iteration$ = 1000}
\end{minipage} 

\caption{Absolute difference between the hard and the soft counts across the K different topics, for iteration 50, 100 and 1000, for the first document of the TASA collection.}
\label{fig:hardvssoftcountsExamples}
\end{figure}

 Figure \ref{fig:hardvssoftcounts} shows the absolute sum of differences across the number of iterations. We can observe a steady difference among the two counts, that diminishes but does not go away with the number of iterations.
Furthermore, in Figure \ref{fig:hardvssoftcountsExamples} we depict for the TASA data set and the first document of the collection, the absolute difference between the hard and the soft counts across the K different topics, for iteration 50, 100 and 1000. Similarly to our previous observations, we see that the two counts are similar, but present a small but steady difference across topics, for all iterations.

\section{Bounding the Approximation for $\phi_p$}
\label{sec:boundPhiPApprox}
We wish to bound the error of the approximation step used in the derivation for the $\phi_p$ technique, given in Equation \ref{eq:approximationphip2}:
\begin{align}
&E_{p(\mathbf{z}| \mathbf{w}, \boldsymbol{\beta}, \mathbf{z} \in Z_{Gibbs}(\mathbf{z}^{(i)}))}\Big [\frac{\sum_{d=1}^{D_{Train}}\sum_{j:w_{d,j} = v} z_{djk} + \beta_v}{ \sum_{d=1}^{D_{Train}}\sum_{j=1}^{N_d} z_{djk} + \sum_{v^\prime = 1}^V\beta_{v^\prime}} \Big ] \nonumber \\
\approx & E_{p(\mathbf{z}| \mathbf{w}, \boldsymbol{\beta}, \mathbf{z} \in Z_{Gibbs}(\mathbf{z}^{(i)}))}\Big [\frac{\sum_{d=1}^{D_{Train}}\sum_{j:w_{d,j} = v} z_{djk} + \beta_v}{ \sum_{d=1}^{D_{Train}}\sum_{j=1}^{N_d} z_{djk}^{(i)} + \sum_{v^\prime = 1}^V\beta_{v^\prime}} \Big ] \nonumber \mbox{ .}
\end{align}

Assume that all $n_k^{(i)} > 0$.  Then, starting from the left hand side of the above,  we have:
\begin{align}
 & E_{p(\mathbf{z}| \mathbf{w}, \boldsymbol{\beta}, \mathbf{z} \in Z_{Gibbs}(\mathbf{z}^{(i)}))}\Big [\frac{\sum_{d=1}^{D_{Train}}\sum_{j:w_{d,j} = v} z_{djk} + \beta_v}{ \sum_{d=1}^{D_{Train}}\sum_{j=1}^{N_d} z_{djk} + \sum_{v^\prime = 1}^V\beta_{v^\prime}} \Big ] \nonumber \\
=& E_{p(\mathbf{z}| \mathbf{w}, \boldsymbol{\beta}, \mathbf{z} \in Z_{Gibbs}(\mathbf{z}^{(i)}))}\Big [\frac{\sum_{d=1}^{D_{Train}}\sum_{j:w_{d,j} = v} z_{djk} + \beta_v}{ \sum_{d=1}^{D_{Train}}\sum_{j=1}^{N_d} z_{djk}^{(i)} + \sum_{v^\prime = 1}^V\beta_{v^\prime}} \times \frac{ \sum_{d=1}^{D_{Train}}\sum_{j=1}^{N_d} z_{djk}^{(i)} + \sum_{v^\prime = 1}^V\beta_{v^\prime}}{ \sum_{d=1}^{D_{Train}}\sum_{j=1}^{N_d} z_{djk} + \sum_{v^\prime = 1}^V\beta_{v^\prime}} \Big ] \nonumber \\
=& \sum_{\mathbf{z} \in Z_{Gibbs}(\mathbf{z}^{(i)})}  \frac{\sum_{d=1}^{D_{Train}}\sum_{j:w_{d,j} = v} z_{djk} + \beta_v}{ \sum_{d=1}^{D_{Train}}\sum_{j=1}^{N_d} z_{djk}^{(i)} + \sum_{v^\prime = 1}^V\beta_{v^\prime}} \times \frac{ \sum_{d=1}^{D_{Train}}\sum_{j=1}^{N_d} z_{djk}^{(i)} + \sum_{v^\prime = 1}^V\beta_{v^\prime}}{ \sum_{d=1}^{D_{Train}}\sum_{j=1}^{N_d} z_{djk} + \sum_{v^\prime = 1}^V\beta_{v^\prime}} \nonumber \\
& \ \ \ \ \ \ \ \ \ \ \ \ \ \ \ \times p(\mathbf{z}| \mathbf{w}, \boldsymbol{\beta}, \mathbf{z} \in Z_{Gibbs}(\mathbf{z}^{(i)}))  \nonumber \\
\geq& \sum_{\mathbf{z} \in Z_{Gibbs}(\mathbf{z}^{(i)})}  \frac{\sum_{d=1}^{D_{Train}}\sum_{j:w_{d,j} = v} z_{djk} + \beta_v}{ \sum_{d=1}^{D_{Train}}\sum_{j=1}^{N_d} z_{djk}^{(i)} + \sum_{v^\prime = 1}^V\beta_{v^\prime}} \times \frac{ \sum_{d=1}^{D_{Train}}\sum_{j=1}^{N_d} z_{djk}^{(i)} + \sum_{v^\prime = 1}^V\beta_{v^\prime}}{ \sum_{d=1}^{D_{Train}}\sum_{j=1}^{N_d} z_{djk}^{(i)} + 1 + \sum_{v^\prime = 1}^V\beta_{v^\prime}} \nonumber \\
& \ \ \ \ \ \ \ \ \ \ \ \ \ \ \ \times p(\mathbf{z}| \mathbf{w}, \boldsymbol{\beta}, \mathbf{z} \in Z_{Gibbs}(\mathbf{z}^{(i)}))  \nonumber \\ 
=& \frac{ \sum_{d=1}^{D_{Train}}\sum_{j=1}^{N_d} z_{djk}^{(i)} + \sum_{v^\prime = 1}^V\beta_{v^\prime}}{ \sum_{d=1}^{D_{Train}}\sum_{j=1}^{N_d} z_{djk}^{(i)} + 1 + \sum_{v^\prime = 1}^V\beta_{v^\prime}}E_{p(\mathbf{z}| \mathbf{w}, \boldsymbol{\beta}, \mathbf{z} \in Z_{Gibbs}(\mathbf{z}^{(i)}))}\Big [\frac{\sum_{d=1}^{D_{Train}}\sum_{j:w_{d,j} = v} z_{djk} + \beta_v}{ \sum_{d=1}^{D_{Train}}\sum_{j=1}^{N_d} z_{djk}^{(i)} + \sum_{v^\prime = 1}^V\beta_{v^\prime}} \Big ] \nonumber\\
=& \frac{n_k^{(i)} + \sum_{v^\prime = 1}^V\beta_{v^\prime}}{n_k^{(i)} + \sum_{v^\prime = 1}^V\beta_{v^\prime} + 1} E_{p(\mathbf{z}| \mathbf{w}, \boldsymbol{\beta}, \mathbf{z} \in Z_{Gibbs}(\mathbf{z}^{(i)}))}\Big [\frac{\sum_{d=1}^{D_{Train}}\sum_{j:w_{d,j} = v} z_{djk} + \beta_v}{ \sum_{d=1}^{D_{Train}}\sum_{j=1}^{N_d} z_{djk}^{(i)} + \sum_{v^\prime = 1}^V\beta_{v^\prime}} \Big ] \mbox{ .}
\end{align}
Similarly, for a lower bound, we have:
\begin{align}
 & E_{p(\mathbf{z}| \mathbf{w}, \boldsymbol{\beta}, \mathbf{z} \in Z_{Gibbs}(\mathbf{z}^{(i)}))}\Big [\frac{\sum_{d=1}^{D_{Train}}\sum_{j:w_{d,j} = v} z_{djk} + \beta_v}{ \sum_{d=1}^{D_{Train}}\sum_{j=1}^{N_d} z_{djk} + \sum_{v^\prime = 1}^V\beta_{v^\prime}} \Big ] \nonumber \\
\leq& \sum_{\mathbf{z} \in Z_{Gibbs}(\mathbf{z}^{(i)})}  \frac{\sum_{d=1}^{D_{Train}}\sum_{j:w_{d,j} = v} z_{djk} + \beta_v}{ \sum_{d=1}^{D_{Train}}\sum_{j=1}^{N_d} z_{djk}^{(i)} + \sum_{v^\prime = 1}^V\beta_{v^\prime}} \times \frac{ \sum_{d=1}^{D_{Train}}\sum_{j=1}^{N_d} z_{djk}^{(i)} + \sum_{v^\prime = 1}^V\beta_{v^\prime}}{ \sum_{d=1}^{D_{Train}}\sum_{j=1}^{N_d} z_{djk}^{(i)} - 1 + \sum_{v^\prime = 1}^V\beta_{v^\prime}}\nonumber \\ 
& \ \ \ \ \ \ \ \ \ \ \ \ \ \ \ \times p(\mathbf{z}| \mathbf{w}, \boldsymbol{\beta}, \mathbf{z} \in Z_{Gibbs}(\mathbf{z}^{(i)}))  \nonumber \\ 
=& \frac{n_k^{(i)} + \sum_{v^\prime = 1}^V\beta_{v^\prime}}{n_k^{(i)} + \sum_{v^\prime = 1}^V\beta_{v^\prime} - 1}E_{p(\mathbf{z}| \mathbf{w}, \boldsymbol{\beta}, \mathbf{z} \in Z_{Gibbs}(\mathbf{z}^{(i)}))}\Big [\frac{\sum_{d=1}^{D_{Train}}\sum_{j:w_{d,j} = v} z_{djk} + \beta_v}{ \sum_{d=1}^{D_{Train}}\sum_{j=1}^{N_d} z_{djk}^{(i)} + \sum_{v^\prime = 1}^V\beta_{v^\prime}} \Big ] \mbox{ .}
\end{align}

\section{Empirical validation of the approximation for $\phi^p$}
\label{sec:validateEq47}

When deriving our $\phi^p$ estimator, in order to prove that $\phi^p$ represents the expected value of the standard CGS $\phi$ estimator, we take the following approximation: we consider that the denominator of Equation \ref{eq:phiExpectation} is not changing, between two adjacent Gibbs samples. Since $z_{djk}$ is within $\pm 1$ of $z^{(i)}_{djk}$ when taking adjacent Gibbs samples and in a typical corpus $n_k >>1$, we can assume that $n_k \simeq n_k\pm 1$. Using this approximation, we are able to consider that the denominator in the left hand of Equation \ref{eq:approximationphip2} is a constant when taking an adjacent sample. Hence, the expected value of the fraction will be the expected value of the numerator, divided by the denominator.

In order to empirically validate this argument, we considered the following experiment: We have run CGS on the New York Times data set for 100 iterations and for $K=100$. Next we computed explicitly, the left and the right sides of Equation \ref{eq:approximationphip2} for all topics and word types. 
Subsequently, we have measured the accuracy of the approximation of Equation \ref{eq:approximationphip2} for a varying size of documents (and therefore $n_k$), by plotting the absolute difference $|$left hand side - right hand side$|$ for each topic, as a function of the size of $D$, or $\frac{|a-b|}{K}$ where

\begin{align}
\label{eq:approxdiff}
a = E_{p(\mathbf{z}| \mathbf{w}, \boldsymbol{\beta}, \mathbf{z} \in Z_{Gibbs}(\mathbf{z}^{(i)}))}\Big [\frac{\sum_{d=1}^{D_{Train}}\sum_{j:w_{d,j} = v} z_{djk} + \beta_v}{ \sum_{d=1}^{D_{Train}}\sum_{j=1}^{N_d} z_{djk} + \sum_{v^\prime = 1}^V\beta_{v^\prime}} \Big ] \nonumber 
\\
b = \frac{\sum_{d=1}^{D_{Train}}\sum_{j:w_{d,j} = v} E_{p(\mathbf{z}| \mathbf{w}, \boldsymbol{\beta}, \mathbf{z} \in Z_{Gibbs}(\mathbf{z}^{(i)})) }[z_{djk}] + \beta_v}{ \sum_{d=1}^{D_{Train}}\sum_{j=1}^{N_d} z_{djk}^{(i)} +  \sum_{v^\prime = 1}^V\beta_{v^\prime}}
\end{align}

\begin{figure}[tb]
   \includegraphics[width=\linewidth]{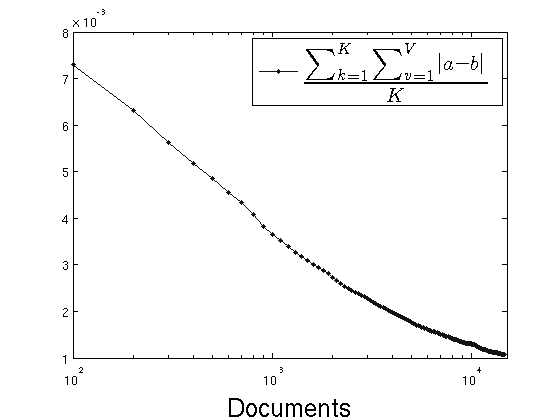}
\caption{Absolute difference, per topic, between the left and right hand of Equation \ref{eq:approximationphip2}. As the number of documents increases, the difference approaches 0. Here, $a$, $b$ are given in Equation \ref{eq:approxdiff} and the sum is divided by $K$ in order to obtain the per-topic differences.}
\label{fig:phipApprox}
\end{figure}

Figure \ref{fig:phipApprox} plots the results from the above experiment. As expected, by increasing the considered subset of documents and therefore $n_k$, the error of the approximation for each topic approaches zero.

\section{}
\label{app1}
\begin{figure}[tb]
\begin{minipage}{\textwidth}
 \begin{minipage}{0.49\textwidth}
   \includegraphics[width=\linewidth]{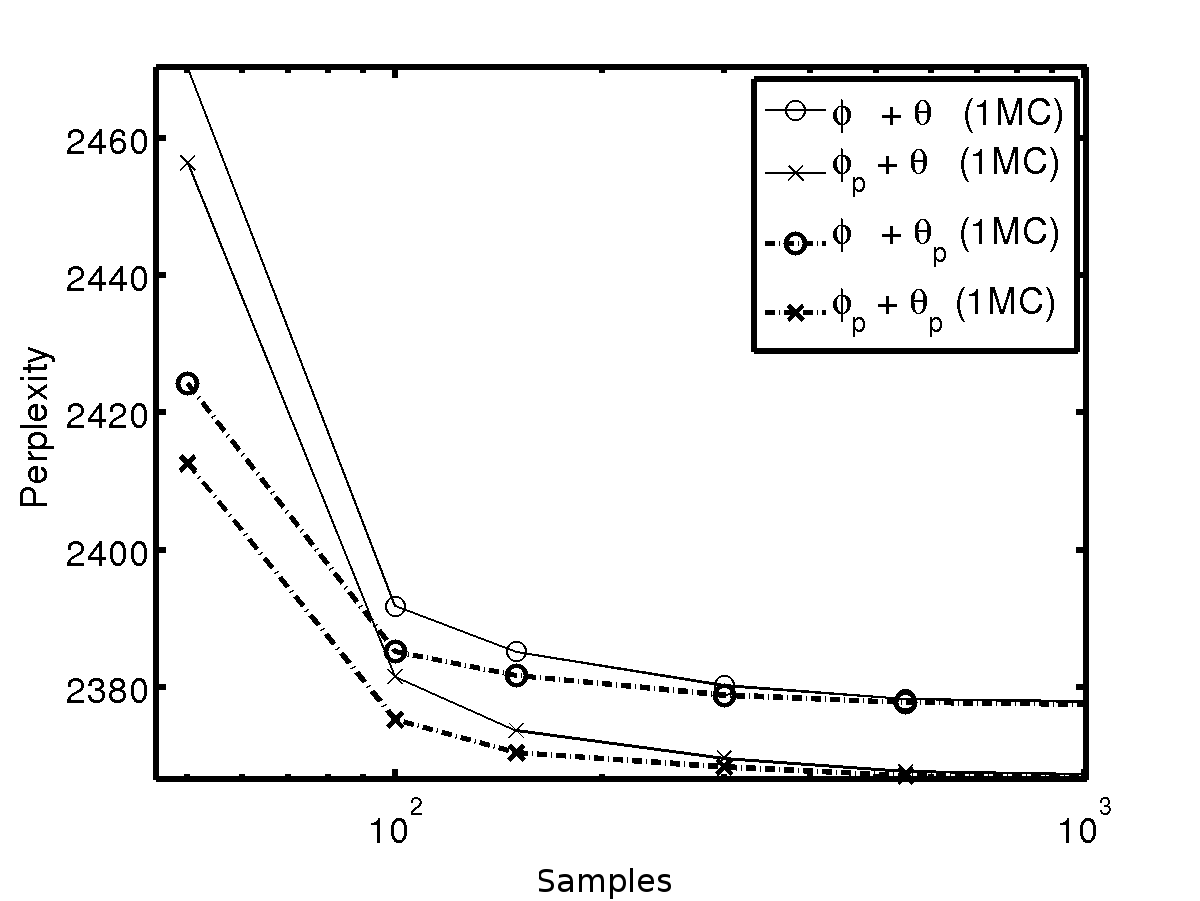}
 \end{minipage}  
 \hspace*{\fill}
 \begin{minipage}{0.49\textwidth}
   \includegraphics[width=\linewidth]{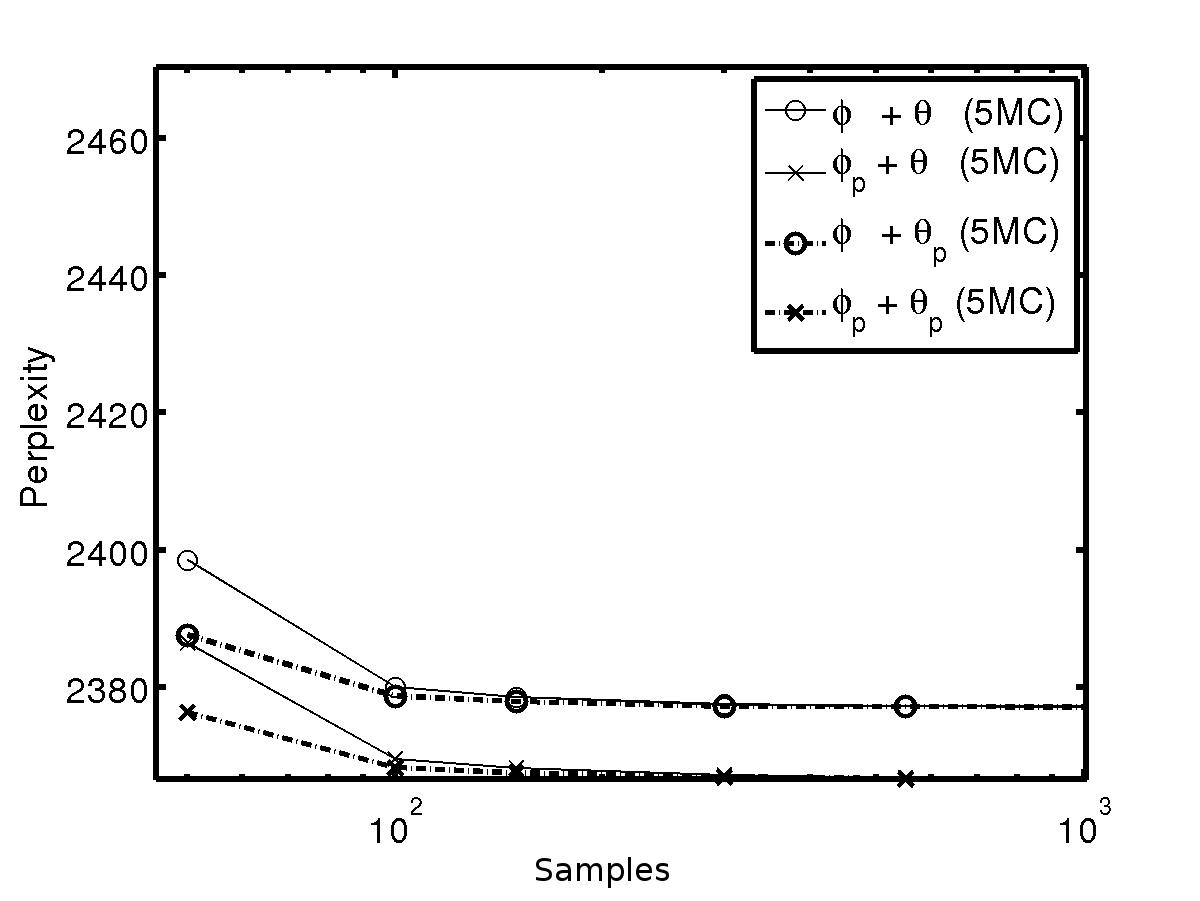}
 \end{minipage}
\subcaption{$K$ = 20}
\end{minipage}

\begin{minipage}{\textwidth}
\begin{minipage}{0.49\textwidth}
  \includegraphics[width=\linewidth]{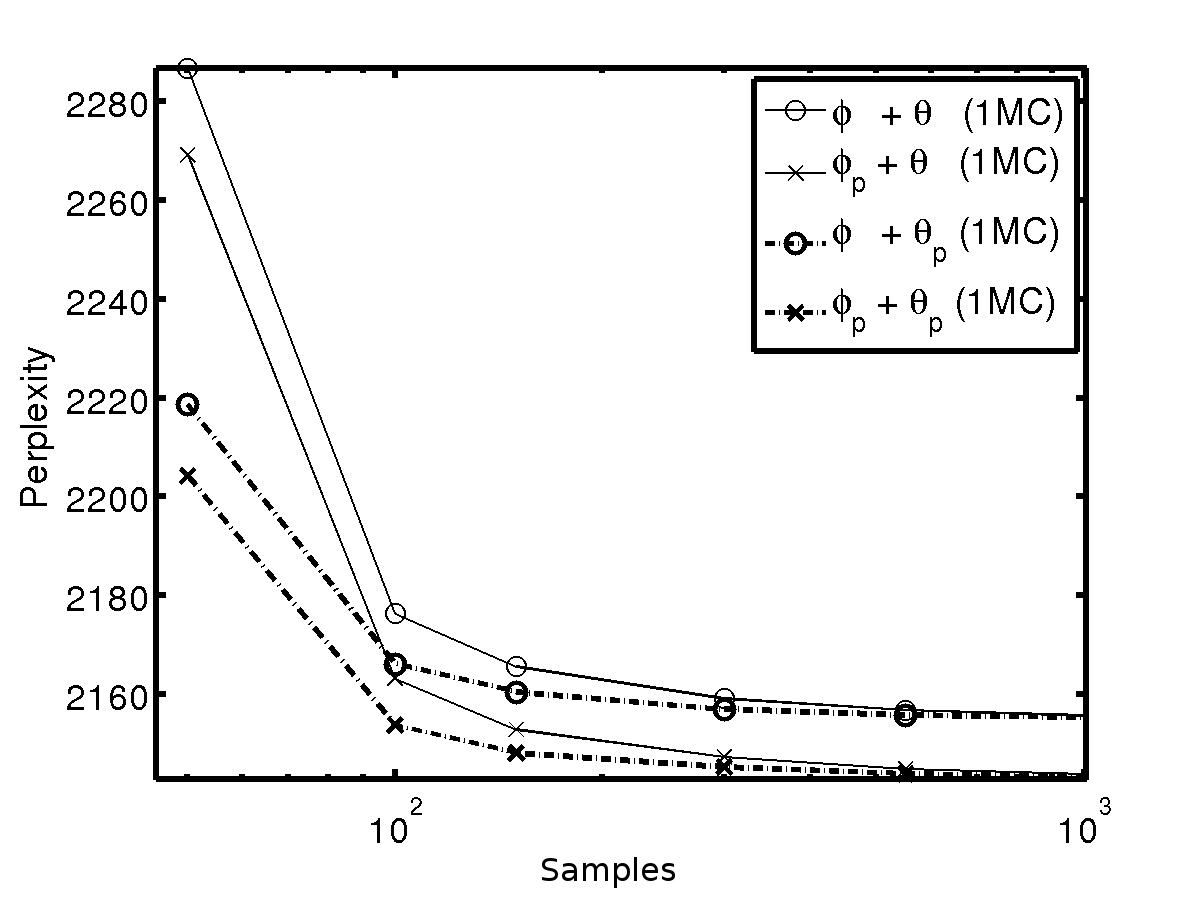}
\end{minipage}  
\hspace*{\fill}
\begin{minipage}{0.49\textwidth}
  \includegraphics[width=\linewidth]{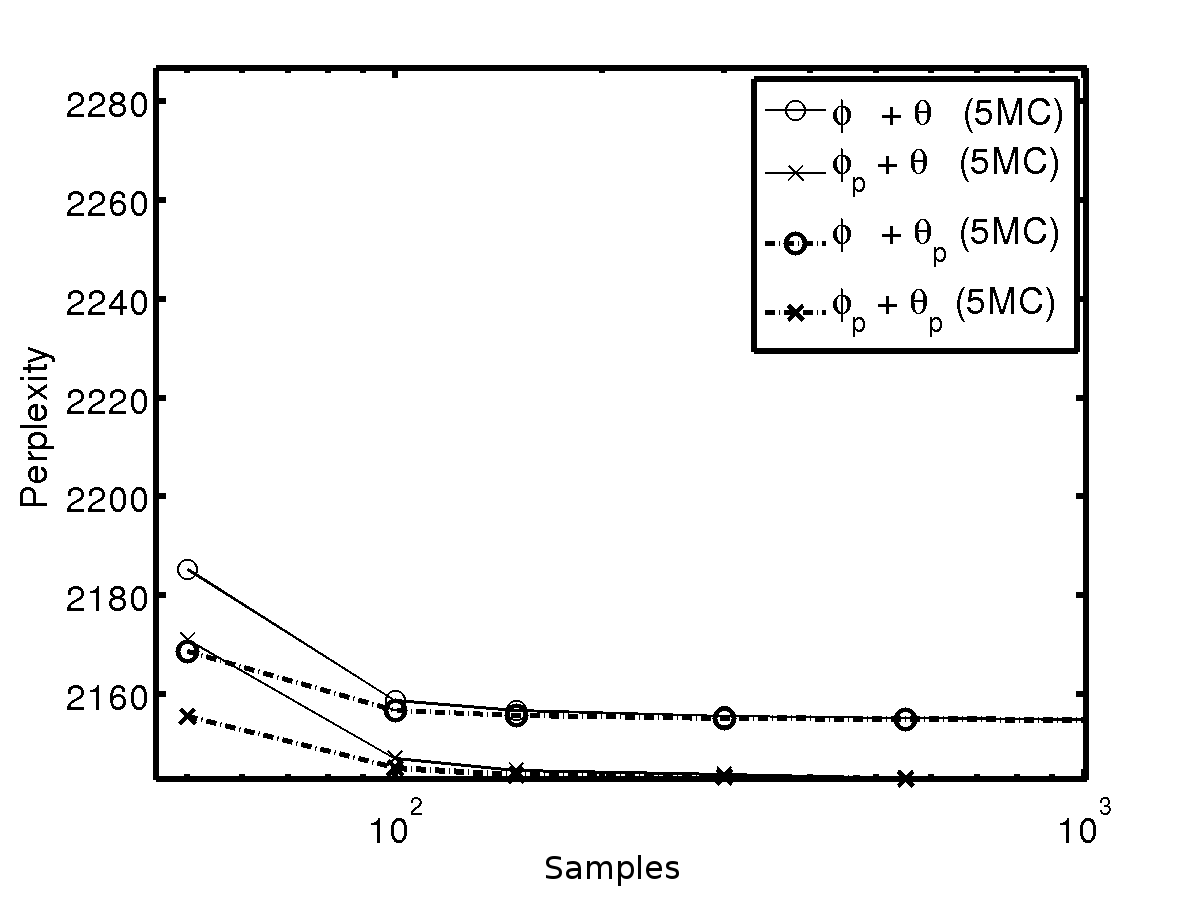}
\end{minipage}
\subcaption{$K$ = 50}
\end{minipage}

\begin{minipage}{\textwidth}
\begin{minipage}{0.49\textwidth}
  \includegraphics[width=\linewidth]{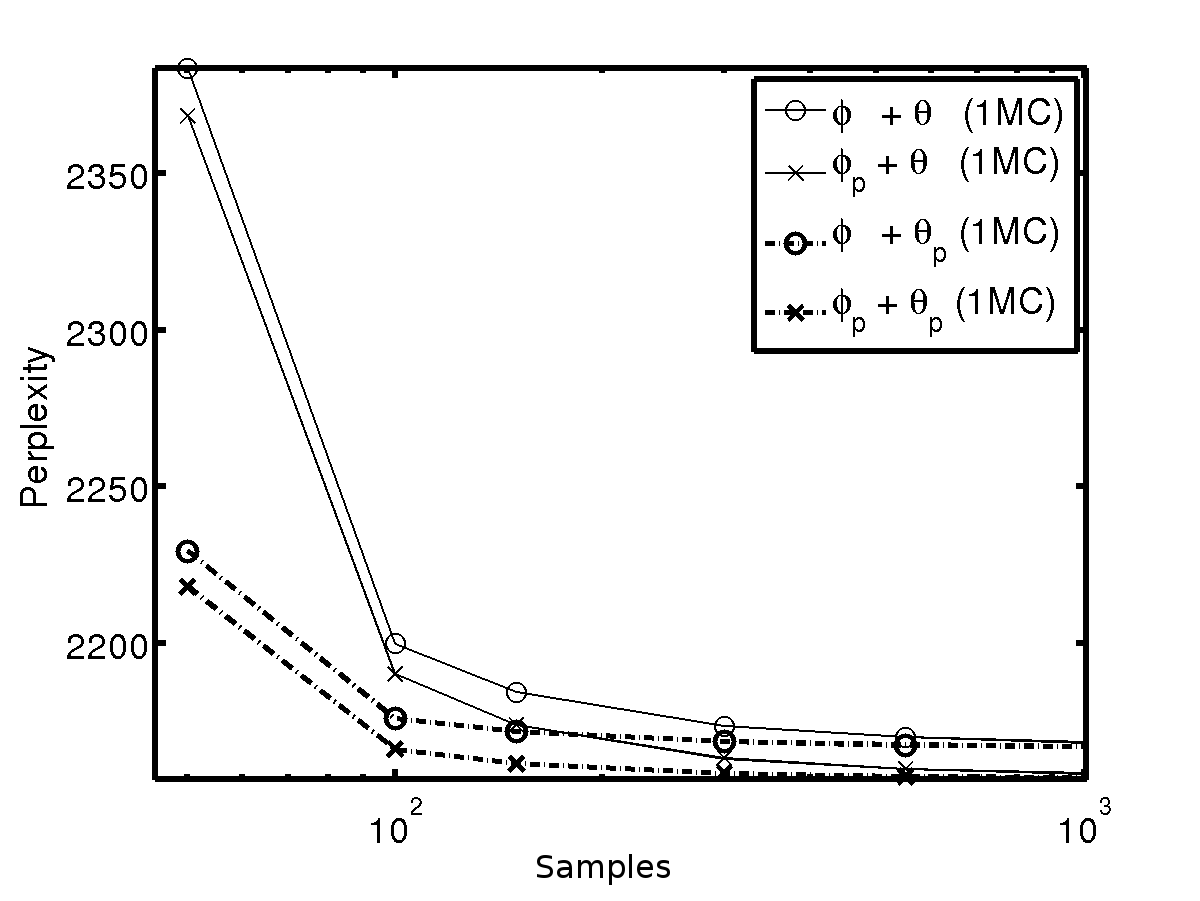}
\end{minipage}  
\hspace*{\fill}
\begin{minipage}{0.49\textwidth}
  \includegraphics[width=\linewidth]{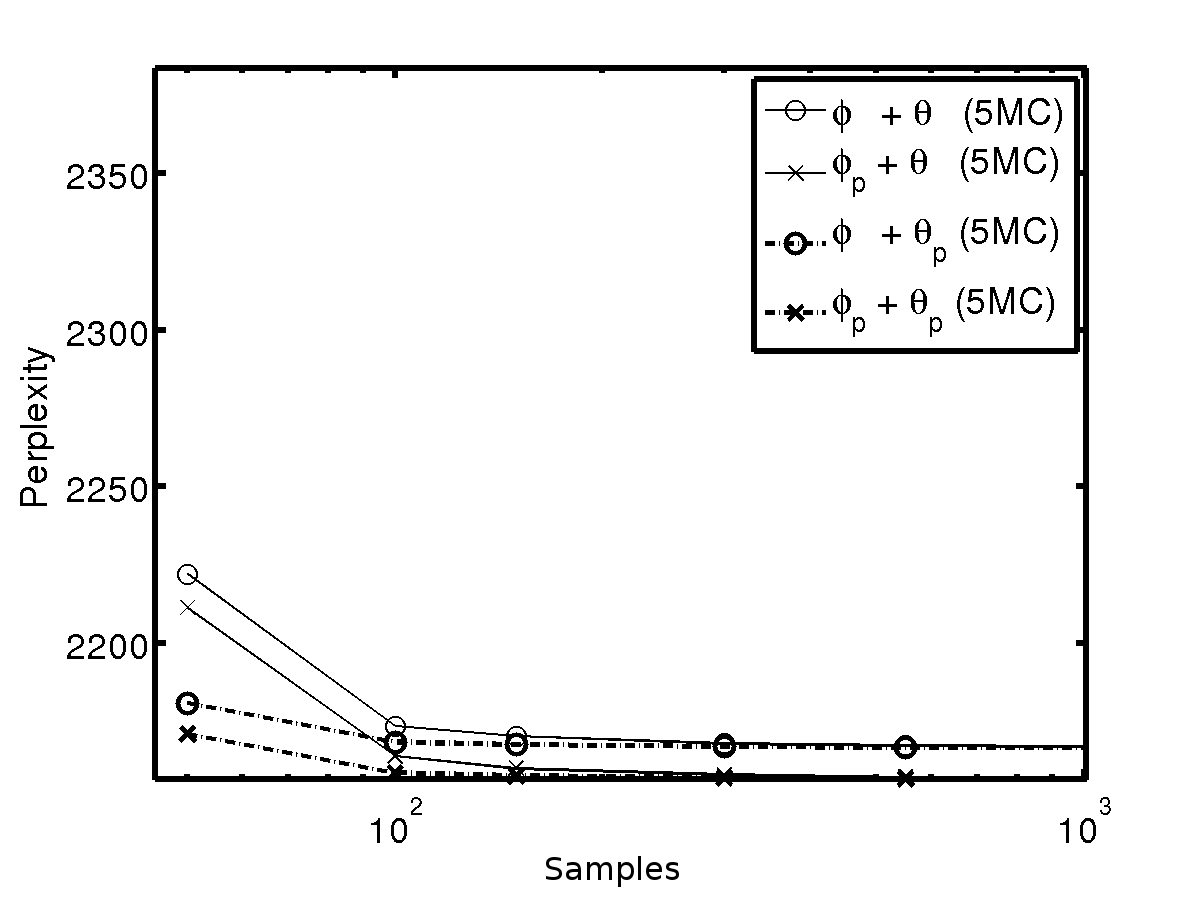}
\end{minipage}
  \subcaption{$K$ = 500}
\end{minipage}

\caption{Perplexity against the number of samples for the CGS$_p$ method and standard CGS for the BioASQ data set. Results are taken by averaging over 5 different runs.}
\label{fig:exp1bioasq}
\end{figure}

\begin{figure}[t]
\begin{minipage}{\textwidth}
 \begin{minipage}{0.49\textwidth}
   \includegraphics[width=\linewidth]{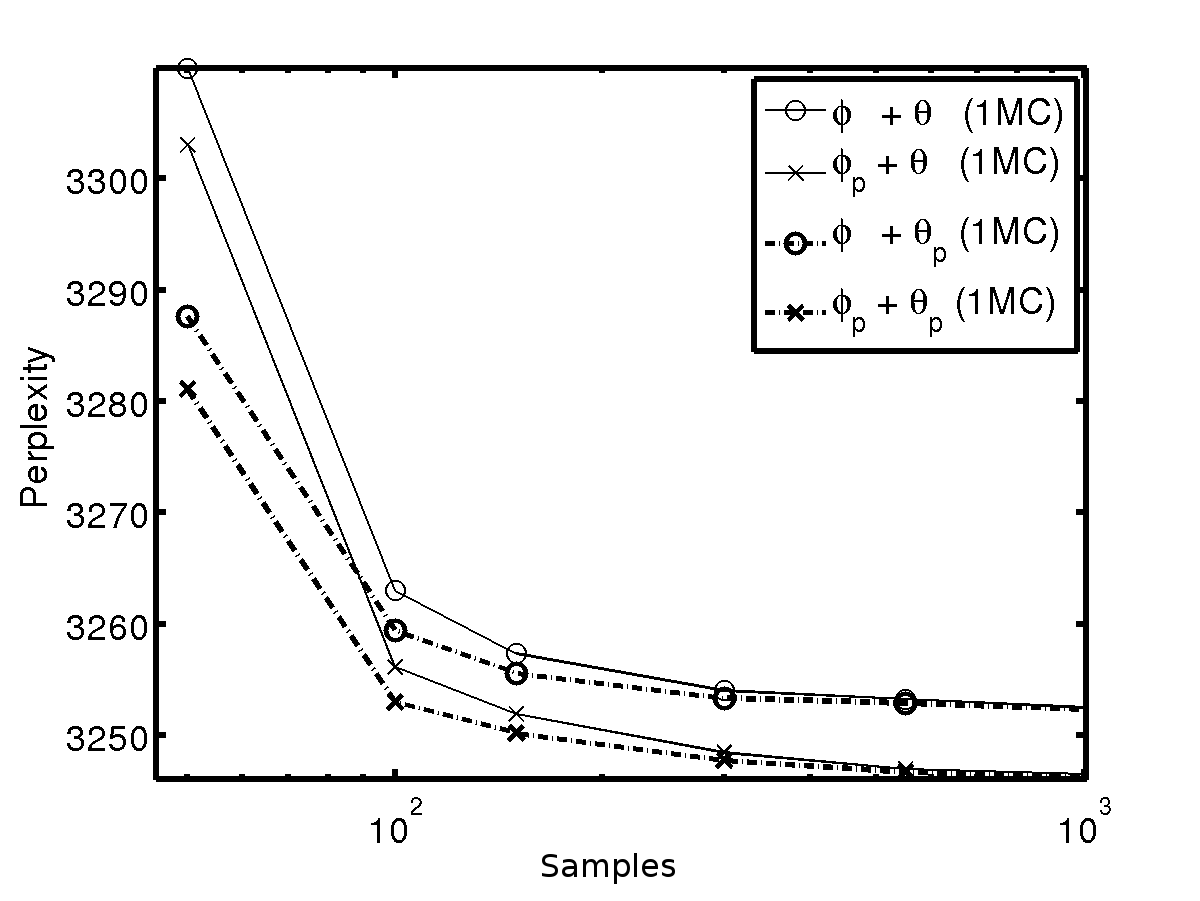}
 \end{minipage}  
 \hspace*{\fill}
 \begin{minipage}{0.49\textwidth}
   \includegraphics[width=\linewidth]{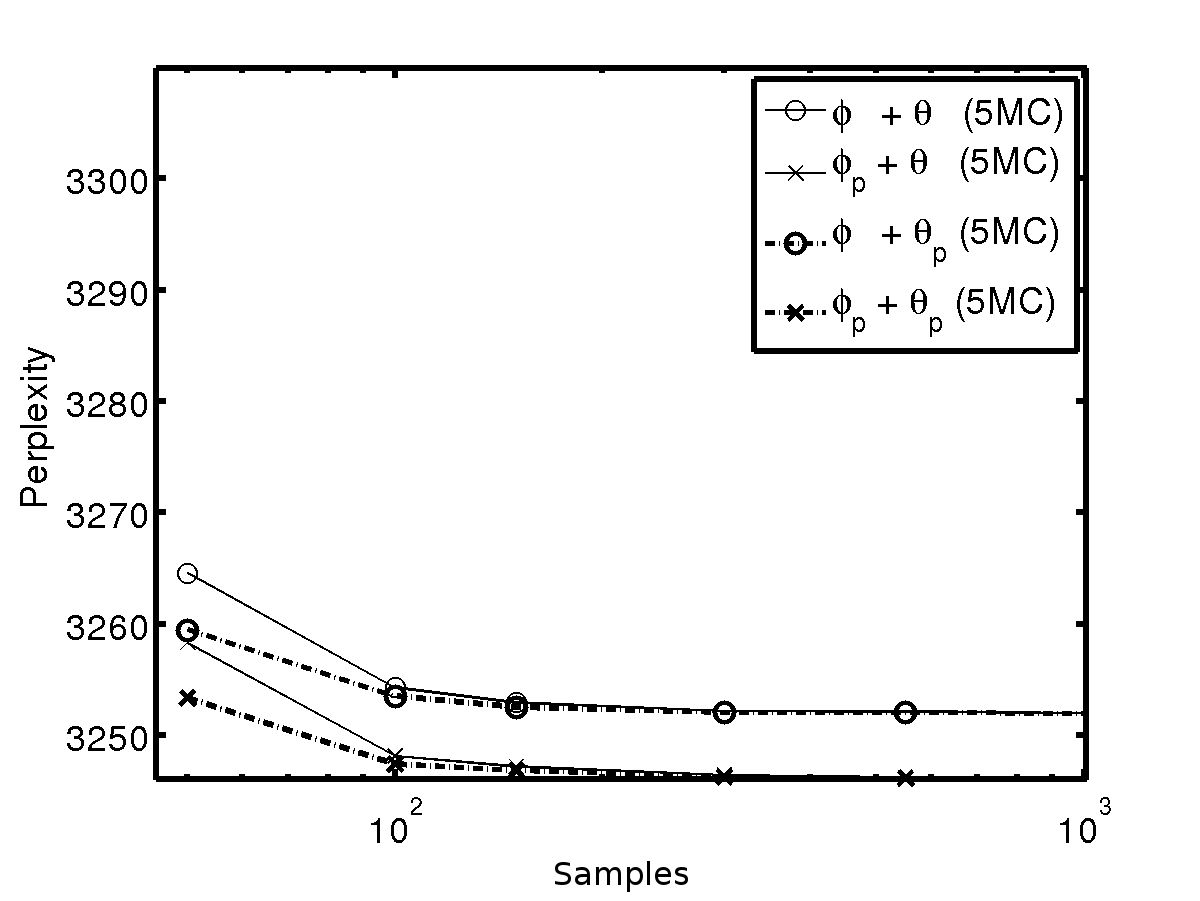}
 \end{minipage}
\subcaption{$K$ = 20}
\end{minipage}

\begin{minipage}{\textwidth}
\begin{minipage}{0.49\textwidth}
  \includegraphics[width=\linewidth]{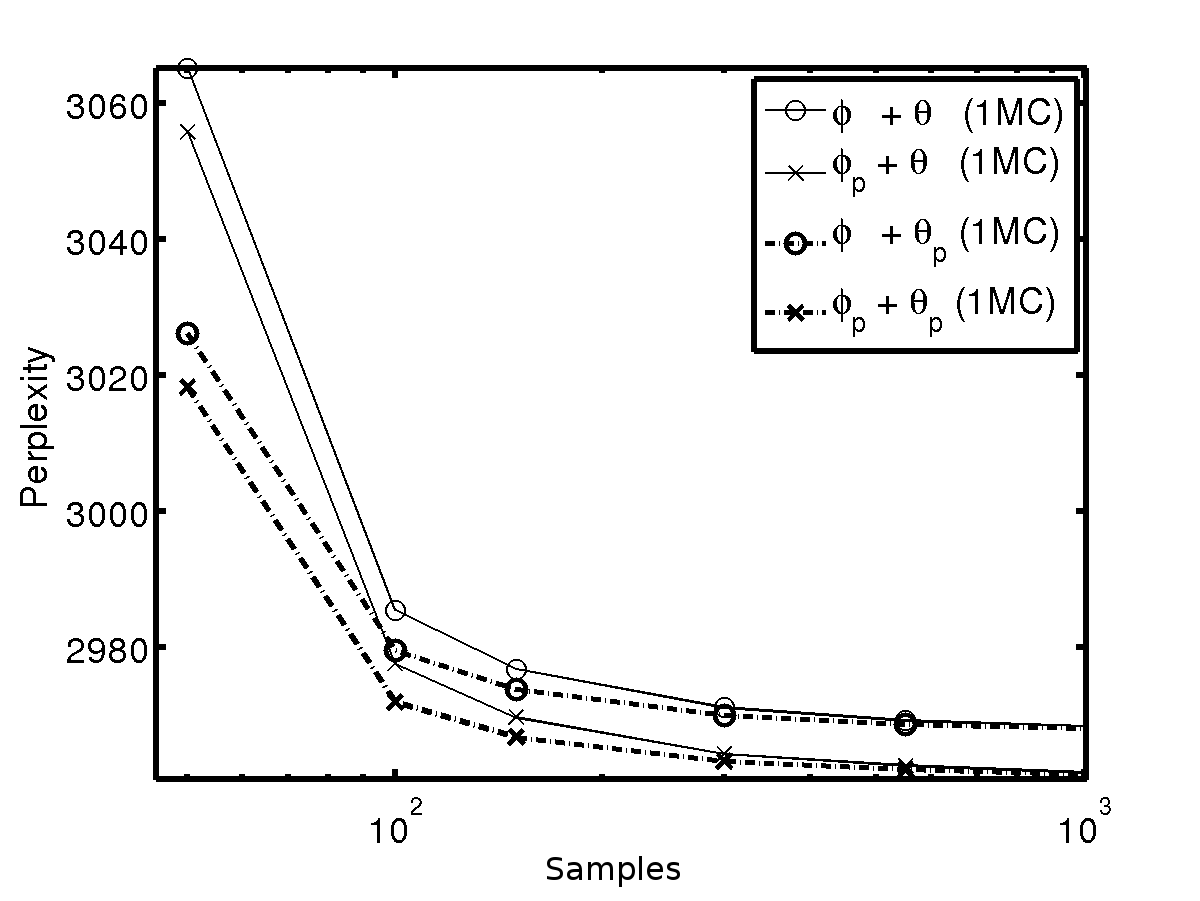}
\end{minipage}  
\hspace*{\fill}
\begin{minipage}{0.49\textwidth}
  \includegraphics[width=\linewidth]{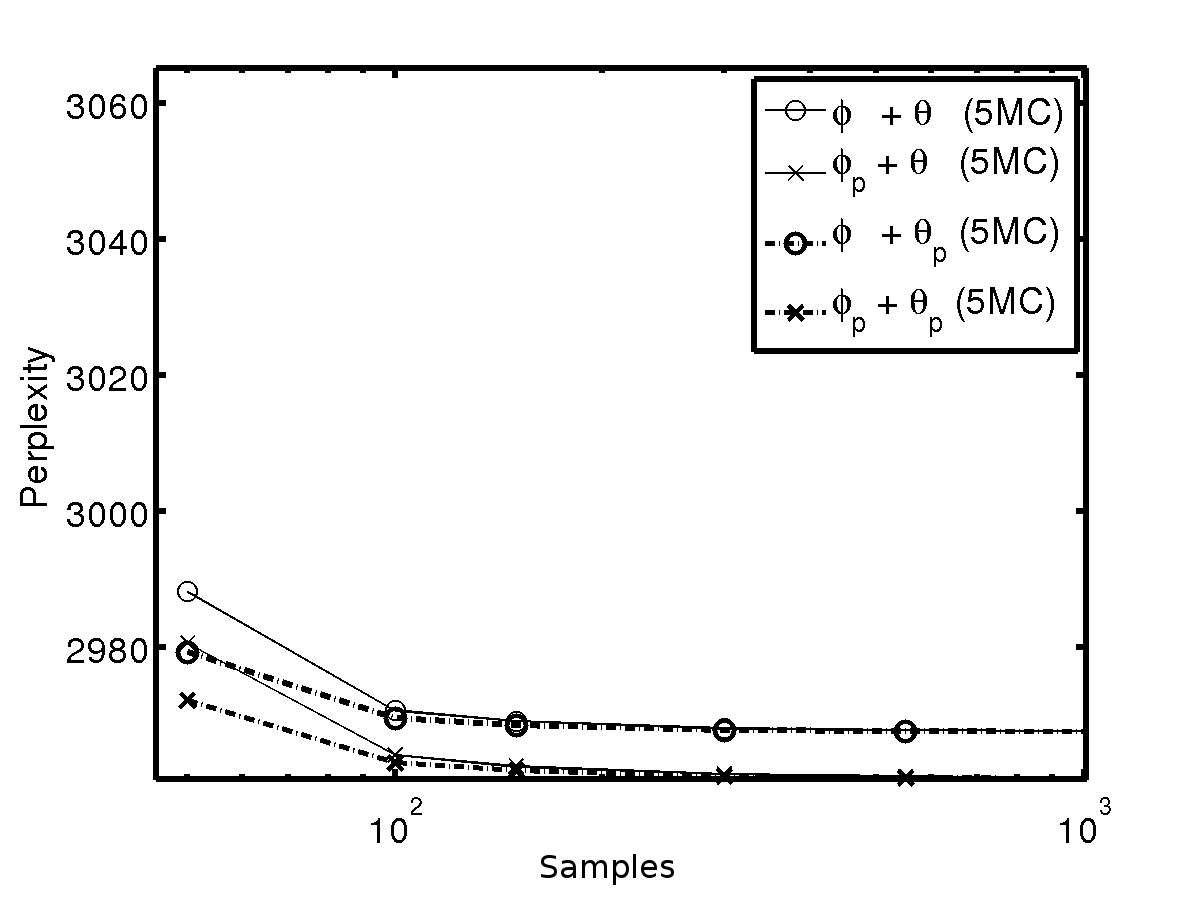}
\end{minipage}
\subcaption{$K$ = 50}
\end{minipage}

\begin{minipage}{\textwidth}
\begin{minipage}{0.49\textwidth}
  \includegraphics[width=\linewidth]{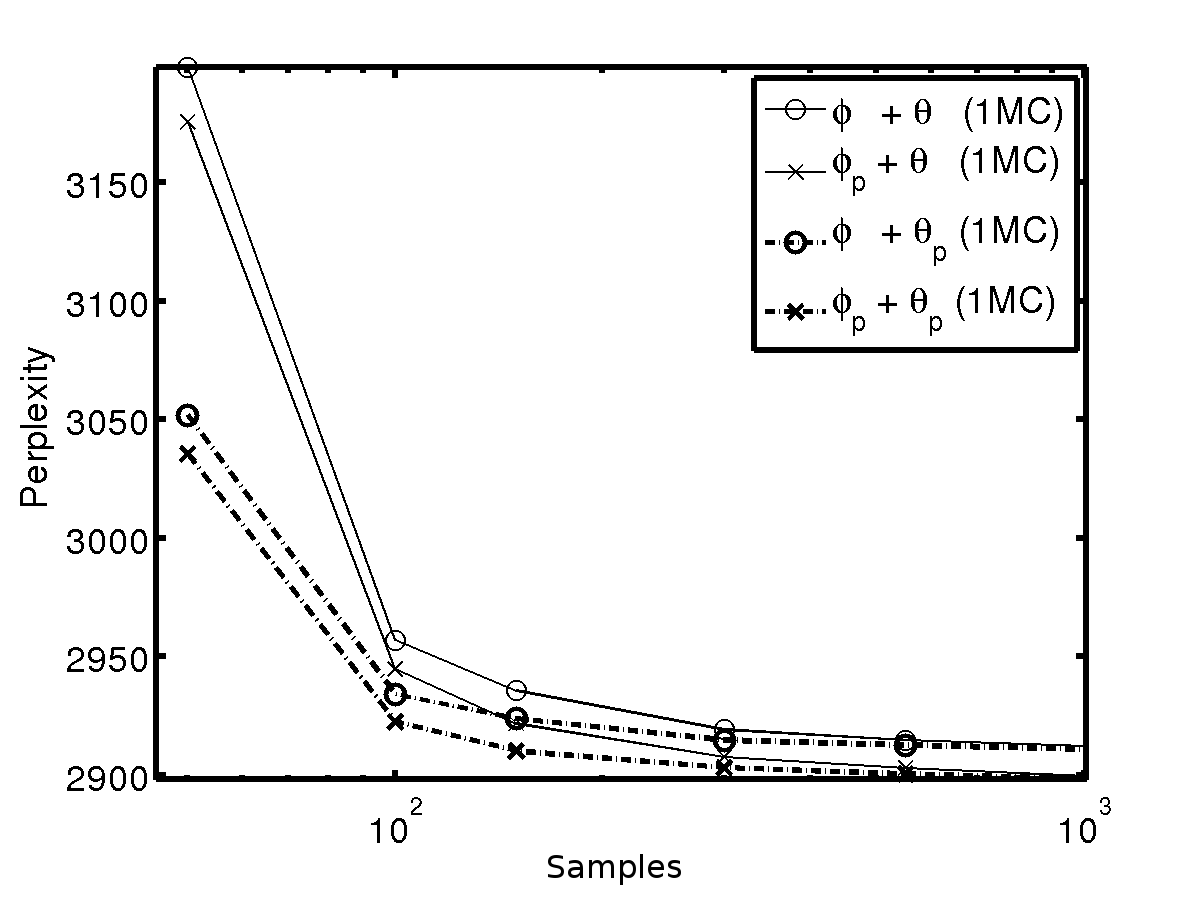}
\end{minipage}  
\hspace*{\fill}
\begin{minipage}{0.49\textwidth}
  \includegraphics[width=\linewidth]{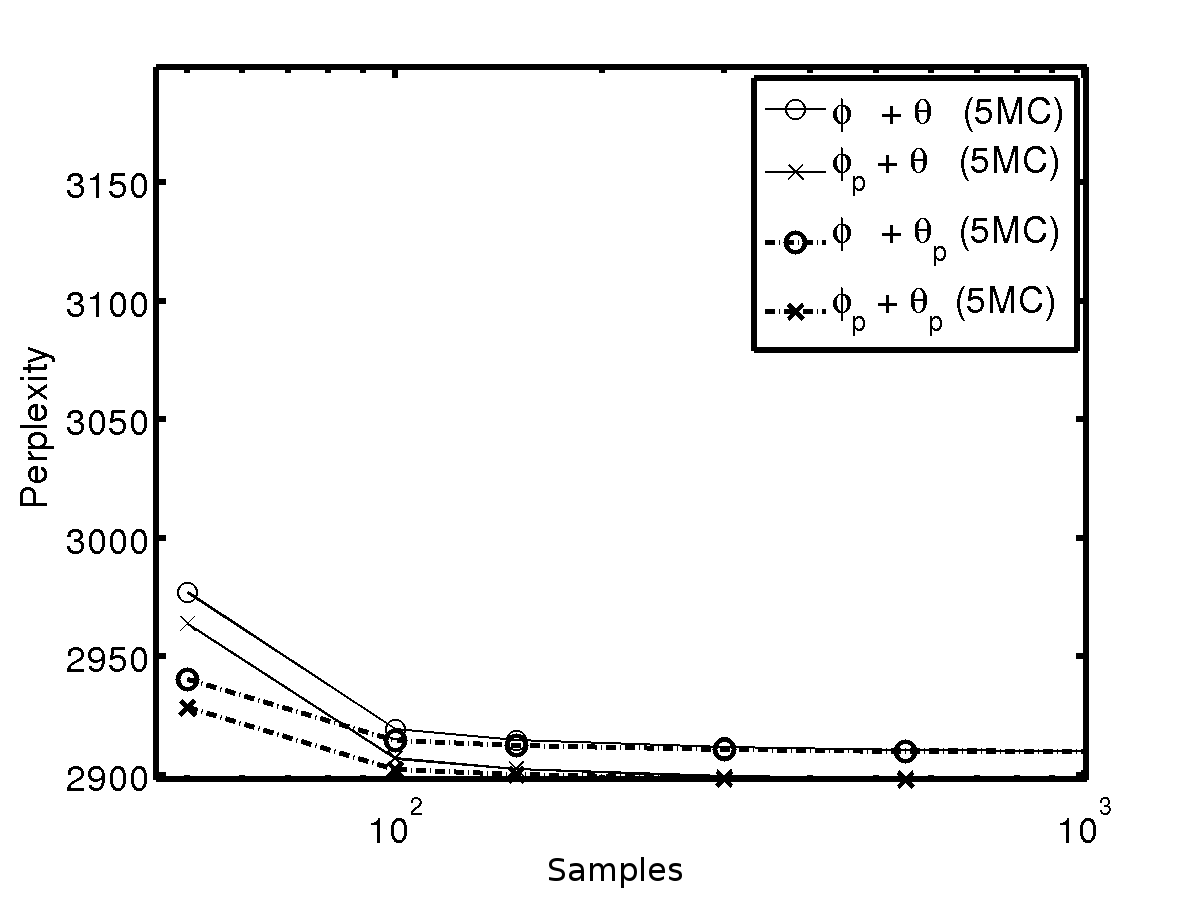}
\end{minipage}
  \subcaption{$K$ = 500}
\end{minipage}

\caption{Perplexity against the number of samples for the CGS$_p$ methods and standard CGS for the New York Times data set. Results are taken by averaging over 5 different runs.}
\label{fig:exp1NYT}
\end{figure}

\begin{figure}[t]
\begin{minipage}{\textwidth}
 \begin{minipage}{0.49\textwidth}
   \includegraphics[width=\linewidth]{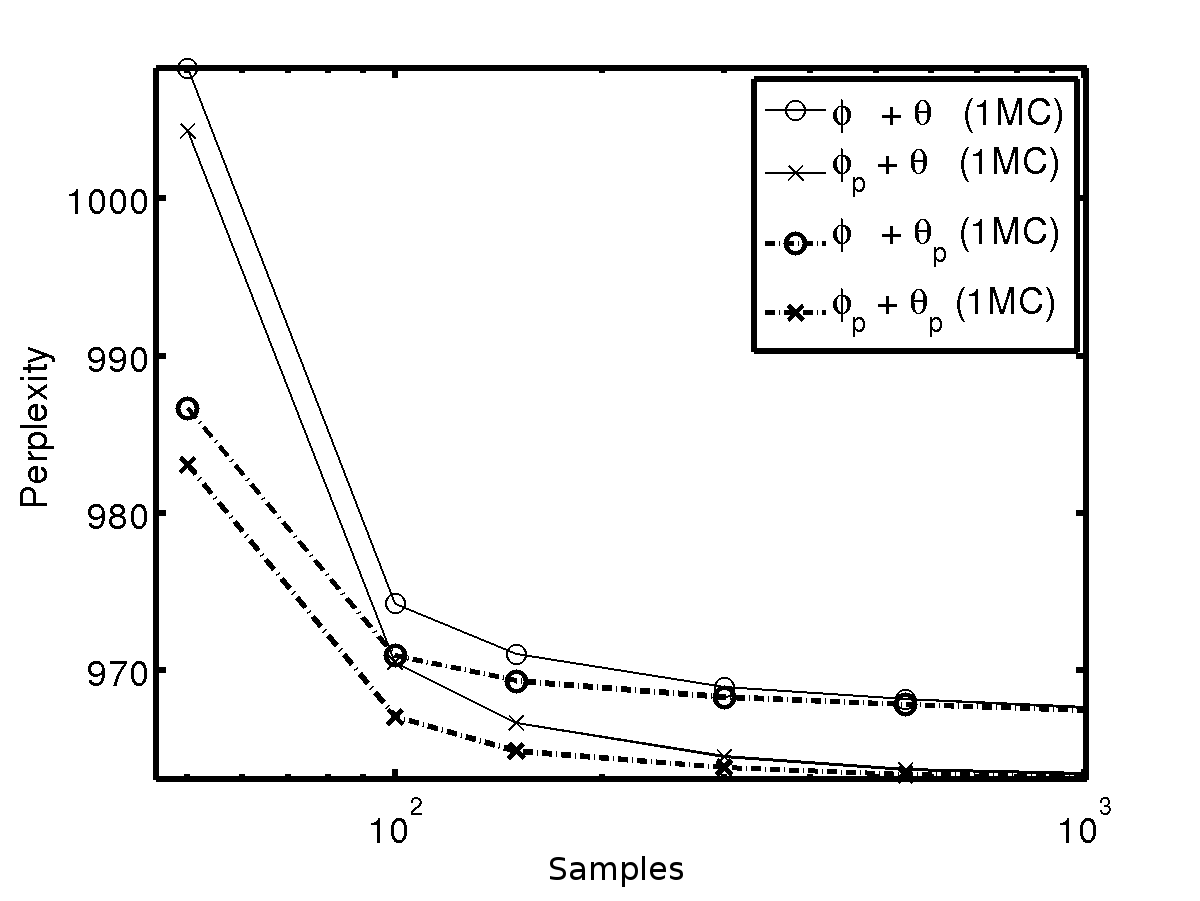}
 \end{minipage}  
 \hspace*{\fill}
 \begin{minipage}{0.49\textwidth}
   \includegraphics[width=\linewidth]{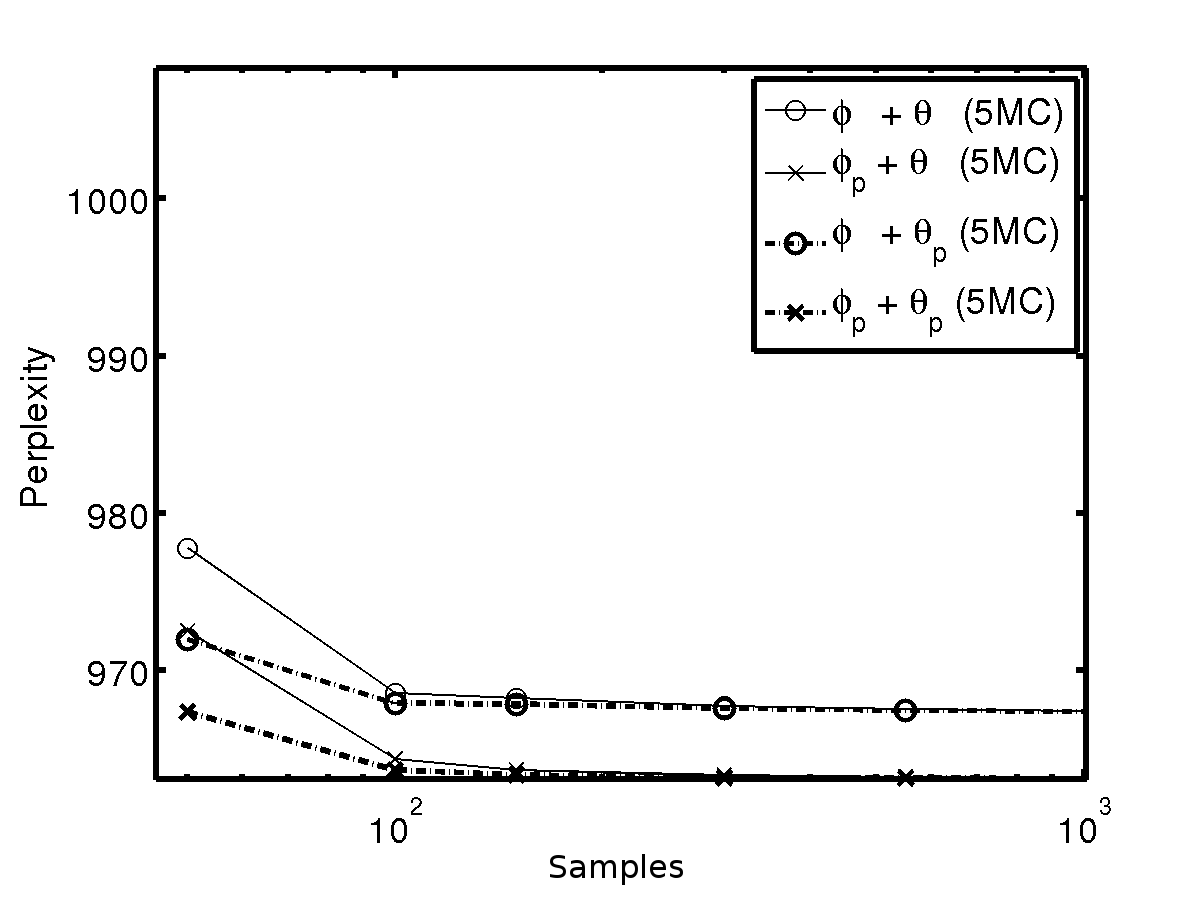}
 \end{minipage}
\subcaption{$K$ = 20}
\end{minipage}

\begin{minipage}{\textwidth}
\begin{minipage}{0.49\textwidth}
  \includegraphics[width=\linewidth]{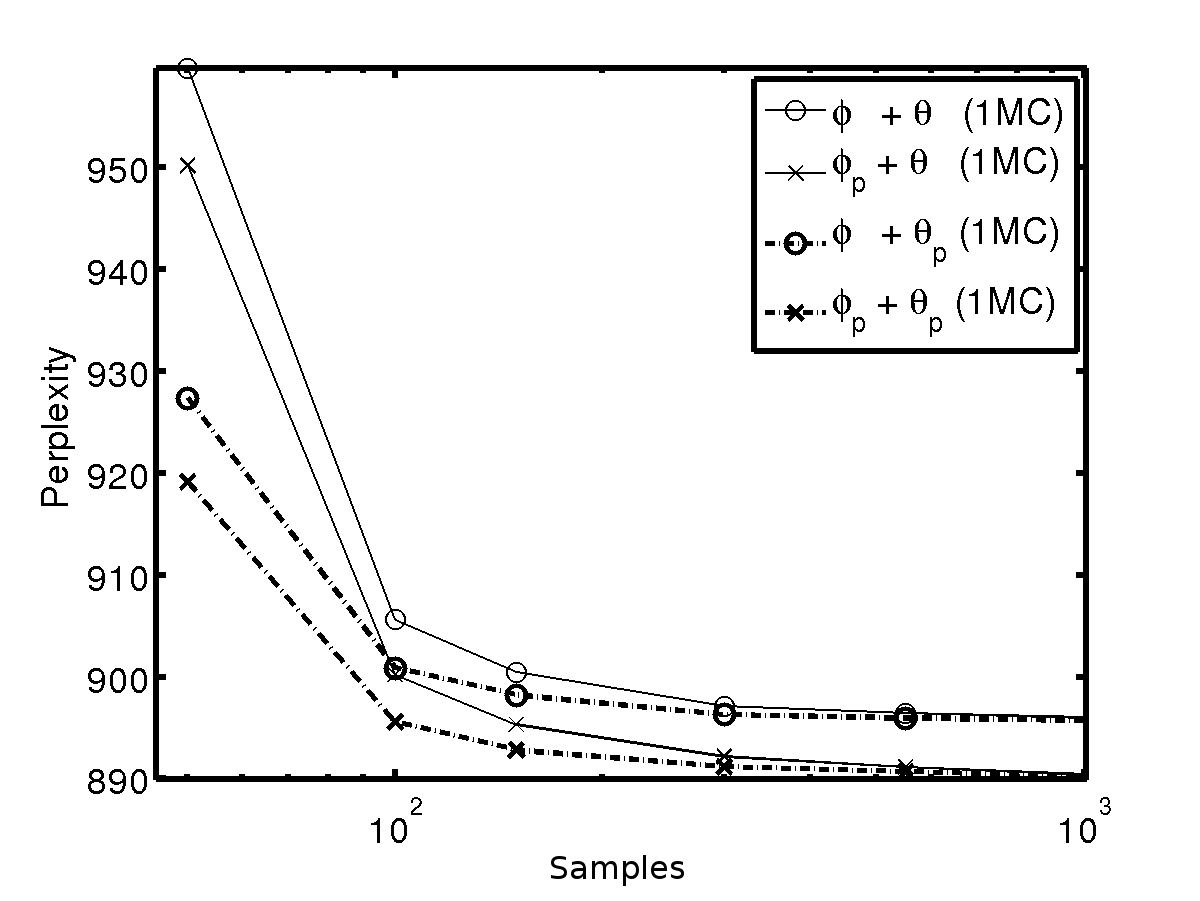}
\end{minipage}  
\hspace*{\fill}
\begin{minipage}{0.49\textwidth}
  \includegraphics[width=\linewidth]{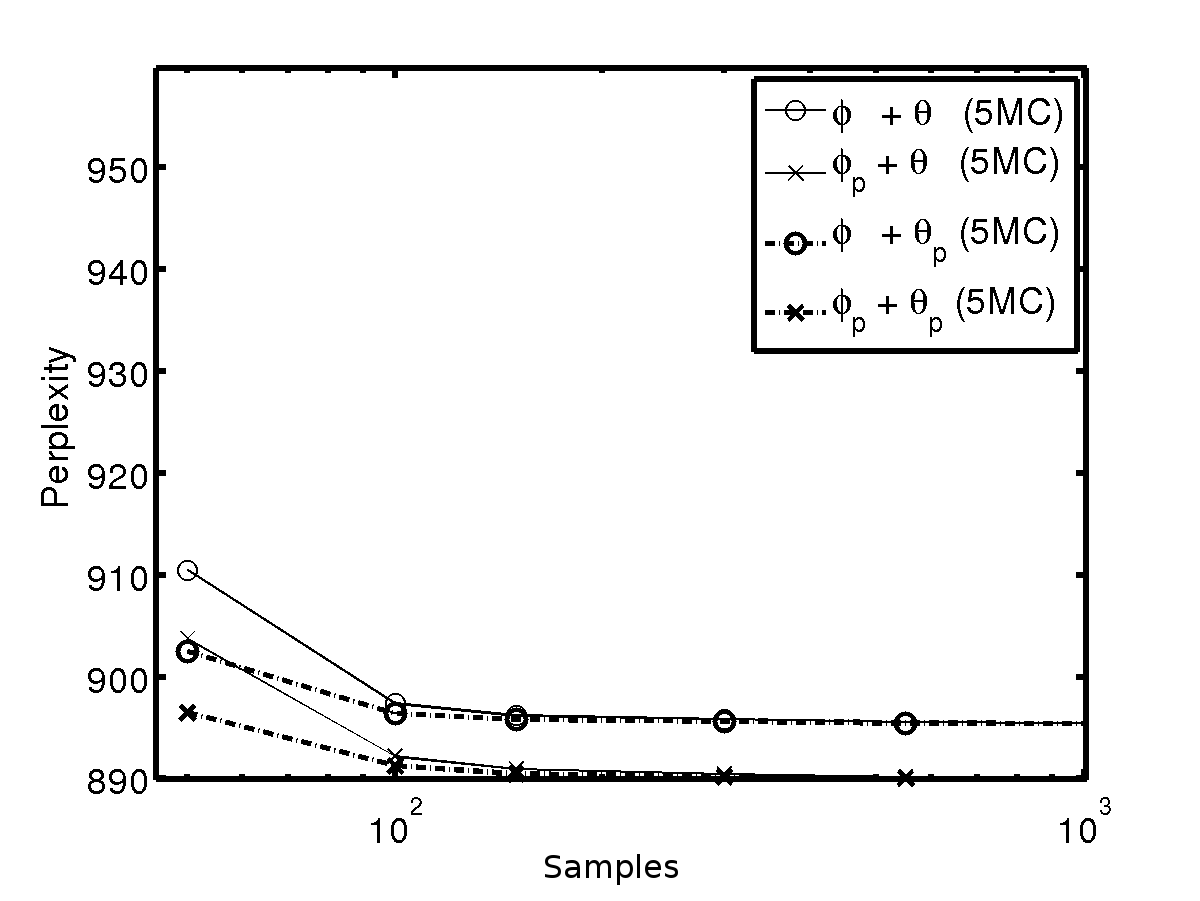}
\end{minipage}
\subcaption{$K$ = 50}
\end{minipage}

\begin{minipage}{\textwidth}
\begin{minipage}{0.49\textwidth}
  \includegraphics[width=\linewidth]{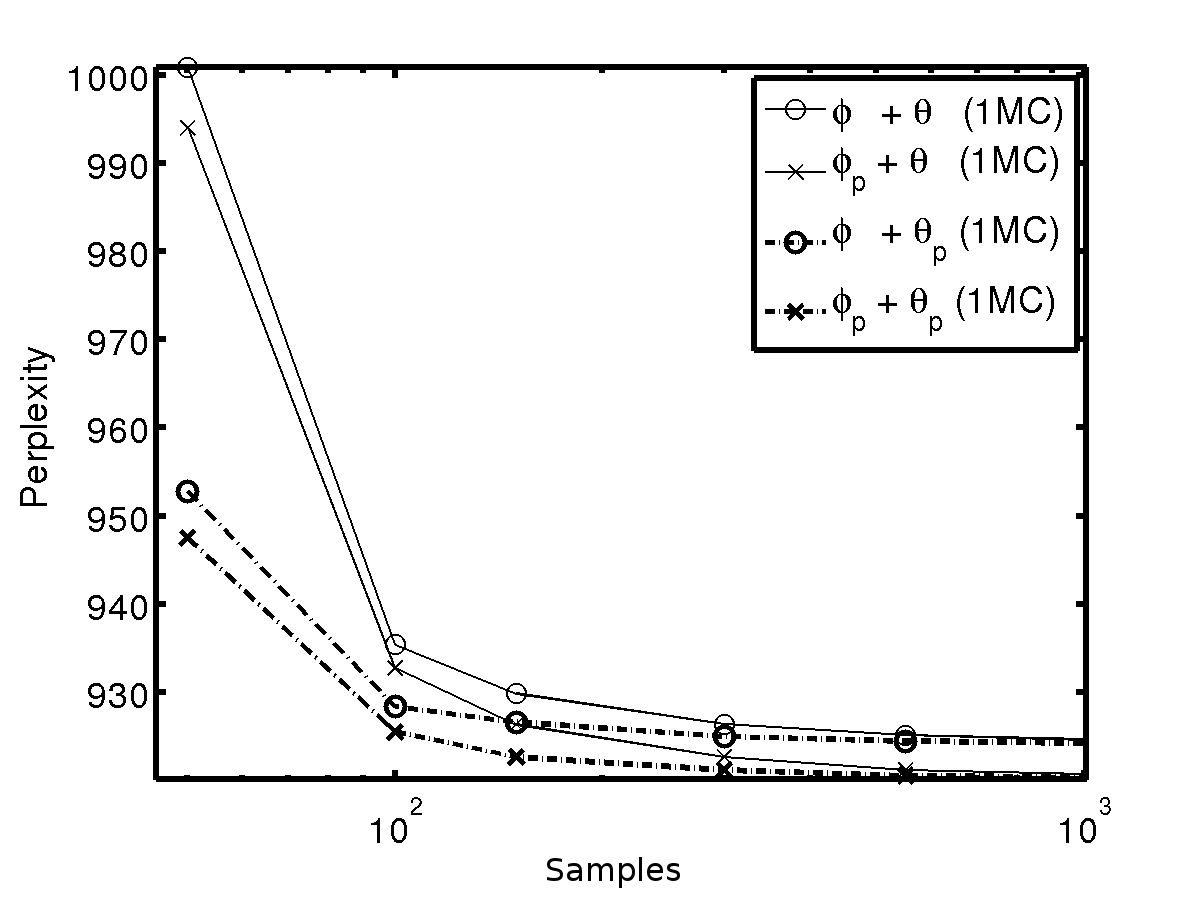}
\end{minipage}  
\hspace*{\fill}
\begin{minipage}{0.49\textwidth}
  \includegraphics[width=\linewidth]{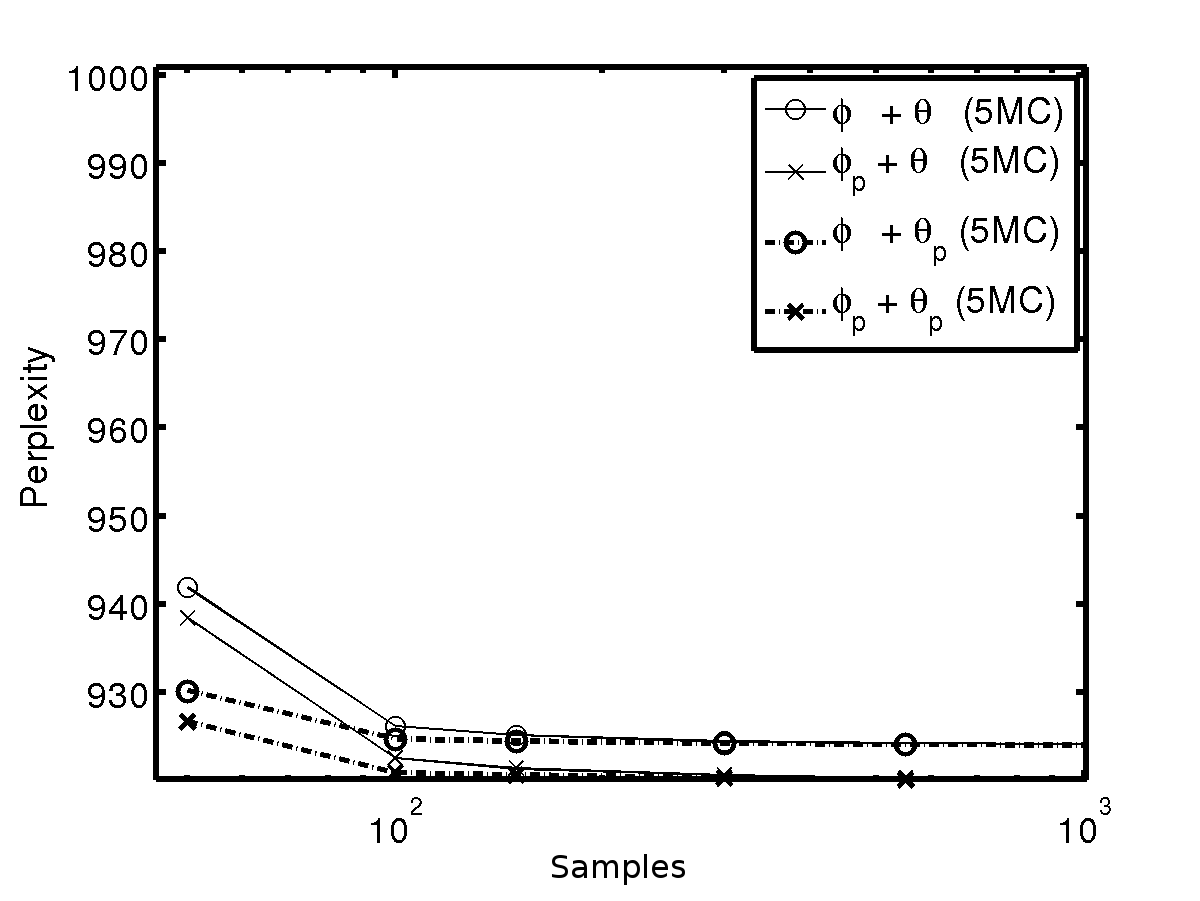}
\end{minipage}
  \subcaption{$K$ = 500}
\end{minipage}

\caption{Perplexity against the number of samples for the CGS$_p$ methods and standard CGS for the Reuters-21578 data set. Results are taken by averaging over 5 different runs.}
\label{fig:exp1reuters}
\end{figure}

\begin{figure}[t]
\begin{minipage}{\textwidth}
 \begin{minipage}{0.49\textwidth}
   \includegraphics[width=\linewidth]{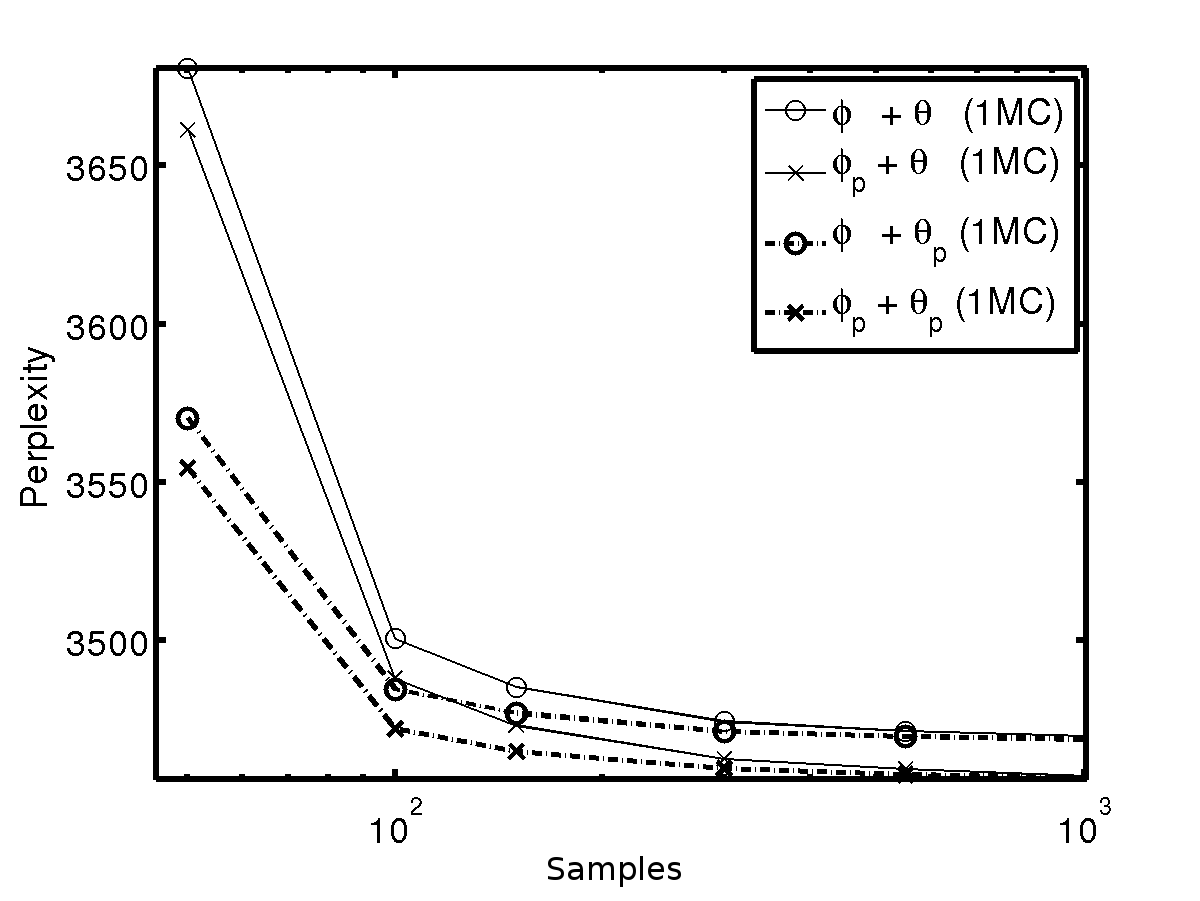}
 \end{minipage}  
 \hspace*{\fill}
 \begin{minipage}{0.49\textwidth}
   \includegraphics[width=\linewidth]{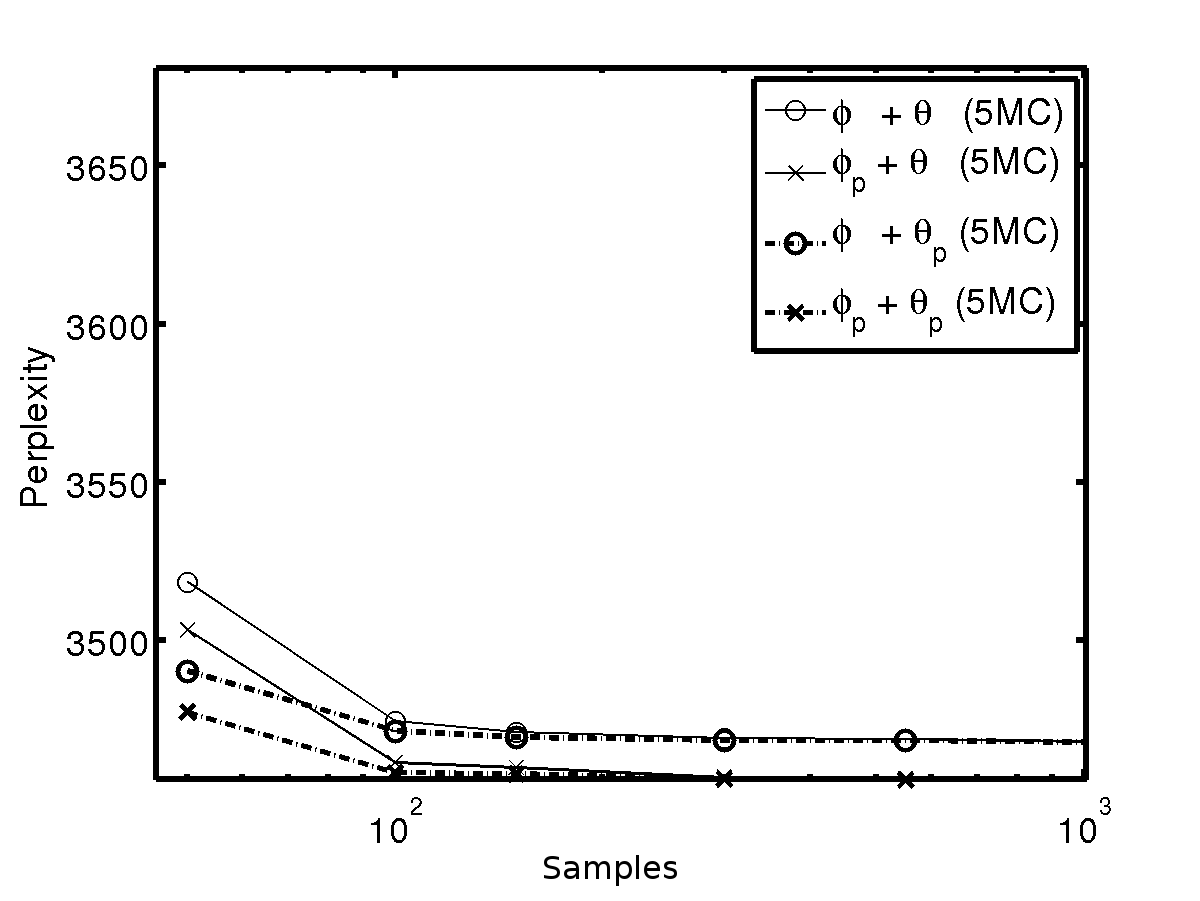}
 \end{minipage}
\subcaption{$K$ = 20}
\end{minipage}

\begin{minipage}{\textwidth}
\begin{minipage}{0.49\textwidth}
  \includegraphics[width=\linewidth]{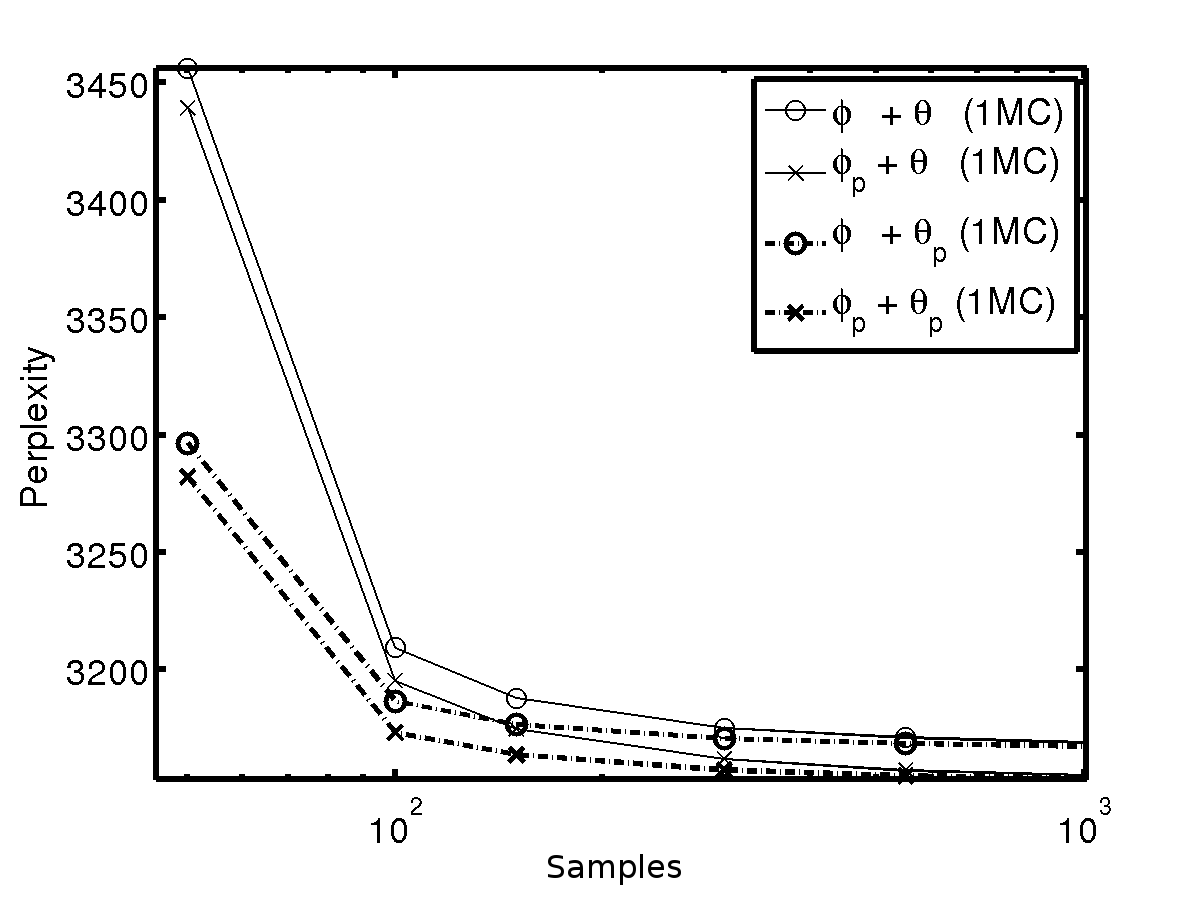}
\end{minipage}  
\hspace*{\fill}
\begin{minipage}{0.49\textwidth}
  \includegraphics[width=\linewidth]{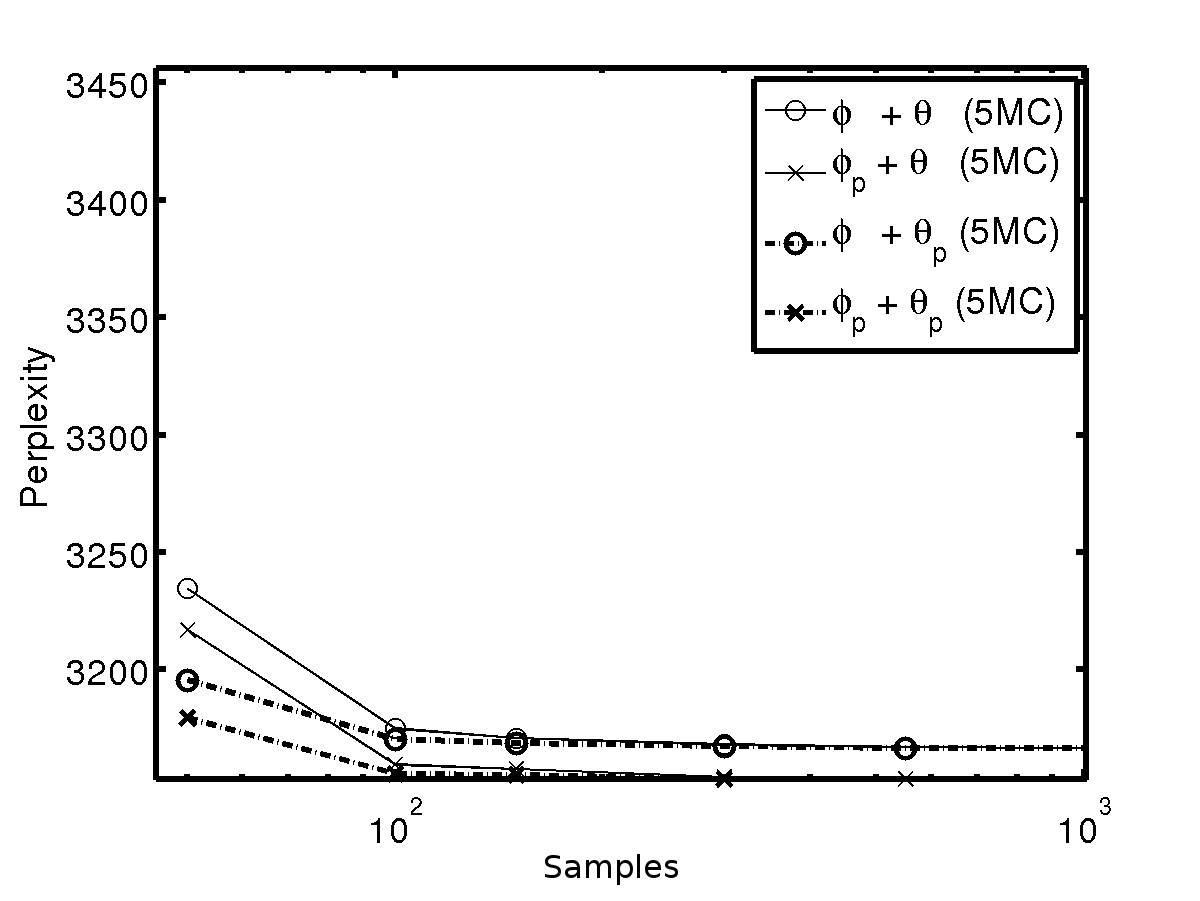}
\end{minipage}
\subcaption{$K$ = 50}
\end{minipage}

\begin{minipage}{\textwidth}
\begin{minipage}{0.49\textwidth}
  \includegraphics[width=\linewidth]{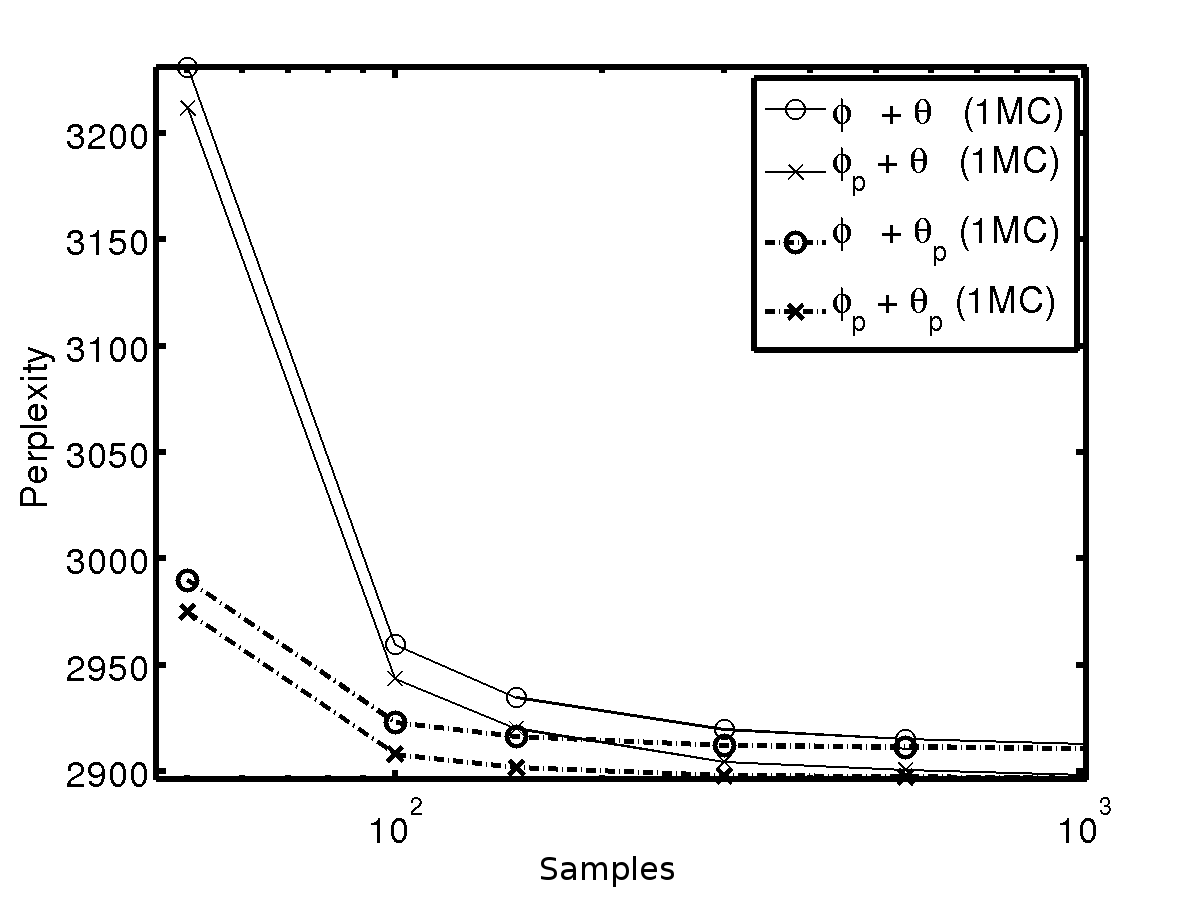}
\end{minipage}  
\hspace*{\fill}
\begin{minipage}{0.49\textwidth}
  \includegraphics[width=\linewidth]{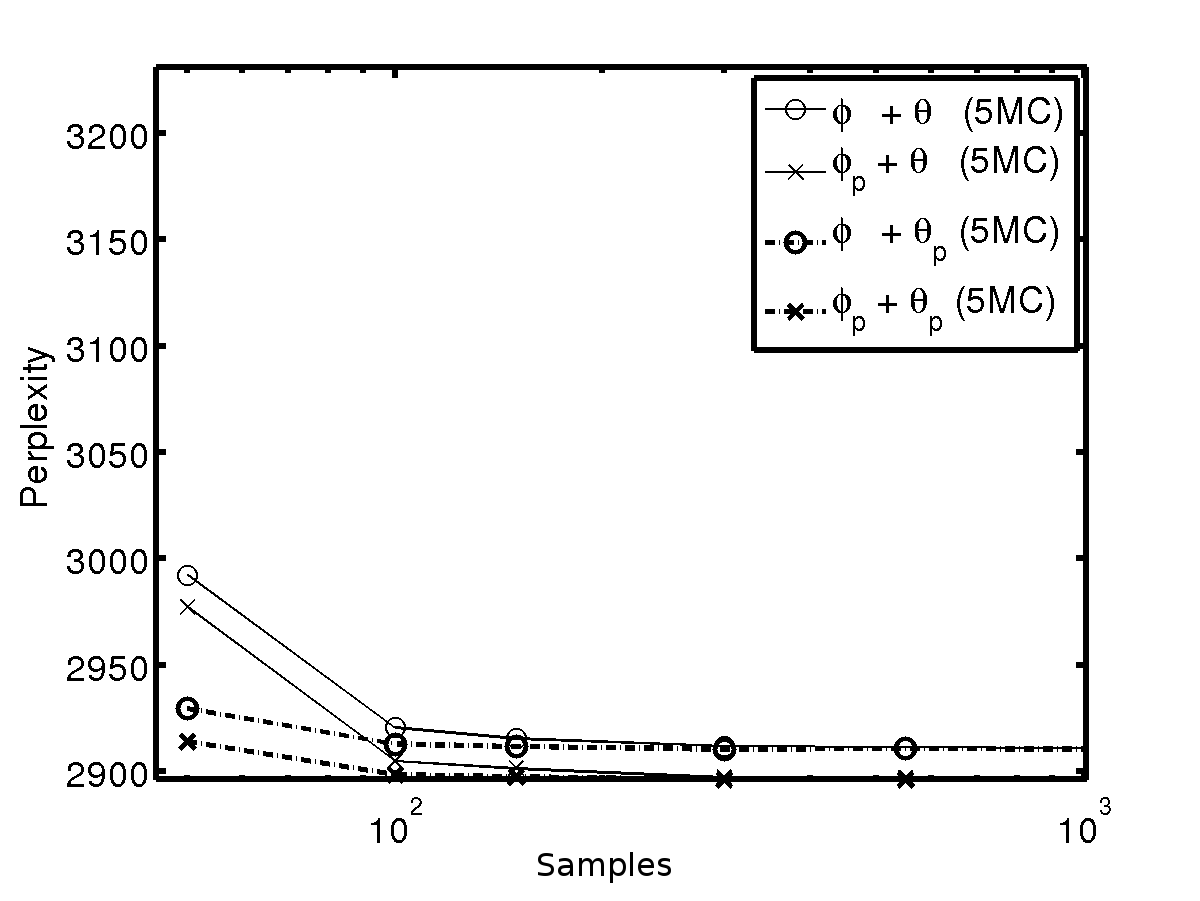}
\end{minipage}
  \subcaption{$K$ = 500}
\end{minipage}

\caption{Perplexity against the number of samples for the CGS$_p$ methods and standard CGS for the TASA data set. Results are taken by averaging over 5 different runs.}
\label{fig:exp1tasa}
\end{figure}

Figures \ref{fig:exp1bioasq} -- \ref{fig:exp1tasa} present additional results for the experiment of Section \ref{sec:exp1}, using different topic configurations ($K$ = 20, 50, 500) for each of the four data sets (BioASQ, New York Times, Reuters-21578, TASA) respectively. These results are aligned with the conclusions made in Section \ref{sec:exp1} and are reported here as further empirical validation of the theory behind CGS$_p$. 

\section{}
\label{app:perplexityVB}

\begin{figure}[tb]
\begin{minipage}{0.49\textwidth}
  \includegraphics[width=\linewidth]{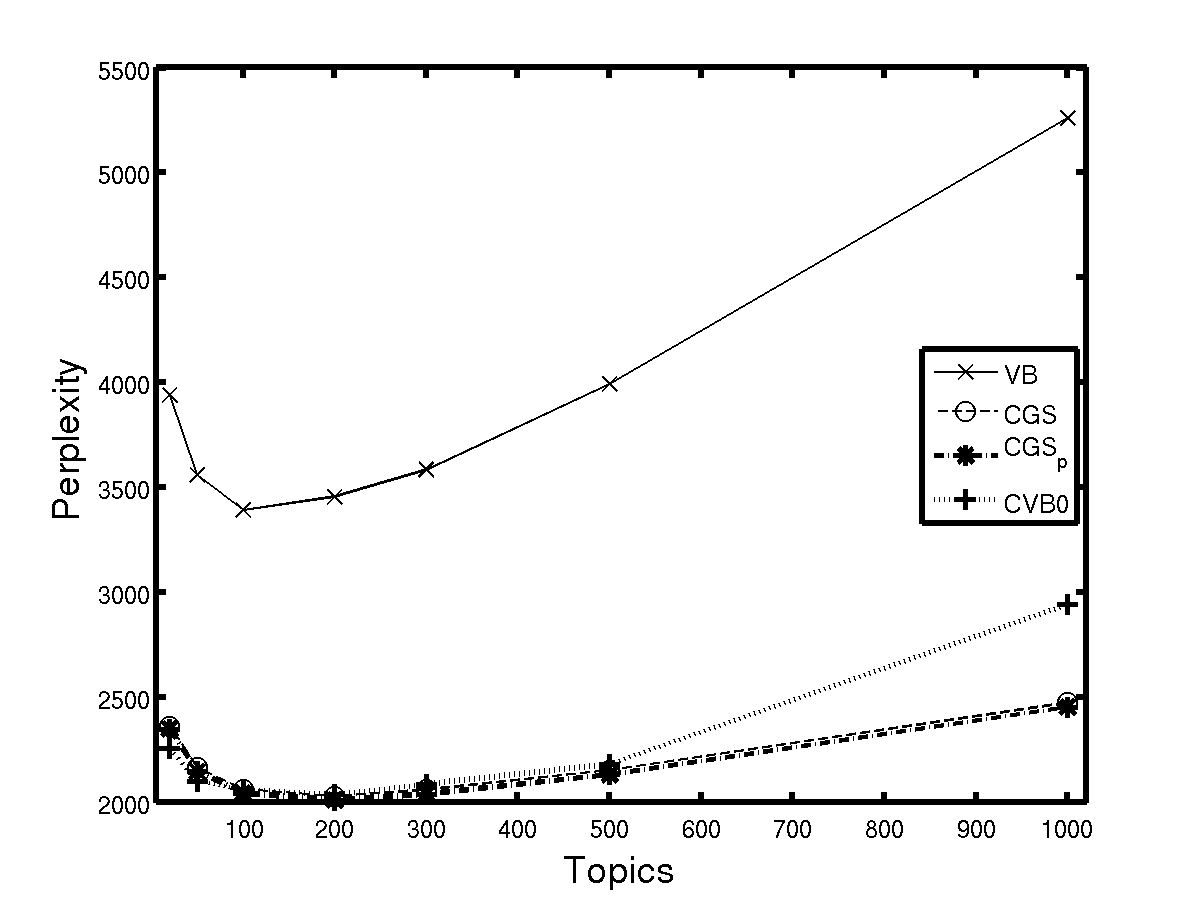}
  \subcaption{BioASQ}
\end{minipage}  
\hspace*{\fill}
\begin{minipage}{0.49\textwidth}
  \includegraphics[width=\linewidth]{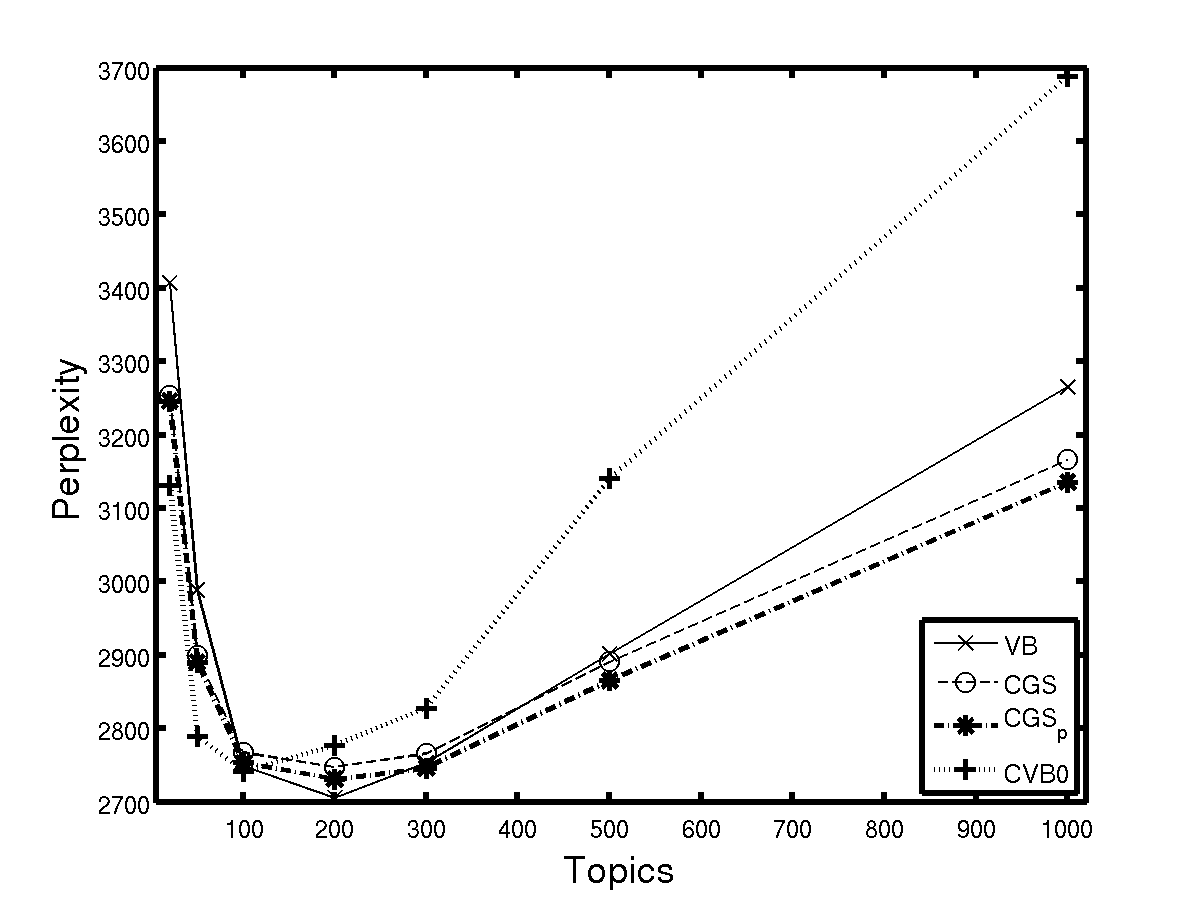}
  \subcaption{New York Times}
\end{minipage}
\begin{minipage}{0.49\textwidth}
  \includegraphics[width=\linewidth]{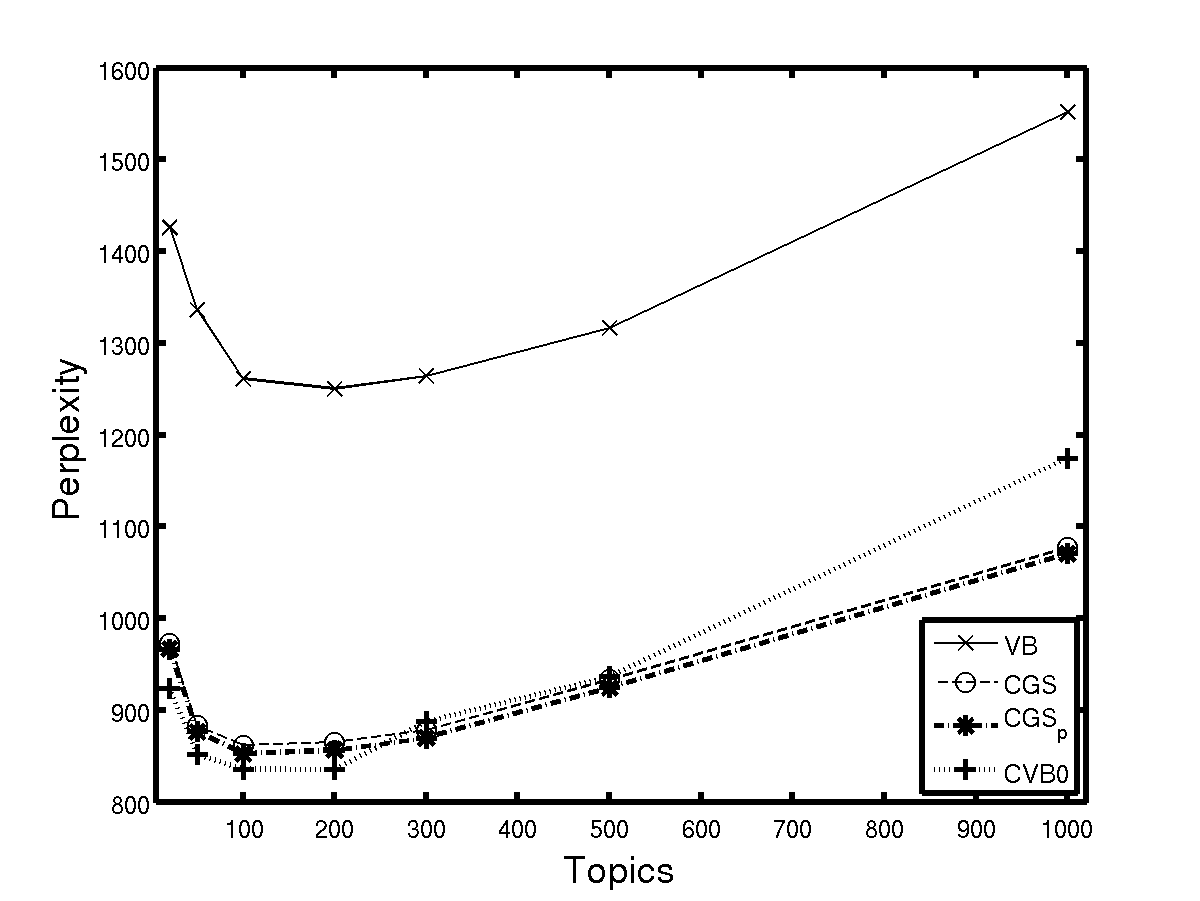}
  \subcaption{Reuters-21578}
\end{minipage}  
\hspace*{\fill}
\begin{minipage}{0.49\textwidth}
  \includegraphics[width=\linewidth]{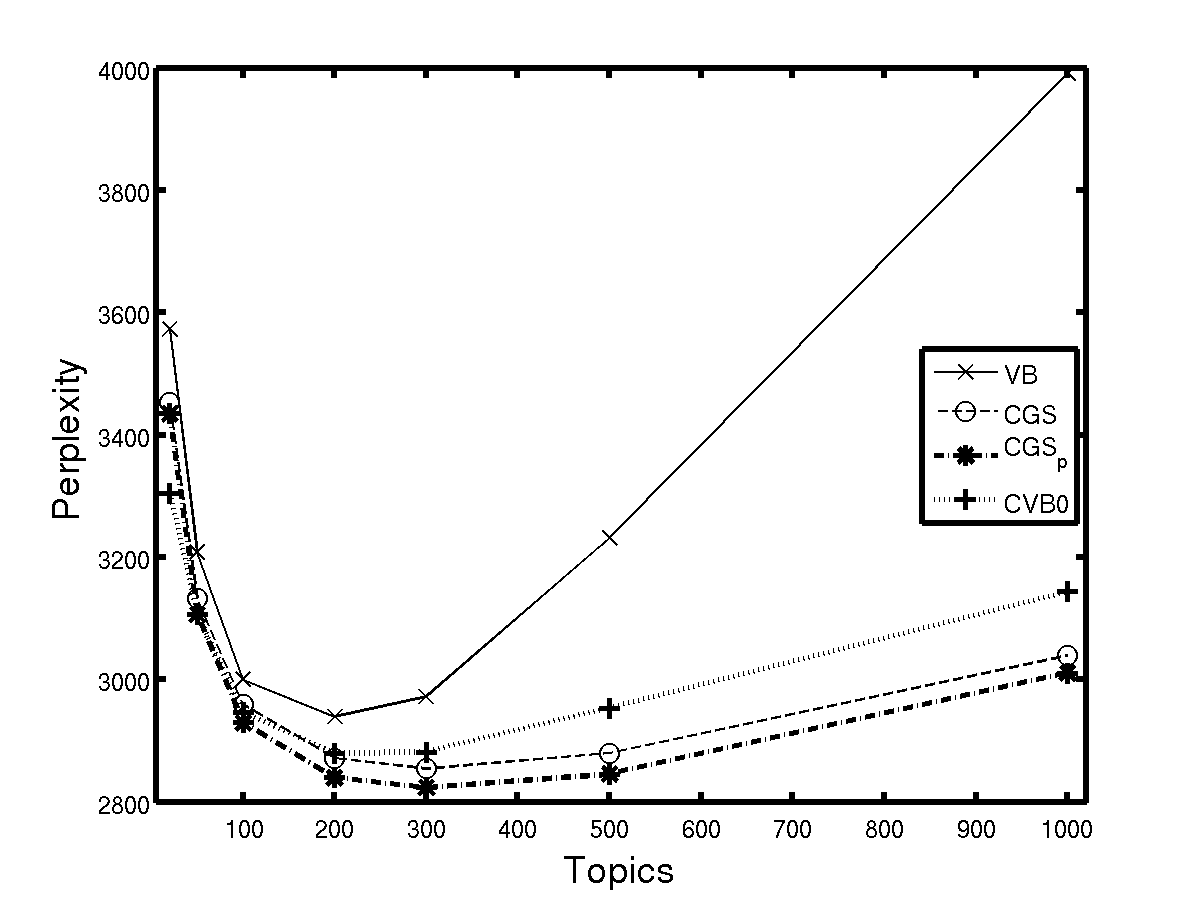}
  \subcaption{TASA}
\end{minipage}
\caption{Perplexity against number of topics for the CGS$_p$ method, standard CGS, CVB0 and VB. Results are taken by averaging over 5 different runs.}
\label{fig:ppx2}
\end{figure}

In this section we report in Figure  \ref{fig:ppx2}, additionally to the results of Section \ref{sec:exp2}, the perplexity across different topic configurations for VB against CVB0, CGS and CGS$_p$. We employed the lda-c package, publicly available at \url{https://github.com/blei-lab/lda-c}, keeping all parameters to default for the Variational EM procedure.

Apart from the New York Times data set, where VB manages to outperform CVB0 for $K>200$ and the CGS algorithms for $K=200$, the algorithm performs worse than other methods in all other cases. These results seem to be aligned with previous results in the literature \citep{DBLP:conf/nips/TehNW06, Asuncion:2009:SIT:1795114.1795118}.

\section{}
\label{app:timeoverhead}

\begin{figure}[tb]
\begin{minipage}{\textwidth}
\begin{minipage}{0.49\textwidth}
  \includegraphics[width=\linewidth]{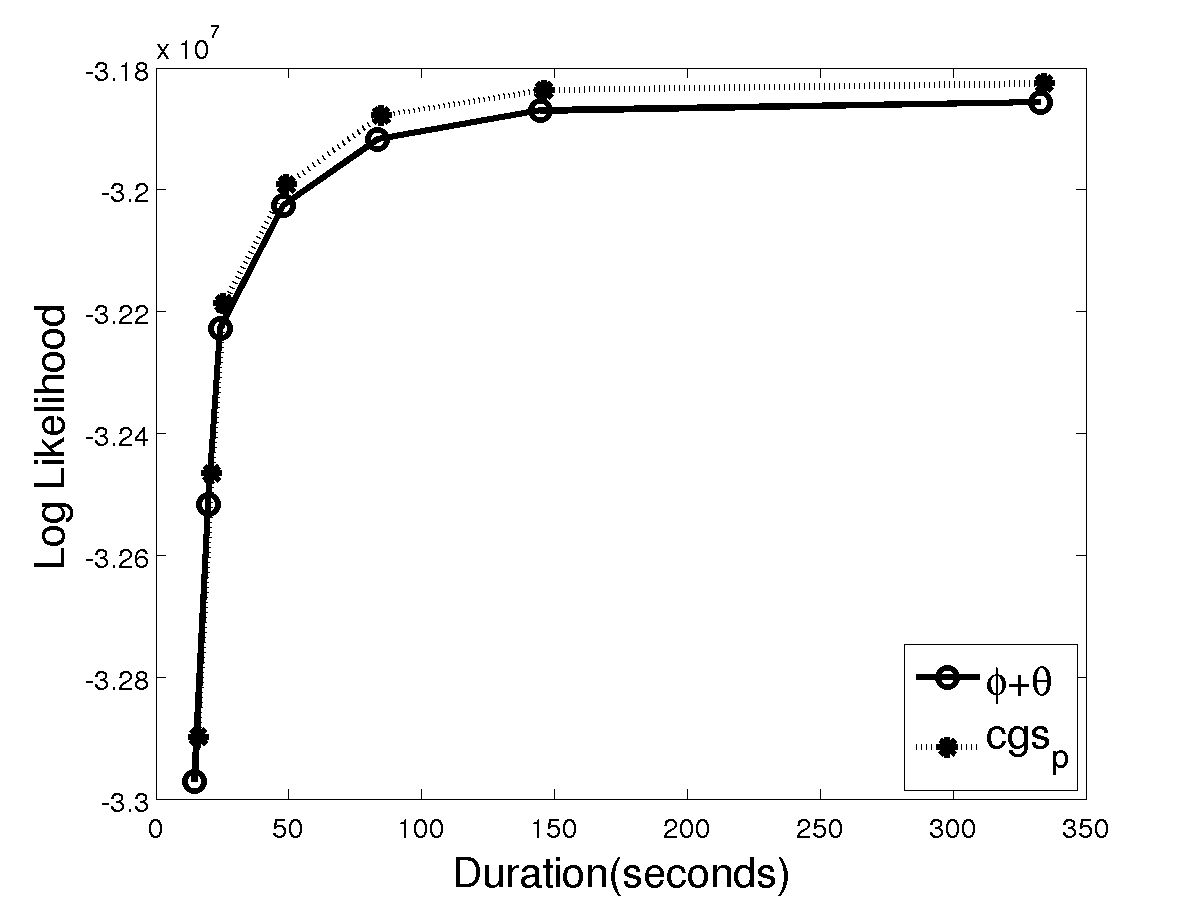}
  \subcaption{$K = 50$}
\end{minipage}  
\hspace*{\fill}
\begin{minipage}{0.49\textwidth}
  \includegraphics[width=\linewidth]{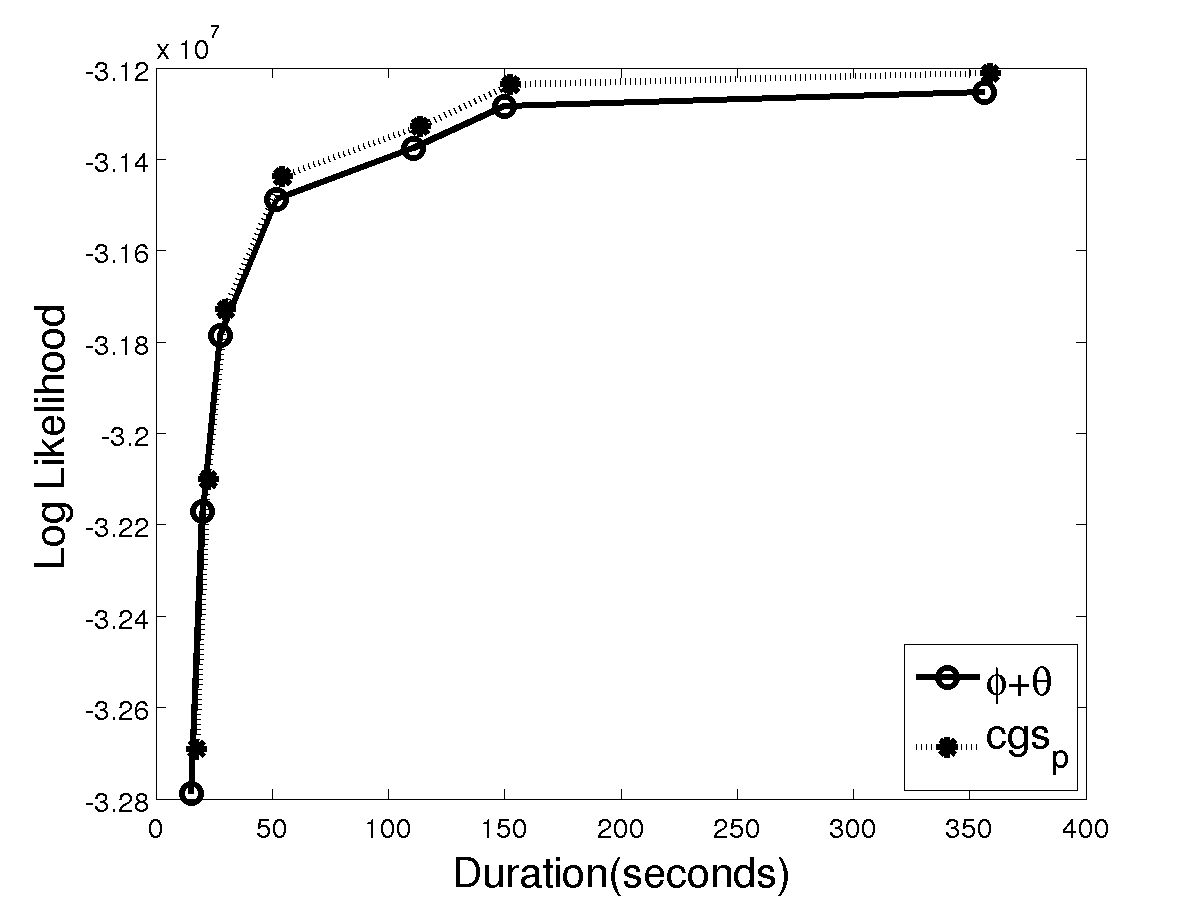}
  \subcaption{$K = 100$}
\end{minipage}
\end{minipage}
\begin{minipage}{\textwidth}
\begin{minipage}{0.49\textwidth}
  \includegraphics[width=\linewidth]{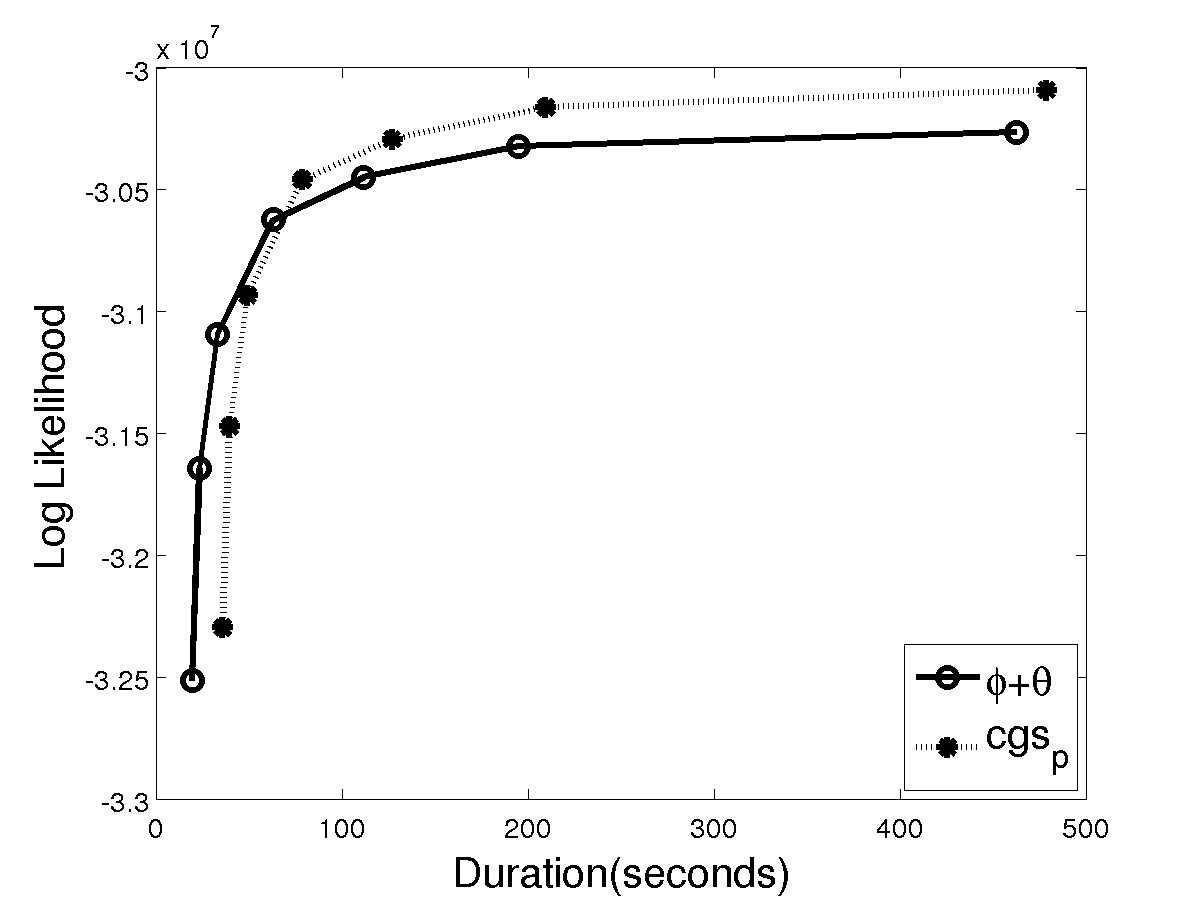}
  \subcaption{$K = 500$}
\end{minipage}  
\hspace*{\fill}
\begin{minipage}{0.49\textwidth}
  \includegraphics[width=\linewidth]{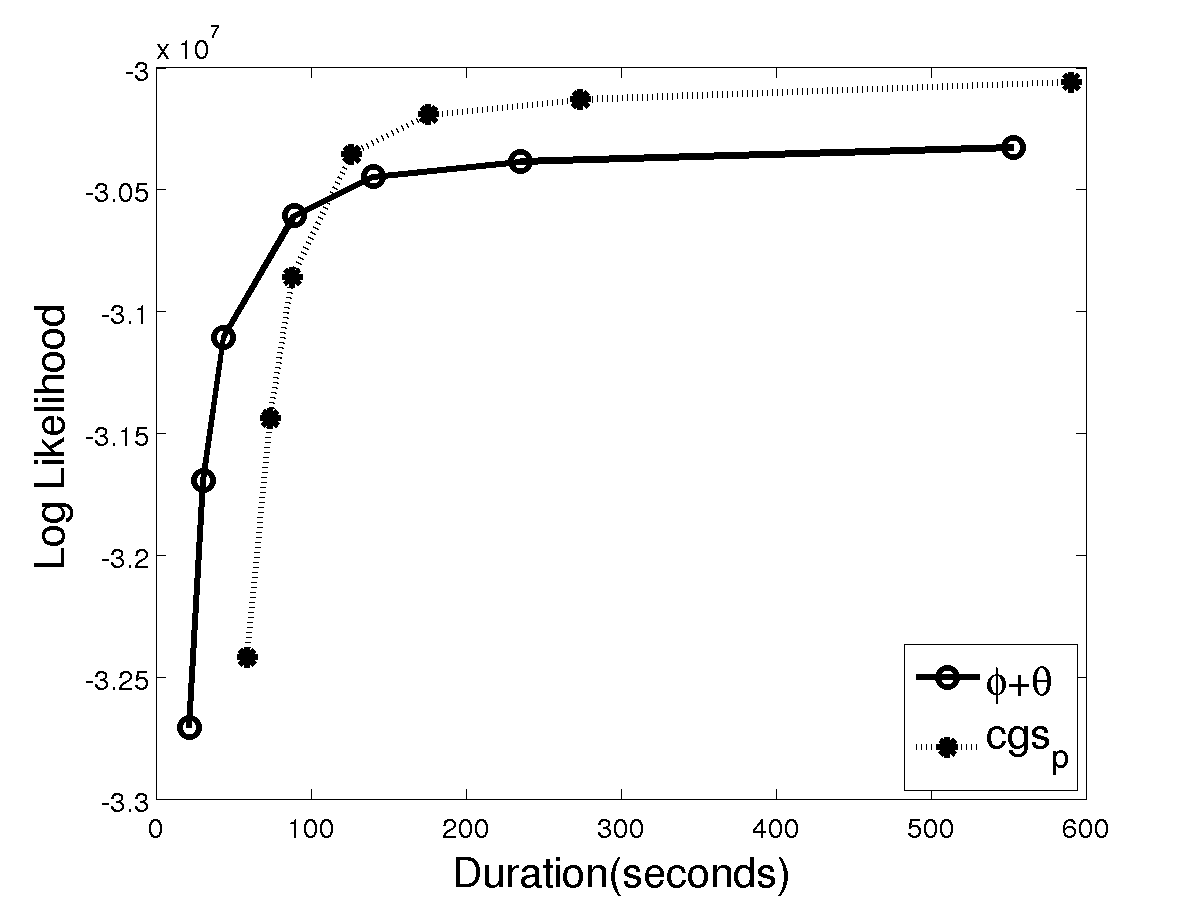}
  \subcaption{$K = 1000$}
\end{minipage}
\end{minipage}
\caption{Log Likelihood vs duration for CGS$_p$ against the standard CGS estimators on BioASQ, with WarpLDA. The time duration for each point in the plot corresponds to the running time of WarpLDA for different numbers of iterations (50, 100, 200, 500, 1000, 2000, 5000), plus the overhead needed to compute each of the estimators at that timestep.}
\label{fig:timeoverheadExp2}
\end{figure}

\begin{figure}[tb]
\begin{minipage}{\textwidth}
\begin{minipage}{0.49\textwidth}
  \includegraphics[width=\linewidth]{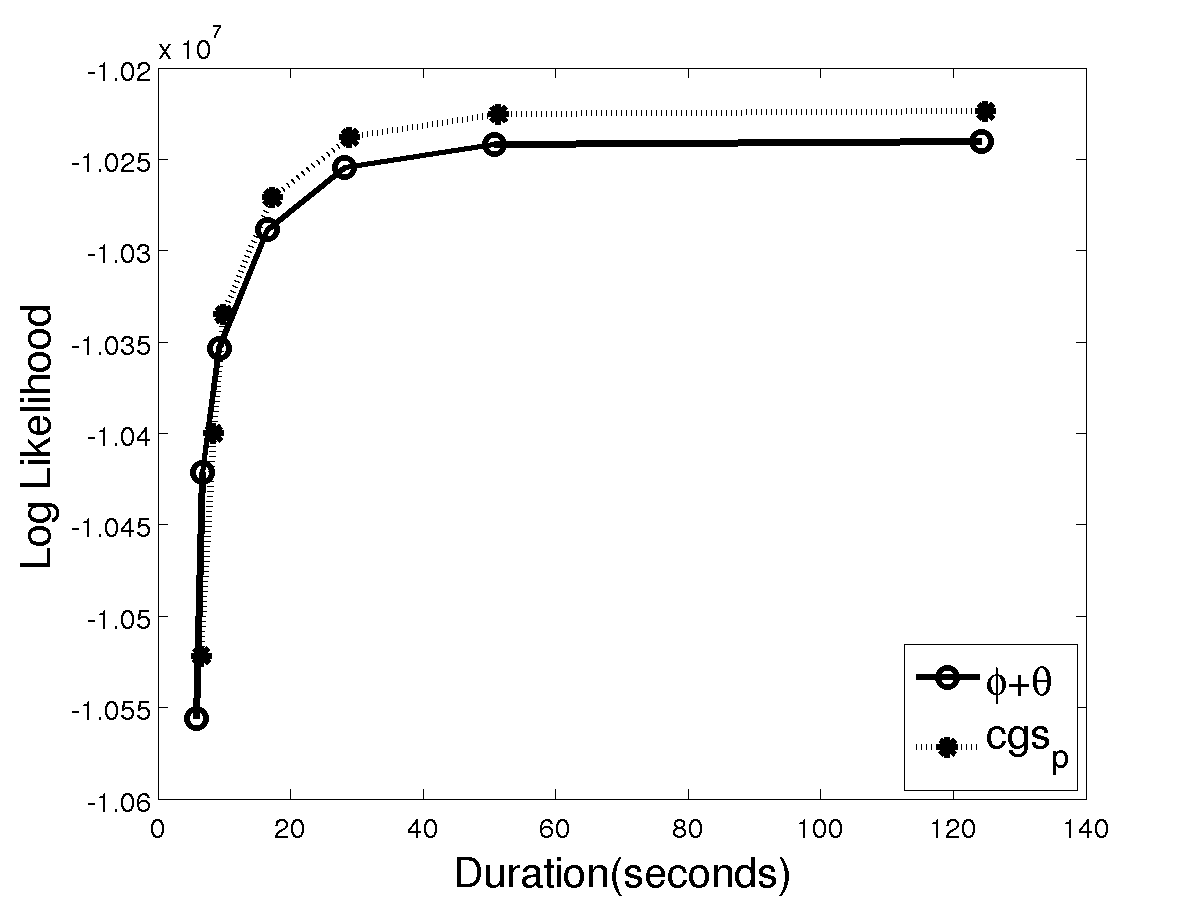}
  \subcaption{$K = 50$}
\end{minipage}  
\hspace*{\fill}
\begin{minipage}{0.49\textwidth}
  \includegraphics[width=\linewidth]{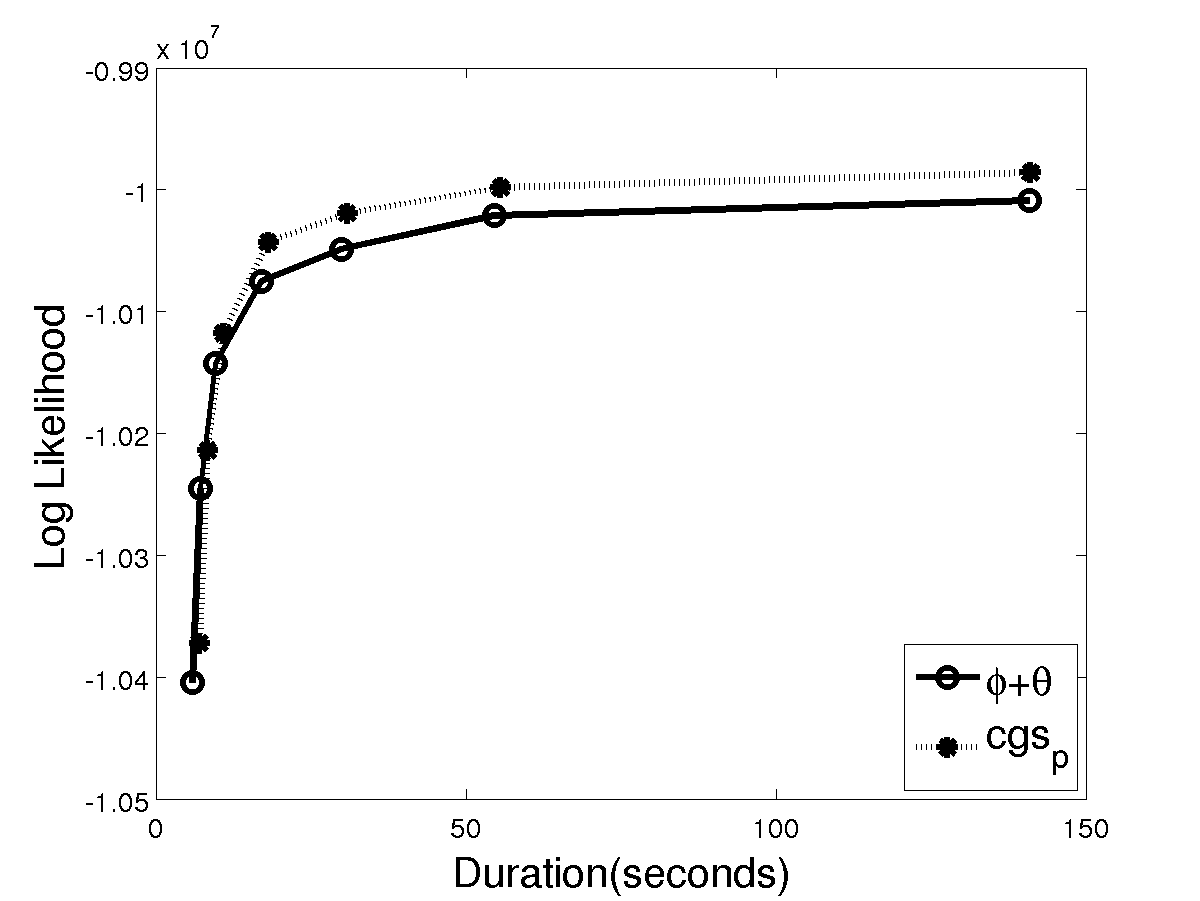}
  \subcaption{$K = 100$}
\end{minipage}
\end{minipage}
\begin{minipage}{\textwidth}
\begin{minipage}{0.49\textwidth}
  \includegraphics[width=\linewidth]{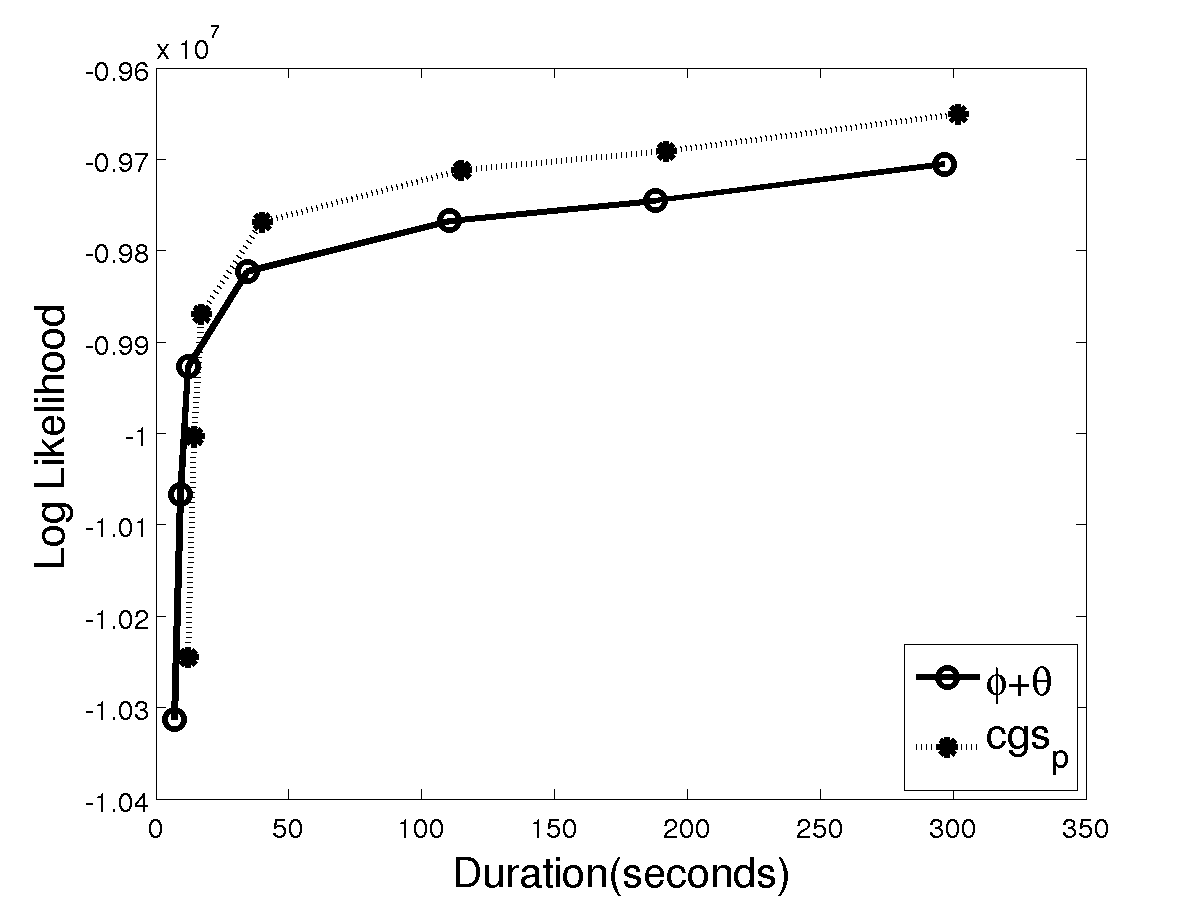}
  \subcaption{$K = 500$}
\end{minipage}  
\hspace*{\fill}
\begin{minipage}{0.49\textwidth}
  \includegraphics[width=\linewidth]{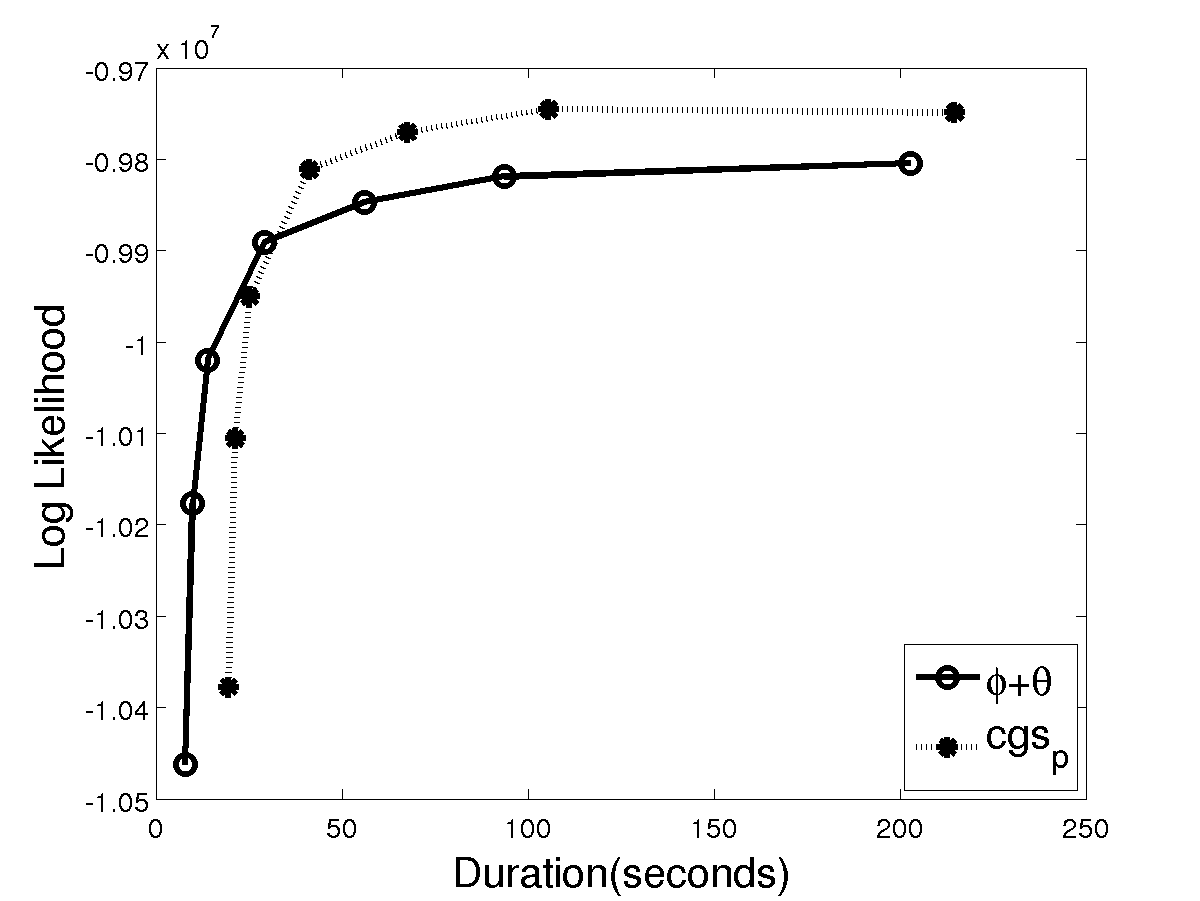}
  \subcaption{$K = 1000$}
\end{minipage}
\end{minipage}
\caption{Log Likelihood vs duration for CGS$_p$ against the standard CGS estimators on Reuters-21578, with WarpLDA. The time duration for each point in the plot corresponds to the running time of WarpLDA for different numbers of iterations (50, 100, 200, 500, 1000, 2000, 5000), plus the overhead needed to compute each of the estimators at that timestep.}
\label{fig:timeoverheadExp3}
\end{figure}

\begin{figure}[tb]
\begin{minipage}{\textwidth}
\begin{minipage}{0.49\textwidth}
  \includegraphics[width=\linewidth]{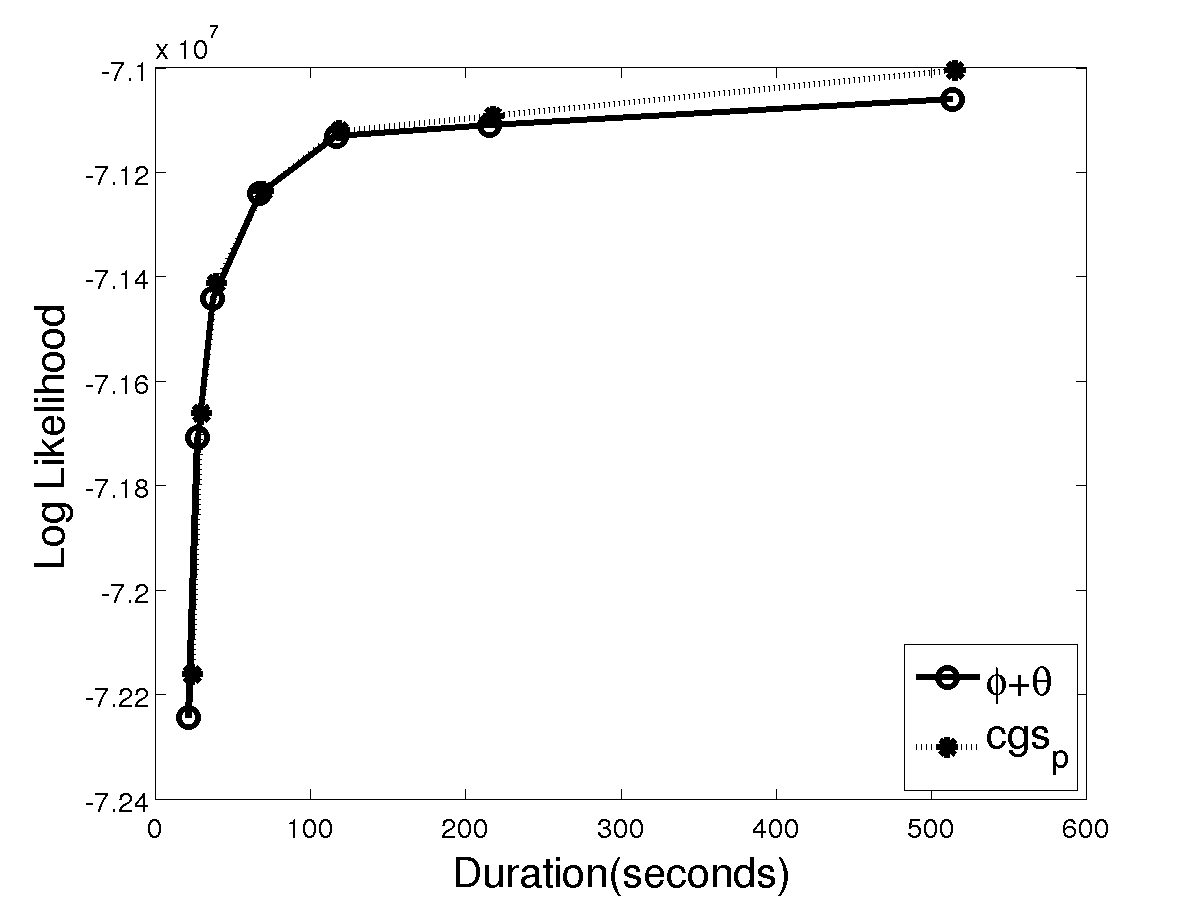}
  \subcaption{$K = 50$}
\end{minipage}  
\hspace*{\fill}
\begin{minipage}{0.49\textwidth}
  \includegraphics[width=\linewidth]{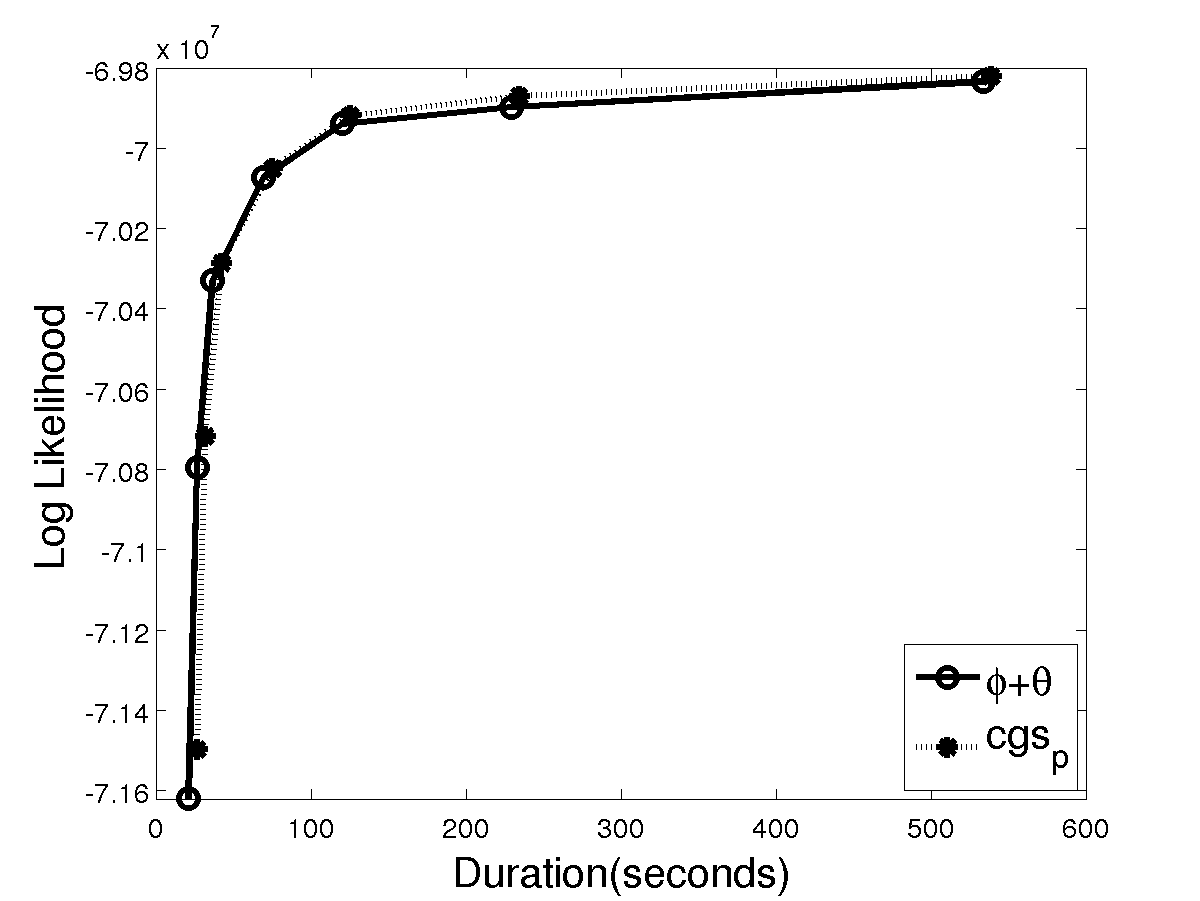}
  \subcaption{$K = 100$}
\end{minipage}
\end{minipage}
\begin{minipage}{\textwidth}
\begin{minipage}{0.49\textwidth}
  \includegraphics[width=\linewidth]{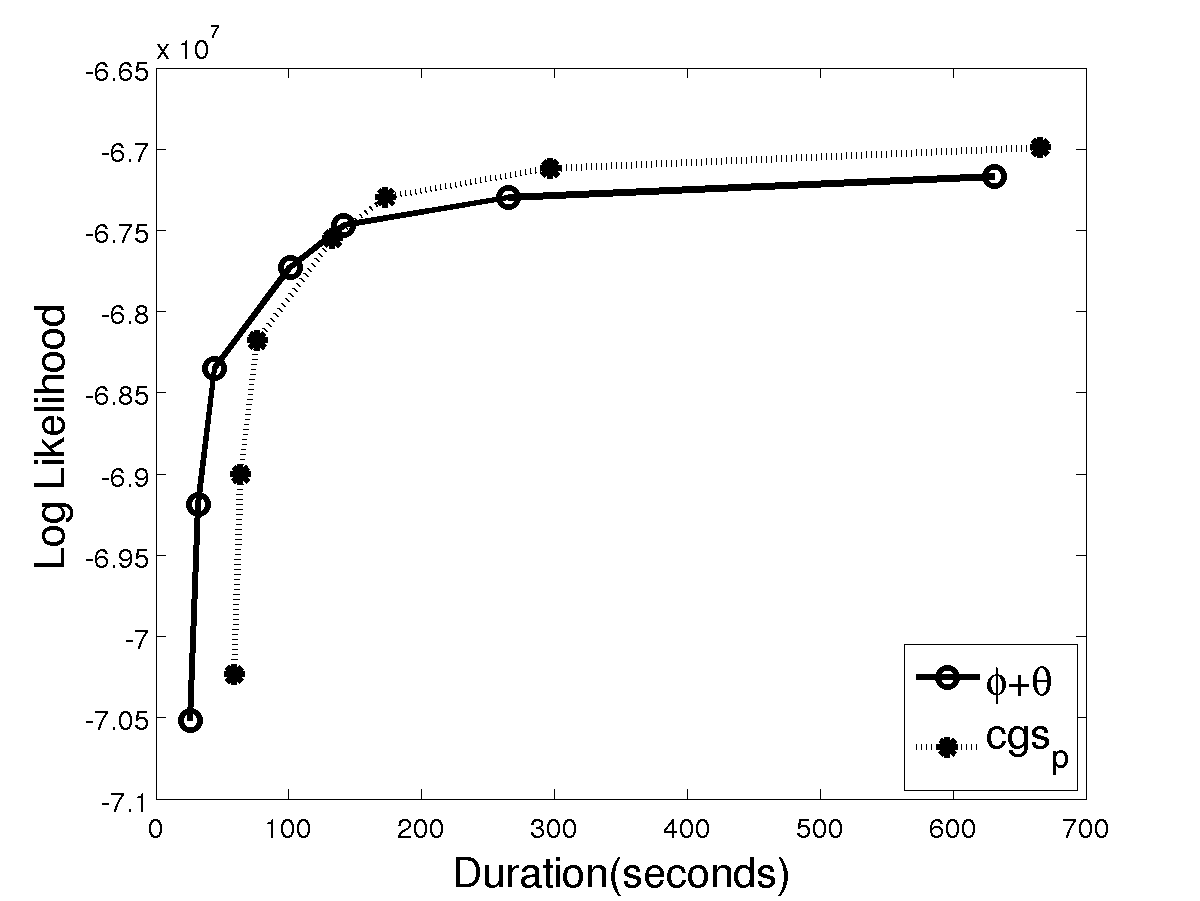}
  \subcaption{$K = 500$}
\end{minipage}  
\hspace*{\fill}
\begin{minipage}{0.49\textwidth}
  \includegraphics[width=\linewidth]{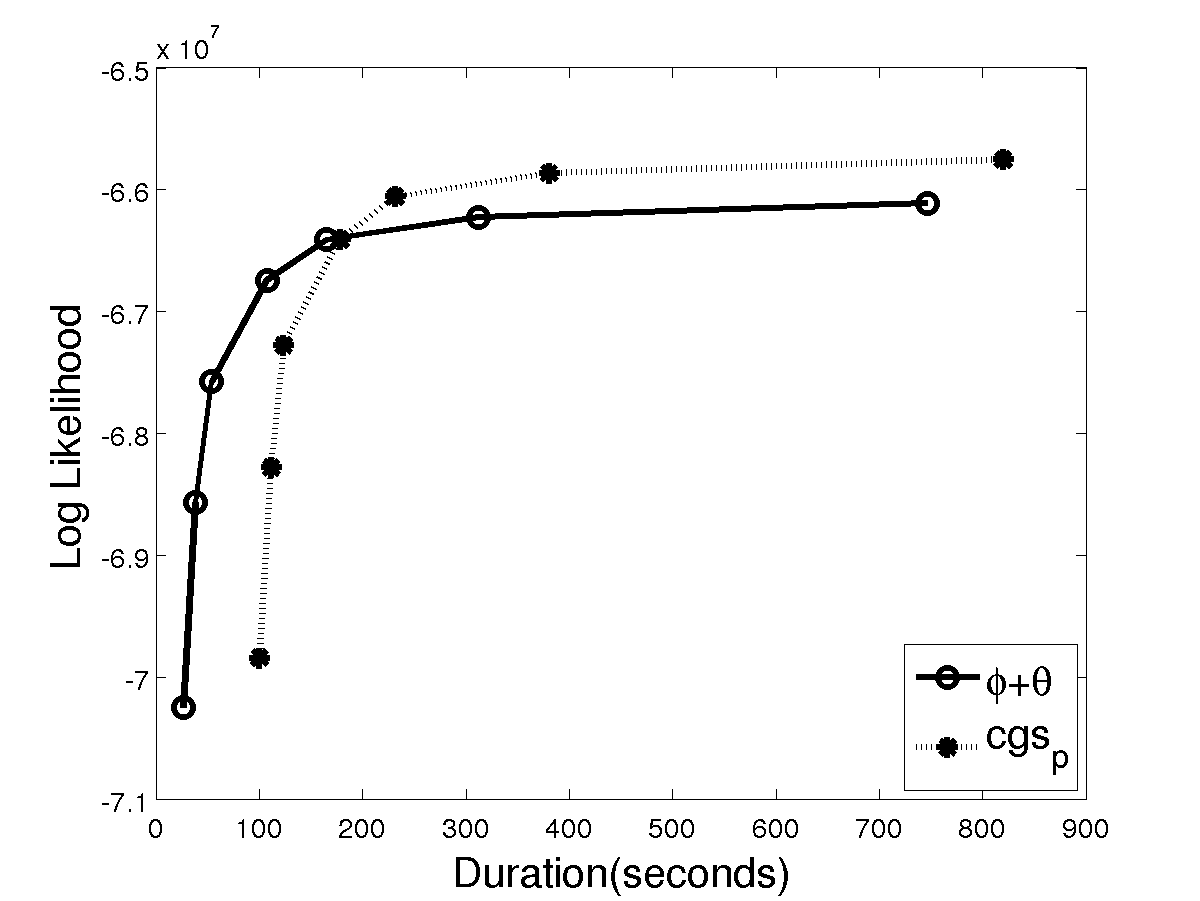}
  \subcaption{$K = 1000$}
\end{minipage}
\end{minipage}
\caption{Log Likelihood vs duration for CGS$_p$ against the standard CGS estimators on New York Times, with WarpLDA. The time duration for each point in the plot corresponds to the running time of WarpLDA for different numbers of iterations (50, 100, 200, 500, 1000, 2000, 5000), plus the overhead needed to compute each of the estimators at that timestep.}
\label{fig:timeoverheadExp4}
\end{figure}

We report in Figures \ref{fig:timeoverheadExp2} - \ref{fig:timeoverheadExp4} additional results for the rest of the data sets, for our experiment of section \ref{sec:overhead}, regarding the run-time overhead of the CGS$_p$ estimators. The results validate the conclusions drawn in section \ref{sec:overhead}.

\section{}
\label{app2}

\begin{figure}[t]
\begin{center}
	\begin{minipage}{0.44\textwidth}
   	\includegraphics[width=\linewidth]{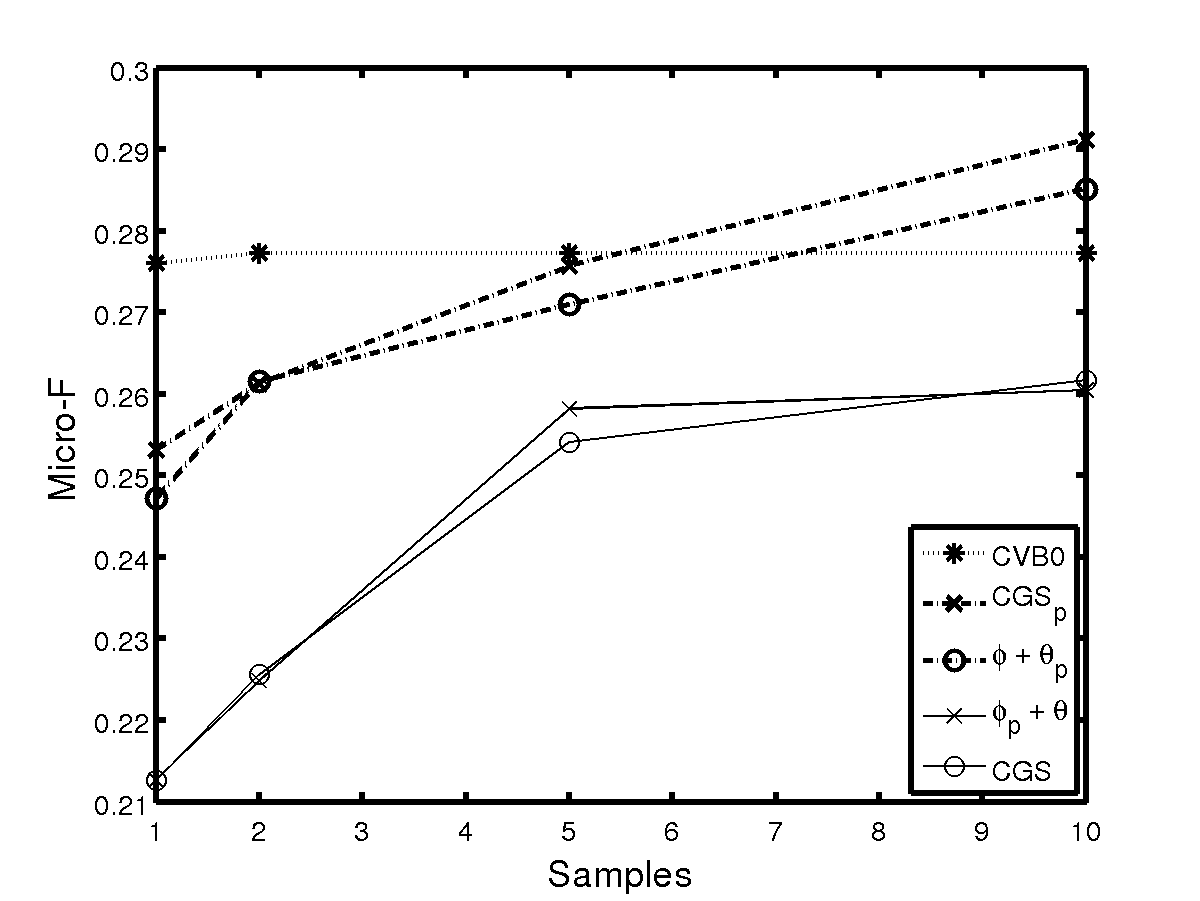}
 	\\
 	\includegraphics[width=\linewidth]{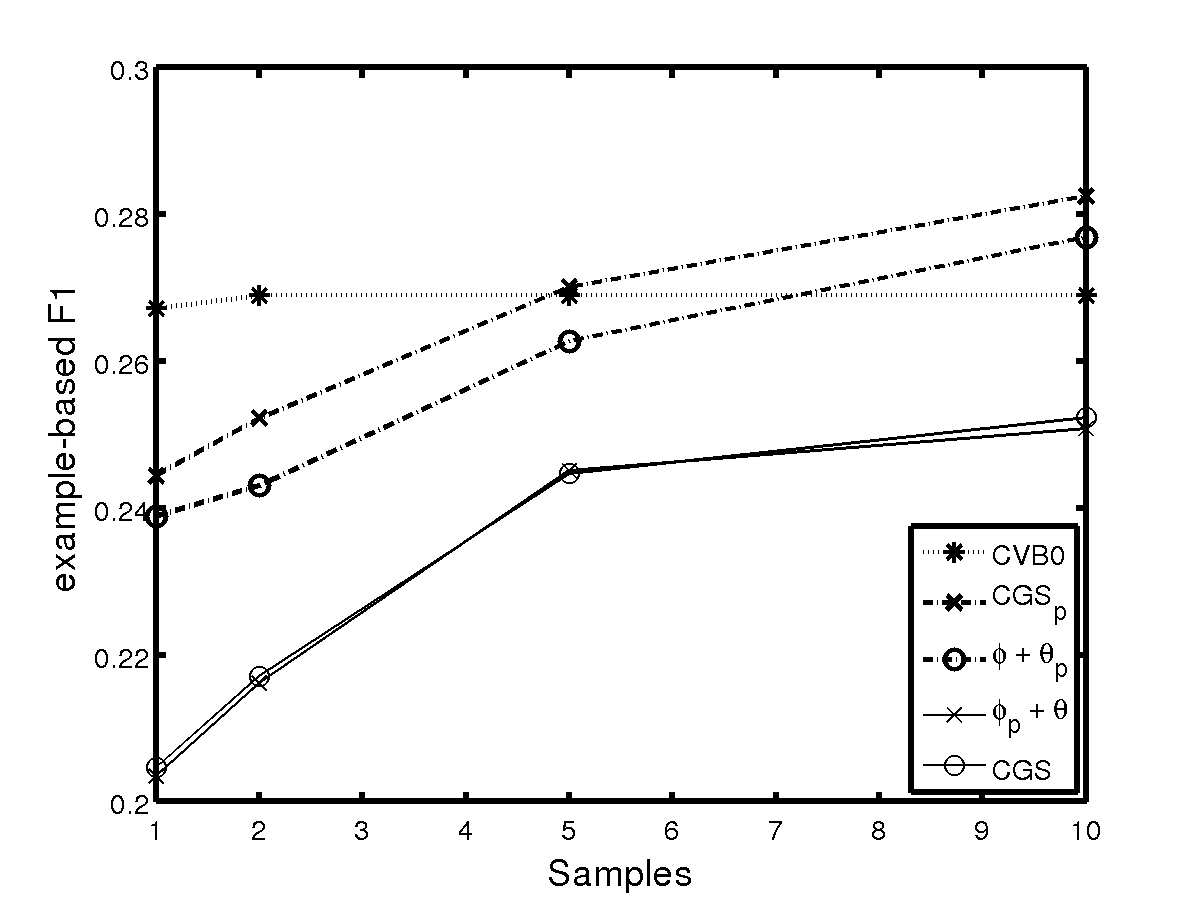}
 	\\
    	\includegraphics[width=\linewidth]{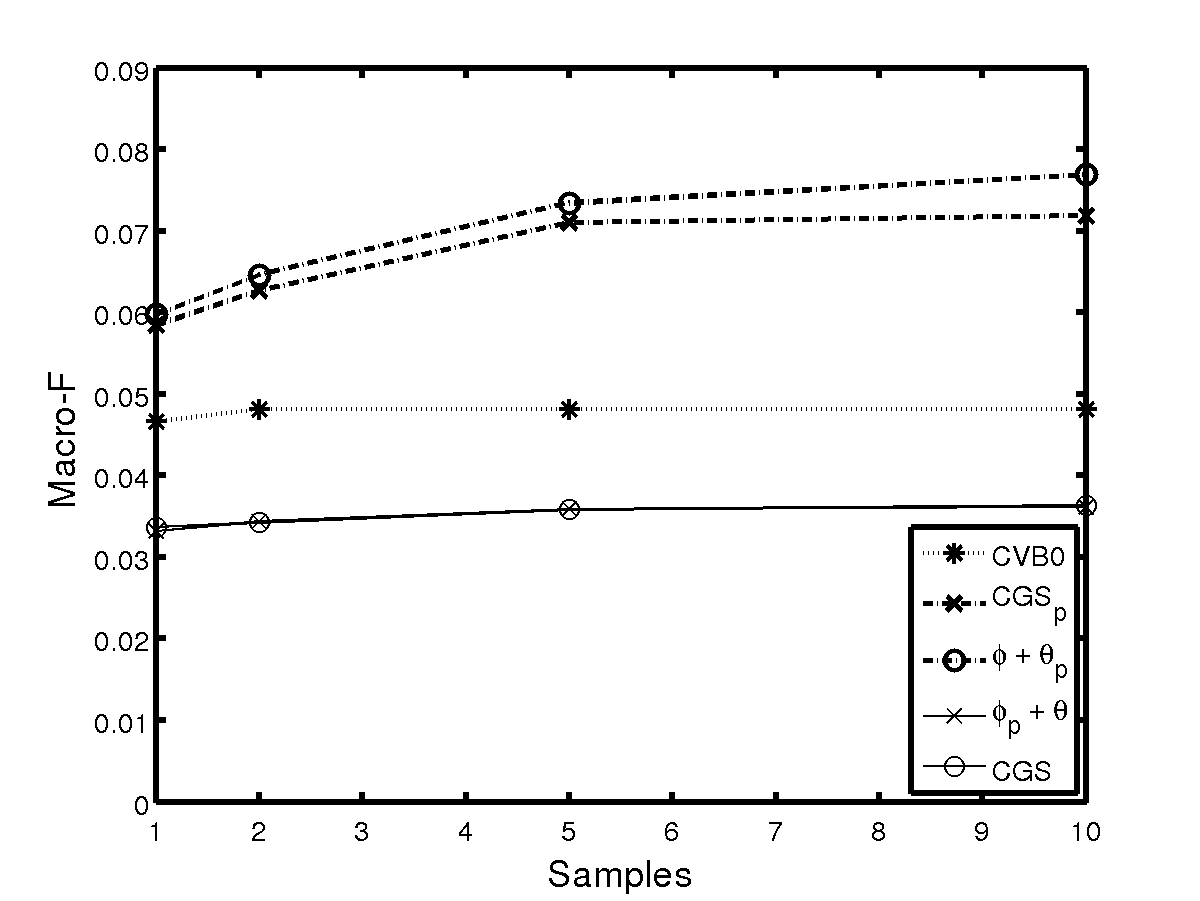}
    \caption{Delicious.}
    \label{fig:exp5a}
    \end{minipage}
	\begin{minipage}{0.44\textwidth}
   	\includegraphics[width=\linewidth]{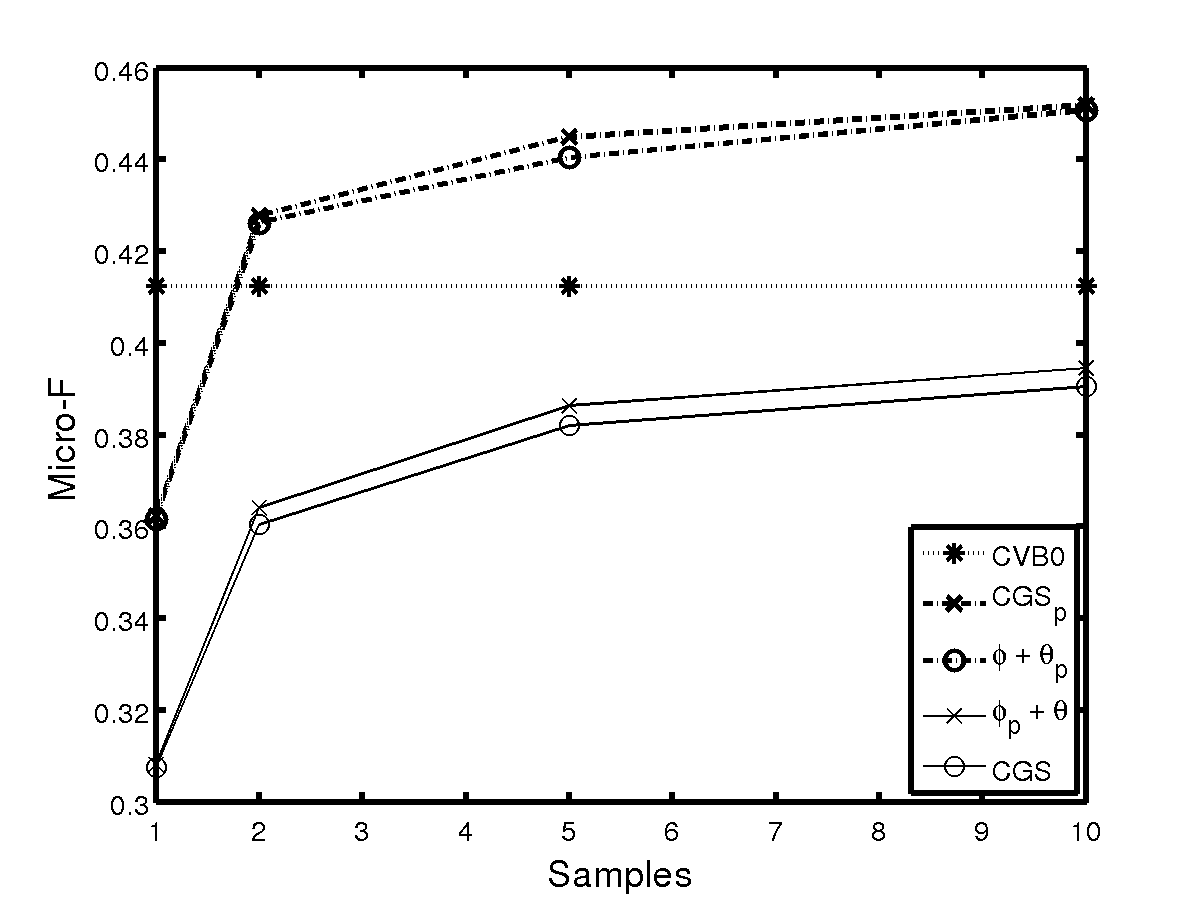}
 	\\
    	\includegraphics[width=\linewidth]{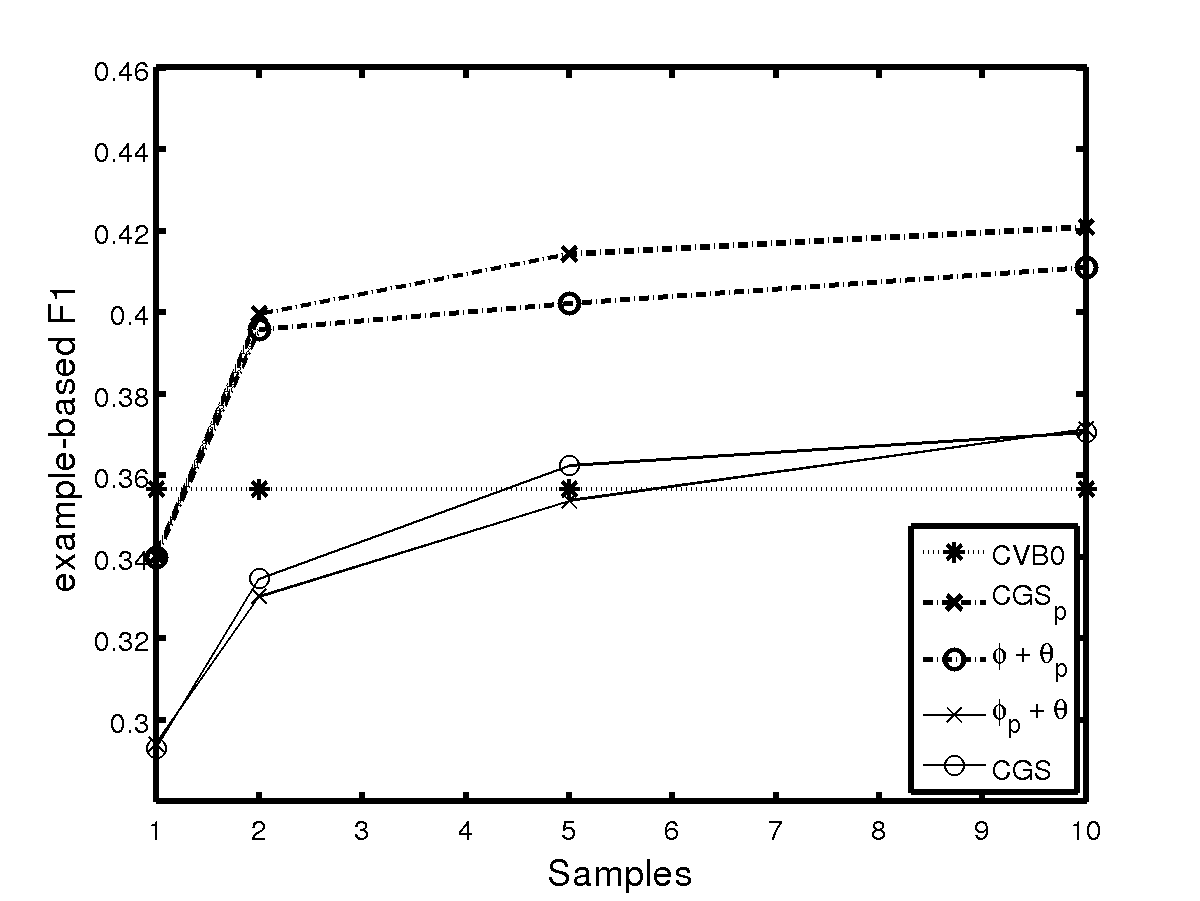}
 	\\
    	\includegraphics[width=\linewidth]{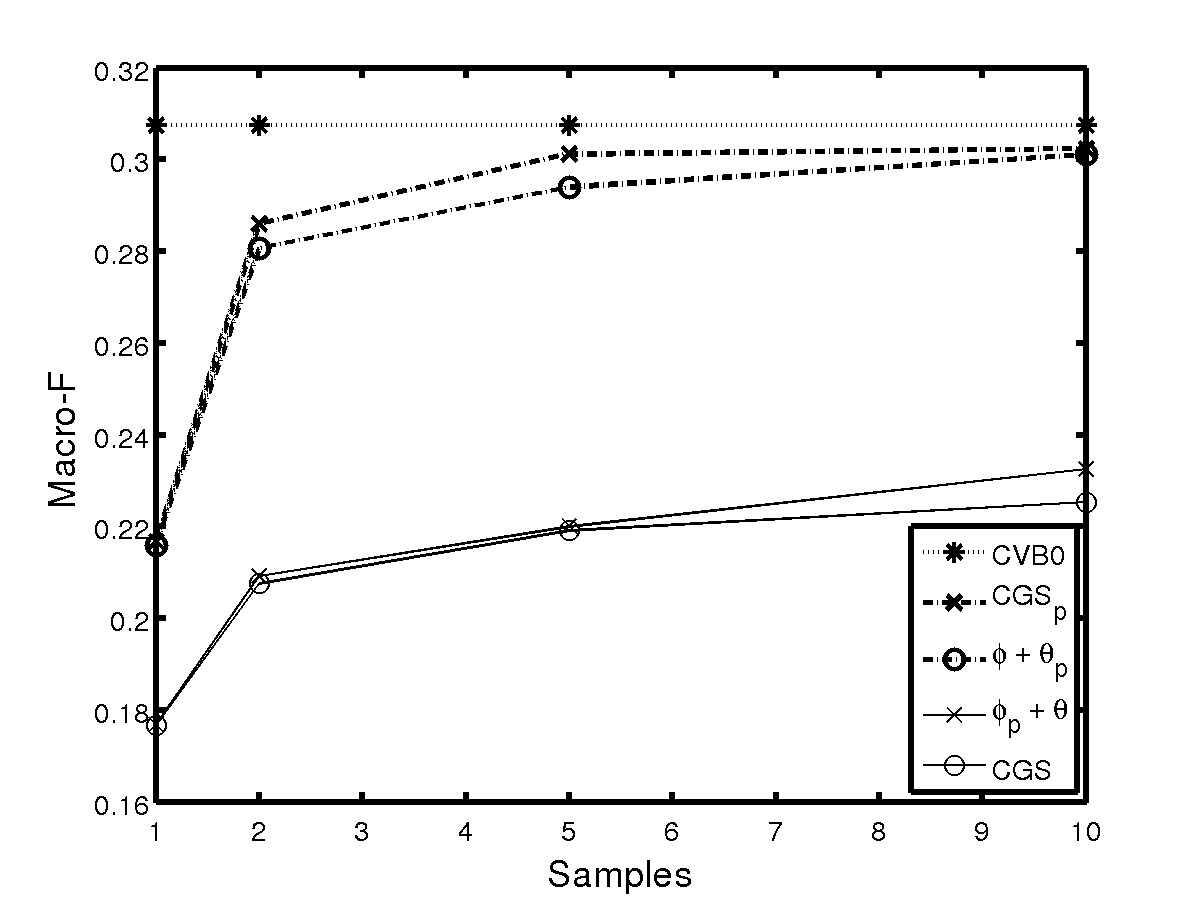}
	\caption{BioASQ.}
    \label{fig:exp5b}
	\end{minipage}

\end{center}
\end{figure}

\begin{figure}[t]
\begin{center}
	\begin{minipage}{0.44\textwidth}
   	\includegraphics[width=\linewidth]{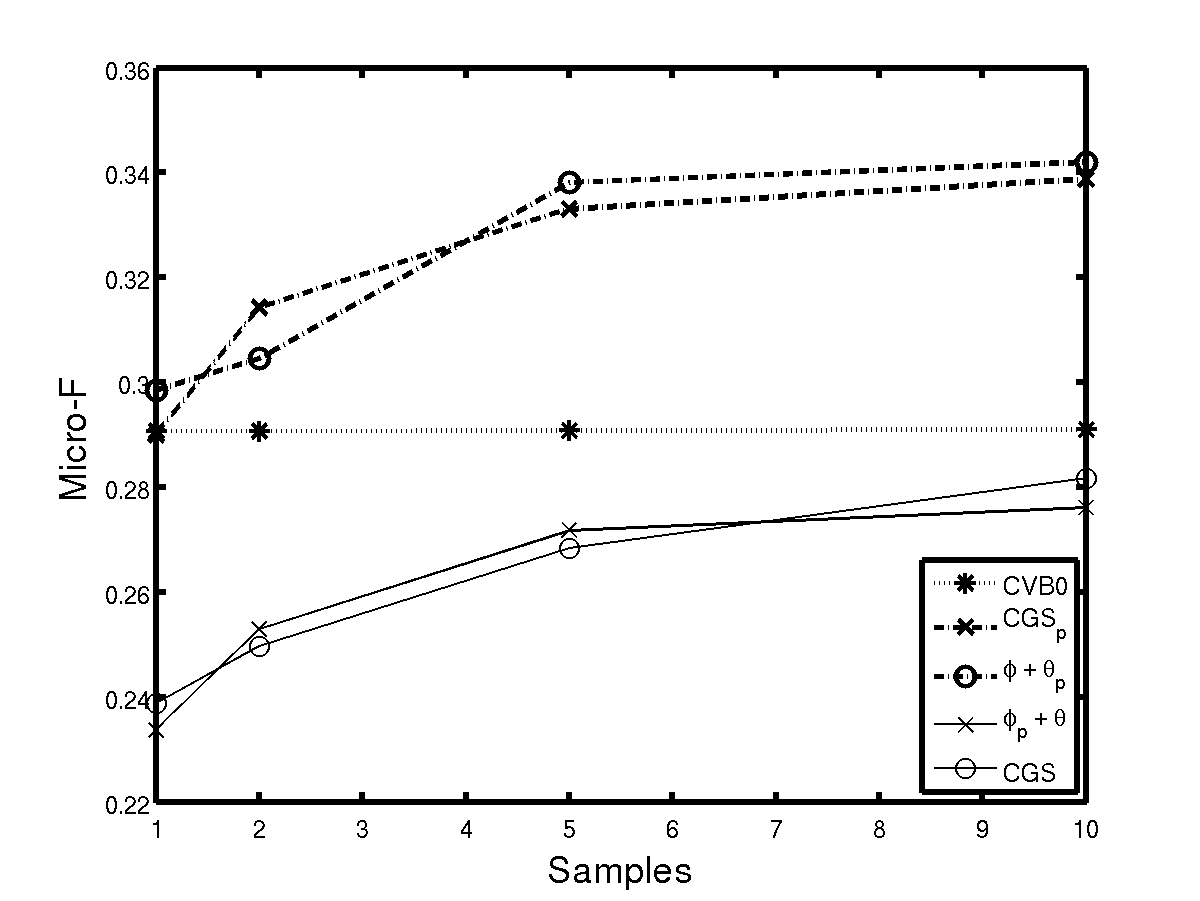}
 	\\
    	\includegraphics[width=\linewidth]{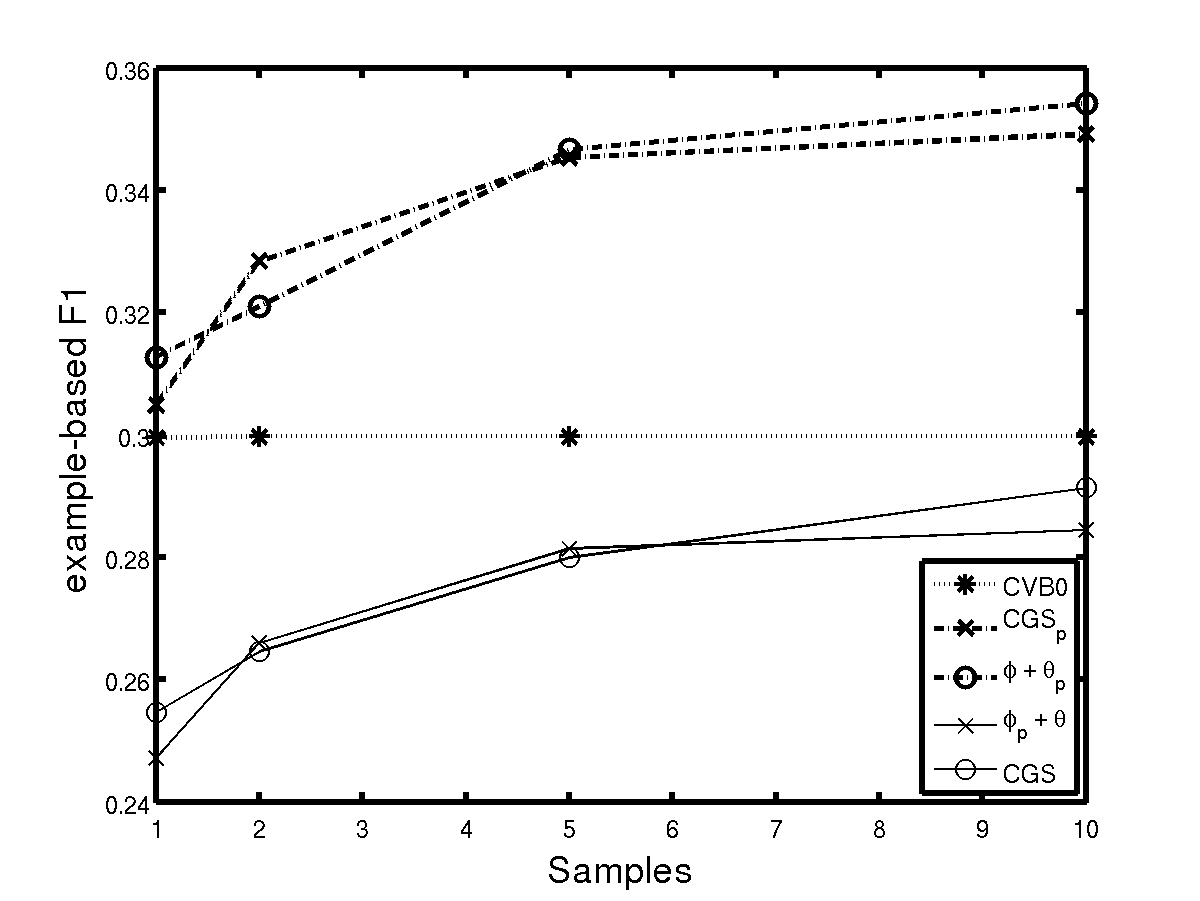}
 	\\
    	\includegraphics[width=\linewidth]{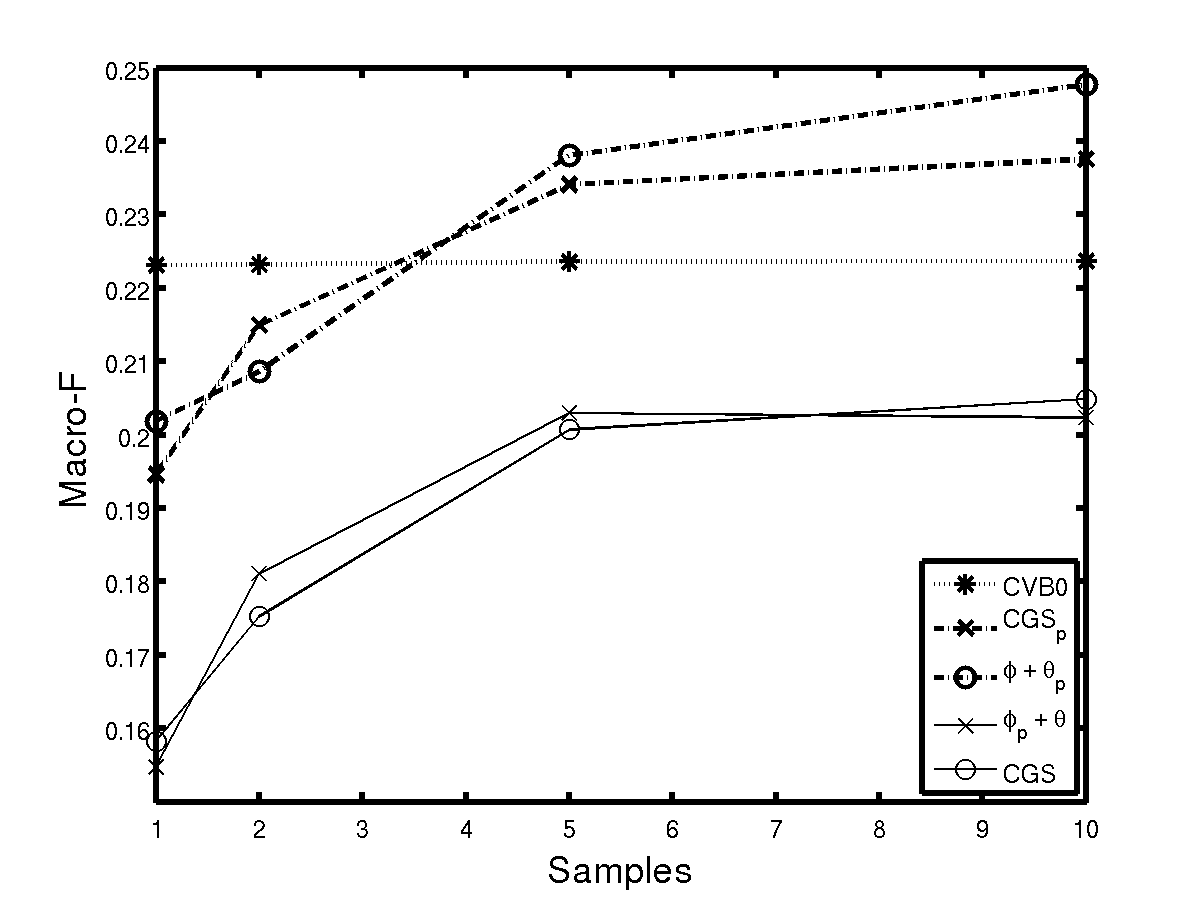}
	\caption{Bibtex.}
    \label{fig:exp5c}
	\end{minipage}
	\begin{minipage}{0.455\textwidth}
   	\includegraphics[width=\linewidth]{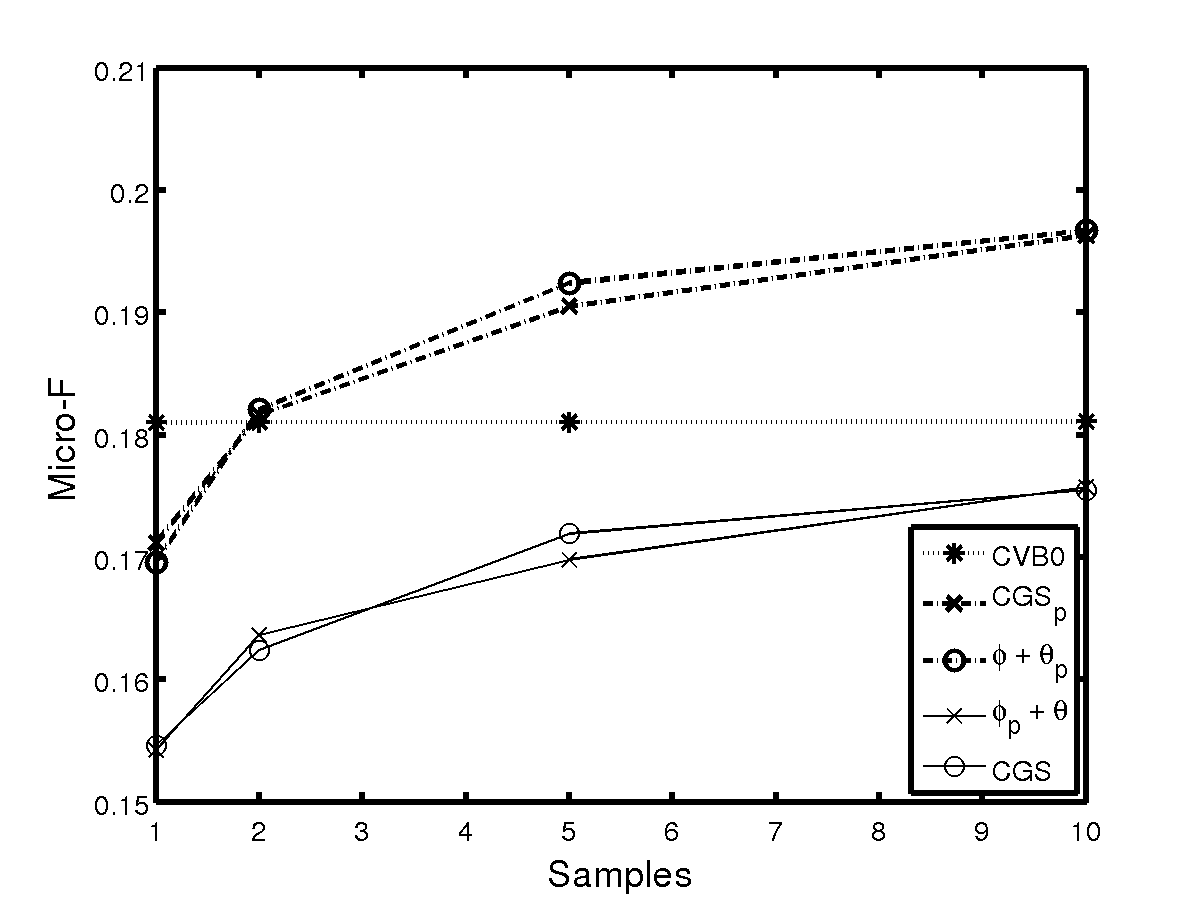}
 	\\
    	\includegraphics[width=\linewidth]{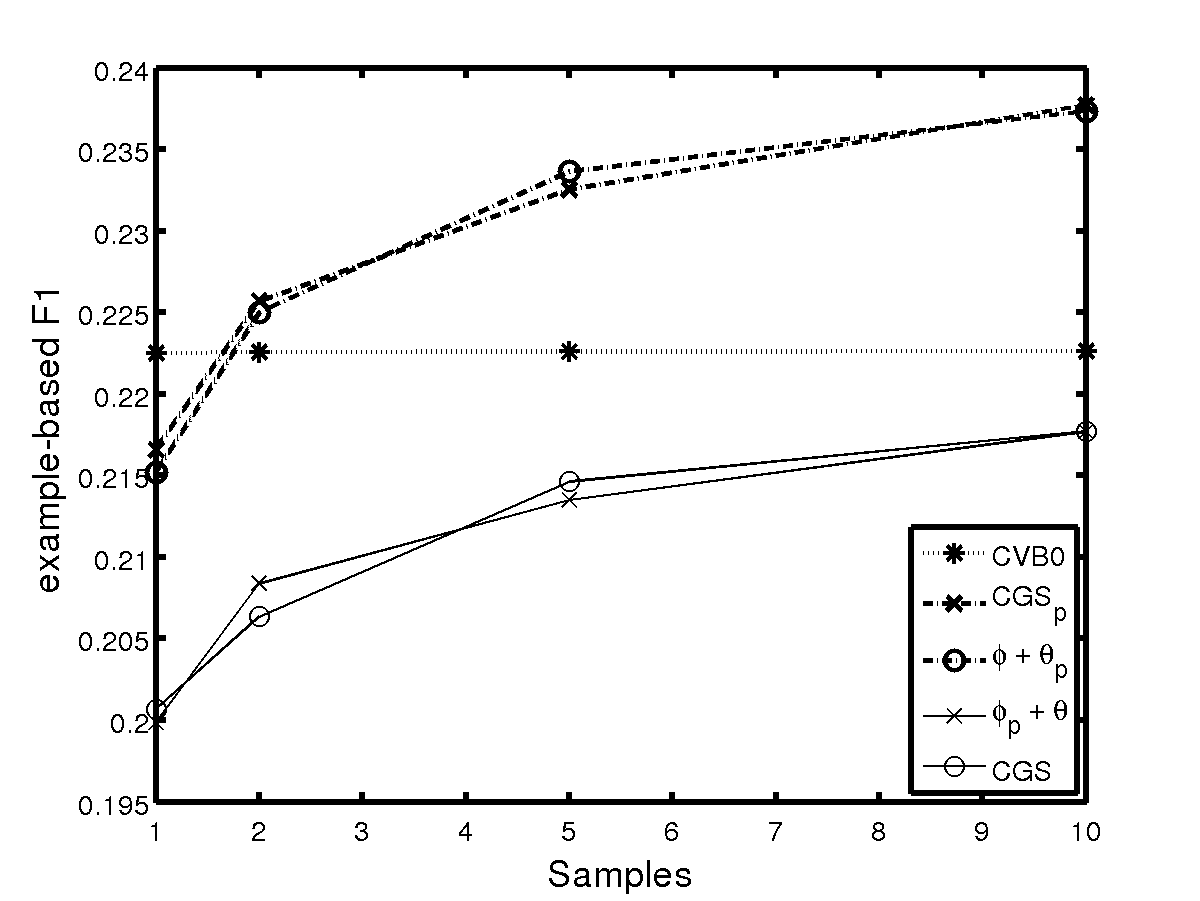}
 	\\
    \includegraphics[width=\linewidth]{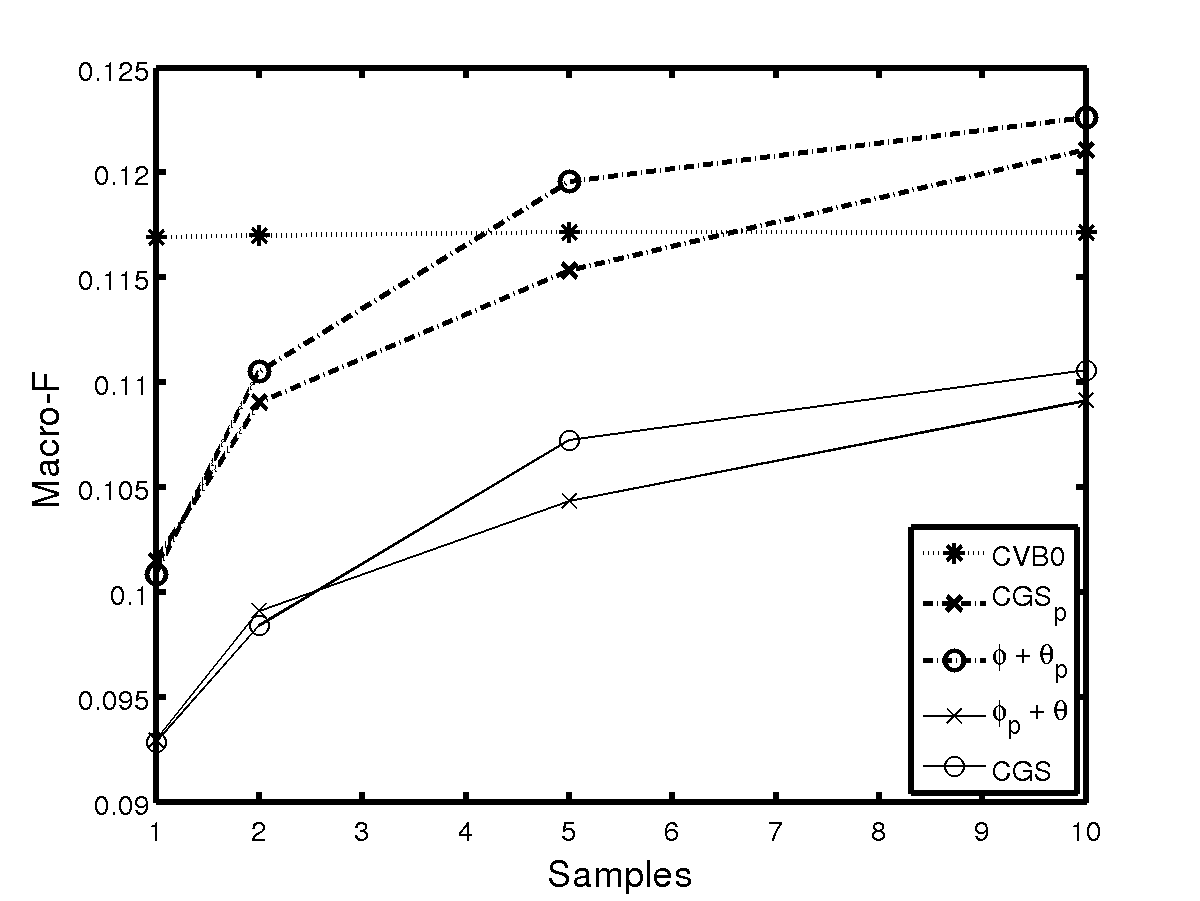}
	\caption{Bookmarks.}
    \label{fig:exp5d}
	\end{minipage}

\end{center}
\end{figure}

We report the results of an additional experiment that we performed in order to investigate how many samples CGS$_p$, $\phi+\theta^p$ and $\phi^p+\theta$ would require to outperform CVB0 (Figures \ref{fig:exp5a} -- \ref{fig:exp5d}). The same parameter setup was used as in the rest of the multi-label experiments of Section \ref{sec:multi}.

\bibliography{main}
\end{document}